%% file: main.tex
\documentclass[aps,pre,reprint,superscriptaddress,longbibliography,nofootinbib]{revtex4-2}
\pdfoutput=1

\usepackage[utf8]{inputenc} 
\usepackage[T1]{fontenc}    
\usepackage{hyperref}       
\usepackage{url}            
\usepackage{booktabs}       
\usepackage{amsfonts}       
\usepackage{nicefrac}       
\usepackage{microtype}      
\usepackage{xcolor}         

\usepackage{bm}
\usepackage{amsmath,amssymb,mathtools}
\usepackage{graphicx} 
\usepackage{caption} 
\usepackage[subrefformat=parens]{subcaption}
\usepackage{listings}  
\usepackage{comment}

\RequirePackage{algorithm}
\RequirePackage{algorithmic}

\input{math_commands}

\input{symbols}
\begin{document}

\title{Phases in a class of associative memories via hidden neurons}
\input{authors}

\input{abstract}
\maketitle  

\input{sec1}
\input{sec2}
\input{sec3}
\input{sec4}
\input{sec6}

\onecolumngrid
\appendix
\input{appendix}

\bibliography{references}

\end{document}

%% file: math_commands.tex
\usepackage{amsmath,amsfonts,bm}

\def\eqref#1{equation~\ref{#1}}

\def\1{\bm{1}}

\def\eps{{\epsilon}}

\def\rb{{\textnormal{b}}}

\def\rh{{\textnormal{h}}}

\def\rt{{\textnormal{t}}}

\def\vb{{\bm{b}}}

\def\vh{{\bm{h}}}

\def\vt{{\bm{t}}}

\DeclareMathAlphabet{\mathsfit}{\encodingdefault}{\sfdefault}{m}{sl}
\SetMathAlphabet{\mathsfit}{bold}{\encodingdefault}{\sfdefault}{bx}{n}

\newcommand{\E}{\mathbb{E}}

\newcommand{\R}{\mathbb{R}}

\newcommand{\softmax}{\mathrm{softmax}}

\DeclareMathOperator{\Tr}{Tr}
\let\ab\allowbreak

%% file: symbols.tex
\newcount\K 
\def\blah{
    \K=0 \loop\ifnum\K<20 
    {\color[rgb]{0.8, 0.8, 0.8} blah blah blah blah blah blah blah blah blah blah} 
    \advance\K by1\repeat 
}

\usepackage{mleftright}
\mleftright

\usepackage{amsthm}
\theoremstyle{plain}

\newtheorem*{thm*}{Theorem}

\newtheorem*{prop*}{Proposition}

\newtheorem*{cor*}{Corollary}

\newtheorem*{lem*}{Lemma}

\renewcommand{\H}{\mathcal{H}}

\newcommand{\Nv}{N_v} \newcommand{\tv}{{\tau_v}}  
\newcommand{\Lv}{{L_v}} \newcommand{\Ev}{{E_v}}
\newcommand{\xv}{v} \newcommand{\gv}{g}  

\newcommand{\Nh}{N_h} \renewcommand{\th}{{\tau_h}}  
\newcommand{\Lh}{{L_h}} \newcommand{\Eh}{{E_h}}
\newcommand{\xh}{h} \newcommand{\gh}{f}  
\newcommand{\xf}{\gh}

\newcommand{\Eint}{E_{\mathrm{int}}}
\newcommand{\xii}[2]{\xi^{(#1, #2)}}  
\newcommand{\coupling}{\lambda}  

\newcommand{\Ea}{{E_\xi^{\mathrm{A}}}} \newcommand{\za}{{Z_\xi^{\mathrm{A}}}} 
\newcommand{\Eb}{{E_\xi^{\mathrm{B}}}} \newcommand{\zb}{{Z_\xi^{\mathrm{B}}}} 
\newcommand{\Ec}{{E_\xi^{\mathrm{C}}}} \newcommand{\zc}{{Z_\xi^{\mathrm{C}}}} 

\newcommand{\xx}{\mathrm{x}}

\newcommand{\simplex}{\Delta}
\newcommand{\detp}{{\det}^\prime}
\newcommand{\hess}{\operatorname{Hess}}
\newcommand{\onevec}{\bm{1}}
\newcommand{\defname}[1]{\emph{#1}}

\DeclareMathOperator{\erf}{{erf}}
\DeclareMathOperator{\sech}{{sech}}

\newcommand{\githubrepo}{https://github.com/Toshihiro-Ota/classh}

\newcommand{\kh}{KH model}  

\newcommand{\mem}{p}

\newcommand{\ntextbf}[1]{\noindent\textbf{#1}}

\newcommand{\eref}[1]{Eq.~(\ref{#1})}
\newcommand{\fref}[1]{Fig.~\ref{#1}}
\newcommand{\tref}[1]{Table~\ref{#1}}
\newcommand{\sref}[1]{Sec.~\ref{#1}}
\newcommand{\appenref}[1]{Appendix~\ref{#1}}

\newcommand{\dd}[2]{\frac{d #1}{d #2}}

\newcommand{\set}[1]{\left\{ #1 \right\}}
\newcommand{\abs}[1]{\left\lvert #1 \right\rvert}
\newcommand{\norm}[1]{\left\lVert #1 \right\rVert}

\newcommand{\EE}[2]{\mathbb{E}_{#1}\left[ #2 \right]} 
\newcommand{\ev}[1]{\left\langle #1 \right\rangle} 

\makeatletter
\newsavebox{\@brx}
\newcommand{\llangle}[1][]{\savebox{\@brx}{\(\m@th{#1\langle}\)}%
  \mathopen{\copy\@brx\mkern2mu\kern-0.9\wd\@brx\usebox{\@brx}}}
\newcommand{\rrangle}[1][]{\savebox{\@brx}{\(\m@th{#1\rangle}\)}%
  \mathclose{\copy\@brx\mkern2mu\kern-0.9\wd\@brx\usebox{\@brx}}}
\makeatother

\newcommand{\diag}{\mathop{\mathrm{diag}}\nolimits}
\newcommand{\ii}{\mathrm{i}}

\DeclareMathOperator{\sgn}{sgn}

\newcommand{\Z}{\mathbb{Z}}

\renewcommand{\P}{\mathbb{P}}

%% file: authors.tex
\newcommand{\ca}{CyberAgent AI Lab, Shibuya, Tokyo 150--0002, Japan}
\newcommand{\rikkyo}{Graduate School of Artificial Intelligence and Science, Rikkyo University, Toshima, Tokyo 171--8501, Japan}
\newcommand{\ithems}{RIKEN iTHEMS, Wako, Saitama 351--0198, Japan}

\affiliation{\ca}
\affiliation{\rikkyo}
\affiliation{\ithems}

\author{Toshihiro Ota}
    \email{ota\_toshihiro@cyberagent.co.jp}
    \affiliation{\ca}
    \affiliation{\ithems}
\author{Masato Taki}
    \email{taki\_m@rikkyo.ac.jp}
    \affiliation{\rikkyo}
    \affiliation{\ithems}

%% file: abstract.tex
\begin{abstract}

Associative memory in the Hopfield network is attractor dynamics in a disordered many-body system, and higher-order and exponential extensions turn its retrieval update into softmax attention.
The polynomial and exponential regimes have been analyzed by different methods, with no common architecture in which to ask what fixes the storage scale.
In this paper we study the bipartite architecture of Krotov and Hopfield, which we call the class $\mathcal{H}$, whose model is fixed by a Lagrangian for each layer, taking the hidden neurons as the order parameter of retrieval.
At polynomial load the replica method yields the replica-symmetric phase diagrams and closed-form capacities, and the crosstalk moment is common to Ising and spherical visible neurons, so their differences come from the visible entropy.
With a softmax hidden layer the load is exponential, and a copy representation maps the thermodynamics onto random-energy-model counting, with paramagnetic, condensed, and frozen phases.
Heating destabilizes retrieval by quantized reassignments of attention, and typical Gaussian patterns remain metastable at every load.
The regimes differ in their crosstalk statistics, central-limit at polynomial load and large-deviation at exponential load, and the class $\mathcal{H}$ splits retrieval into two roles, the visible Lagrangian fixing stability and the hidden one the storage scale, two axes that may also guide the design of new Lagrangians.

\end{abstract}

%% file: sec1.tex
\section{Introduction}
\label{sec:introduction}

Associative memory is a content-addressable mechanism that retrieves a whole memory from a partial cue.
The Hopfield network formalized this retrieval as the collective dynamics of many interacting neurons, with the stored patterns realized as attractors of an energy landscape \cite{hopfield82}.
The network is thereby a disordered many-body system, and the replica method of spin-glass statistical mechanics quantified the competition between the retrieval and the spin-glass phases and the storage capacity of the model \cite{amit1985spin,amit1985storing,amit1987statistical}.

This framework has since been extended considerably \cite{krotov2023new,krotov2025modern}.
Replacing the pairwise interactions by higher-order ones raises the number of retrievable patterns from linear in the number of neurons to polynomial, with a degree that grows with the order of the interaction \cite{gardner1987multiconnected,abbott1987storage,baldi1987number,NIPS2016_eaae339c}, and exponential interactions make it exponentially large in the number of neurons \cite{demircigil2017model}.
With a log-sum-exp energy the retrieval update takes the form of softmax attention, placing associative memory in direct correspondence with the attention mechanism of the transformer \cite{NIPS2017_3f5ee243,ramsauer2021hopfield,ota2023attention}.

The two regimes have been analyzed with different tools: the replica method at polynomial load, and large-deviation and extreme-value statistics at exponential load, which control both the retrieval thresholds \cite{lucibello2024exponential} and the finite-temperature transitions \cite{koulischer2023exploring,hu2024provably,petrova2026geometric}.
What is missing is a setting in which one can ask, within a single architecture, what fixes the storage scale and how the crosstalk statistics change with it.
The bipartite architecture of Krotov and Hopfield provides one \cite{krotov2020large}: a visible and a hidden layer coupled only by pairwise interactions, in which the choice of a Lagrangian for each layer alone determines the model.
We call this two-layer family the \defname{class $\H$}, the $k=2$ member of a hierarchical class $\H_k$ of $k$ coupled layers, and describe it in \sref{sec:preliminaries}.
Its three representatives singled out in \cite{krotov2020large}, studied so far as separate models with different effective energies, are Model A, with Ising visible neurons, Model B, whose hidden layer is a softmax, and Model C, with spherical visible neurons.
We take a further step and treat the hidden neurons themselves as the order parameter of memory retrieval, which describes different visible geometries and hidden nonlinearities in one language.

\sref{sec:models_a_c} treats the statistical mechanics of Models A and C at polynomial load, where the replica method yields the replica-symmetric phase diagrams, with the zero-temperature capacities in closed form.
The moment that measures the crosstalk of the non-retrieved patterns is common to the two models, while their retrieval phases differ sharply: the quadratic spherical model is marginal and stores nothing \cite{bolle2003spherical}, whereas the higher-order interaction restores a retrieval phase.
In \sref{sec:model_b}, we study Model B, whose hidden sector reduces to the attention weights assigned to the stored patterns, so that retrieval is the concentration of attention.
Its natural load is exponential, and a representation of the thermodynamics in terms of a temperature-dependent number of copies, each selecting one stored pattern, turns the problem into one of counting with the structure of the random energy model \cite{derrida1981random,monasson1995structural}.
This yields the paramagnetic, condensed, and frozen phases, and shows that heating destabilizes retrieval not by a smooth erosion of the overlap but by quantized reassignments of attention, the retrieval of a typical Gaussian pattern remaining metastable at every load.

What separates the two regimes is the character of the crosstalk statistics, central-limit and insensitive to the pattern ensemble at polynomial load, large-deviation and ensemble-dependent at exponential load.
One consequence is that at exponential load the rare Gaussian patterns of atypically large norm lie below typical retrieval in free energy, so that a typical memory is never the equilibrium phase and the network operates as a metastable device.
Underlying both statements is the picture the class $\H$ supplies, in which the retrieval problem separates into two roles carried by the two Lagrangians.
The visible Lagrangian fixes, through the visible entropy, the stability of retrieval under a given crosstalk, which is what isolates the difference between Models A and C.
The hidden Lagrangian fixes the storage scale and the character of the disorder statistics, a polynomial nonlinearity yielding polynomial load and central-limit crosstalk, and the log-sum-exp yielding exponential load and large deviations.
In these terms the sequence from the classical Hopfield network through dense associative memory to attention is not a succession of separate theories, but a set of positions on these two axes, compared in the common order-parameter language that the hidden neurons provide.%
\footnote{The code for our numerical experiments is available at \url{\githubrepo}.}

%% file: sec2.tex
\section{Preliminaries}
\label{sec:preliminaries}

To fix notation, in this section we give an overview of the class $\H$, originally proposed in \cite{krotov2020large}, and provide the statistical mechanical setup for our main discussions in the subsequent sections.
Details of the class $\H$ and of the more general class-$\H_k$ associative memories are presented in \appenref{append:class_hk}.

\subsection{Overview of the class $\H$}
\label{sec:overview_of_class_h}

\begin{figure}[t]
    \centering
        \includegraphics[keepaspectratio, scale=0.17]{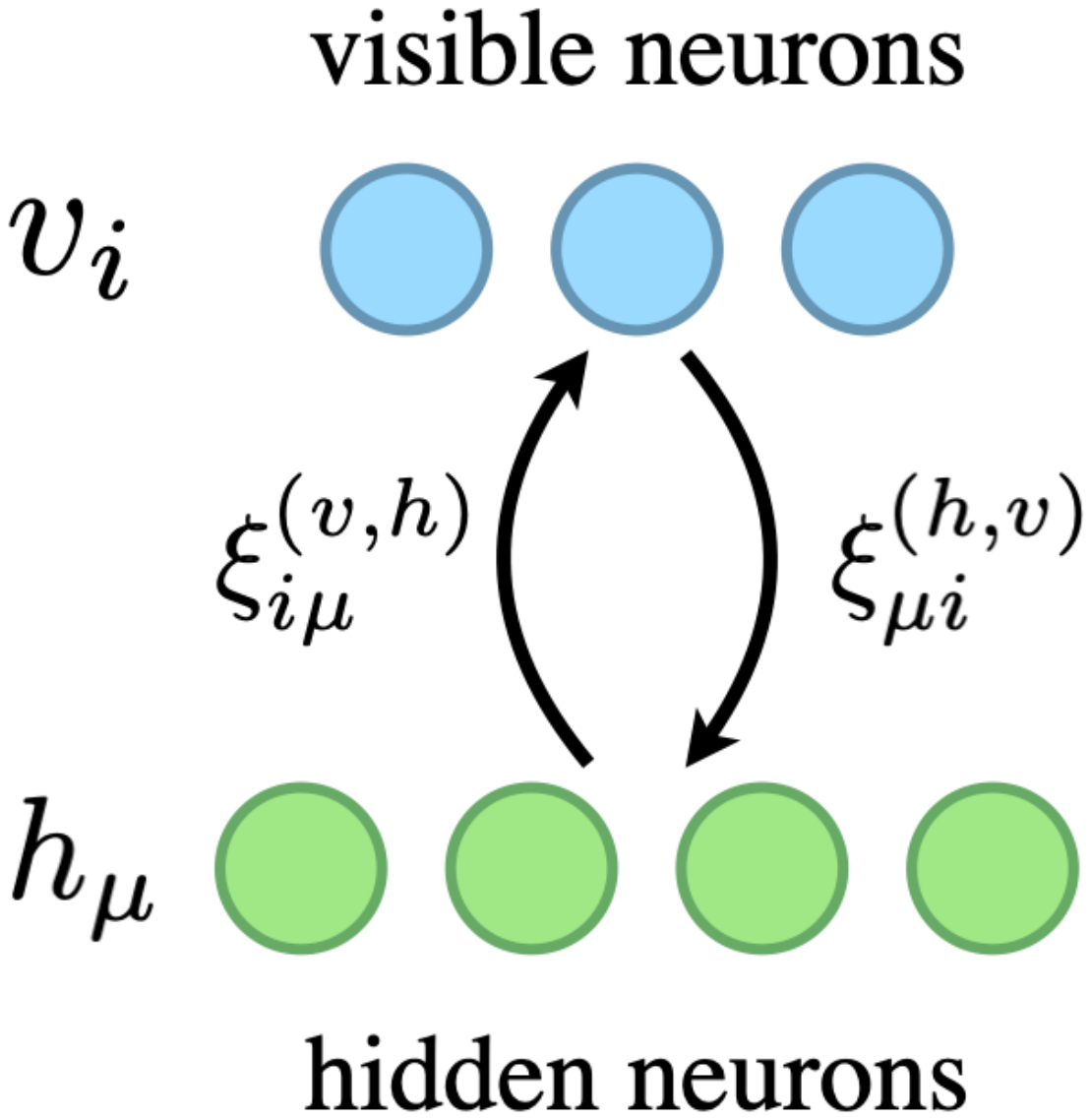}
    \caption{
        The class-$\H$ associative memories.
        The $\Nv$ visible neurons and the $\Nh$ hidden neurons form a bipartite network with no intralayer connections.
        }
    \label{fig:neurons}
\end{figure}

The dynamical variables in this system consist of $\Nv$ visible neurons $\xv(t)\in \R^{\Nv}$ and $\Nh$ hidden neurons $\xh(t)\in \R^{\Nh}$, and their interactions are represented by $\xii{h}{v} \in \R^{\Nh \times \Nv}$ and $\xii{v}{h} \in \R^{\Nv \times \Nh}$, with the constraint $\xii{v}{h} = (\xii{h}{v})^\top$, see \fref{fig:neurons}.
The dynamics of the system is governed by the ``Lagrangians'' $\Lv\colon \R^{\Nv}\to \R$ and $\Lh\colon \R^{\Nh}\to \R$, which determine the activation functions of the neurons as their gradients,
\begin{equation}
    \gh = \nabla \Lh,  \qquad  \gv = \nabla \Lv.
\end{equation}

The dynamical equations of the system and the energy function are given by
\begin{align}
    \tv \dd{\xv(t)}{t}
        = \frac{\coupling}{\th} \xii{v}{h} \gh(\xh(t)) - \xv(t), \\
    \th \dd{\xh(t)}{t}
        = \frac{\coupling}{\tv} \xii{h}{v} \gv(\xv(t)) - \xh(t),
\end{align}
and
\begin{equation}
    \begin{aligned}
        E_\xi(\xv, \xh)  
        &= \frac{1}{\tv} \left( \xv^\top \gv(\xv) - \Lv(\xv) \right)  \\
        &+ \frac{1}{\th} \left( \xh^\top \gh(\xh) - \Lh(\xh) \right)
        - \frac{\coupling}{\th \tv} \gh(\xh)^\top \xii{h}{v} \gv(\xv)  \\
        &\eqcolon \frac{1}{\tv} \Ev(\xv) + \frac{1}{\th} \Eh(\xh) + \frac{\coupling}{\th \tv} \Eint(\xv, \xh),
        \label{eq:energy-h}
    \end{aligned}
\end{equation}
where $\tv$ and $\th$ are the relaxation time constants of the visible and hidden neurons, respectively, and $\coupling$ is a coupling constant.
The combinations $\xv^\top \gv(\xv) - \Lv(\xv)$ and $\xh^\top \gh(\xh) - \Lh(\xh)$ are the Legendre transforms of the Lagrangians evaluated at the activations $\gv(\xv)$ and $\gh(\xh)$.
In this sense, $E_\xi$ is the ``Hamiltonian'' associated with the pair of Lagrangians.
In fact, the dynamical equations above can be viewed as one half of Hamilton's canonical equations restricted to the Legendre constraint surface, which makes the dynamics dissipative, see \appenref{append:class_hk}.

Provided that the Hessians of the Lagrangians are positive (semi-)definite, this energy function monotonically decreases along the solution trajectory of the dynamical equations,
\begin{equation}
    \dd{E_\xi(\xv(t), \xh(t))}{t} \leq 0.
\end{equation}
If, in addition, the overall energy function is bounded from below, the trajectory is guaranteed to converge to a fixed-point attractor state, which corresponds to one of the local minima of the energy function.
Such fixed points may be identified with the stored memories, and the convergence toward them with memory retrieval.

In the adiabatic limit, $\tv \gg \th$, the hidden neurons relax much faster than the visible ones and can be adiabatically eliminated: taking $\th \to 0$ in the dynamical equations, we obtain
\begin{equation}
    \xh(t) = \frac{\coupling}{\tv} \xii{h}{v} \gv(\xv(t)),
\end{equation}
so that the hidden neurons instantaneously follow the visible configuration, and their states $\xh_\mu(t)$ measure the overlap between the activation of the visible neurons and the patterns $\xii{h}{v}_\mu$.
In this sense, each hidden neuron acts as a feature detector for the corresponding pattern, and the hidden neurons serve as the order parameter of memory retrieval in the system \cite{krotov2020large}.

\subsection{Partition function}
\label{sec:partition_function}

To consider the statistical mechanics of the class $\H$ with the energy function \eref{eq:energy-h}, here we introduce a formal partition function in a general setup:
\begin{equation}
    Z_\xi(\beta) = \int d\xv d\xh \exp \left( - \beta E_\xi(\xv, \xh) \right),
    \label{eq:partition-function}
\end{equation}
where $\beta$ is the inverse temperature.  
This expression is formal: for certain choices of Lagrangians, the energy function has flat directions along which the integral diverges (e.g., for $\Lv$ homogeneous of degree one, $\Ev$ vanishes identically and the energy is independent of the overall scale of $\xv$), and thus the precise partition function, including the integration domain and measure, is defined for each model in the corresponding sections below.

We can write this partition function as
\begin{multline}
    Z_\xi(\beta)
        = \int d\xv \, e^{-\frac{\beta}{\tv} \Ev(\xv)}  \\
            \times \int d\xh \exp\left\{-\frac{\beta}{\th} \left(\Eh(\xh) + \frac{\coupling}{\tv} \Eint(\xv, \xh)\right)\right\}.
\end{multline}
In this form, the adiabatic limit is rephrased as $\beta/\th \to \infty$, in which the thermal fluctuations of the hidden neurons are suppressed.
By the saddle-point approximation, the $\xh$ integral then localizes at the stationary point of the integrand, which is given by
\begin{equation}
    \xh_\ast = \frac{\coupling}{\tv} \xii{h}{v} \gv(\xv).
\end{equation}
Thus, in the adiabatic limit, the partition function becomes
\begin{equation}
    \begin{aligned}
        &Z_\xi(\beta)  \\
        &\approx \int d\xv \exp\left\{ -\beta \left( \frac{1}{\tv} \Ev(\xv) - \frac{1}{\th} \Lh(\xh_\ast) \right)\right\} \left( \frac{2\pi}{\beta / \th} \right)^{\Nh/2}  \\
        &= \left( \frac{2\pi}{\beta / \th} \right)^{\Nh/2} \int d\xv \exp\left(-\beta E_\xi(\xv, \xh_\ast)\right).
    \end{aligned}
    \label{eq:partition-function-adiabatic}
\end{equation}
Strictly, the Gaussian prefactor also carries the factor $(\det \hess \Lh(\xh_\ast))^{-1/2}$, which we suppress at this leading order.
Its role is examined for Model B in \appenref{append:model_b_copy_representation}.

As discussed in this section, the hidden neurons play the role of the order parameter of memory retrieval in the system.
In the following sections, we investigate the properties of Models A, B, and C from the viewpoint of the role of the hidden neurons.

%% file: sec3.tex
\section{Models A and C}
\label{sec:models_a_c}

Models A and C are members of a family of models within the class $\H$.
Let us consider a family of Lagrangians,%
\footnote{
    These models can be further generalized, see \appenref{append:family_of_models_a_c}.
    For the purpose of the present paper, it suffices to consider the $\ell_p$-norm.}
\begin{equation}
    \Lv(\xv) = \norm{\xv}_p,  \qquad  \Lh(\xh) = \sum_\mu F(\xh_\mu),
\end{equation}
where $1\leq p\leq \infty$.
The corresponding activation functions are
\begin{align}
    \gv_i(\xv) = \frac{\sgn(\xv_i)\abs{\xv_i}^{p-1}}{\norm{\xv}_p^{p-1}},
    \qquad
    \gh_\mu(\xh) = F^\prime(\xh_\mu).
\end{align}
Since $\xv^\top \gv(\xv) - \Lv = 0$ for these Lagrangians, the bare visible neurons vanish in the energy function, that is, the energy depends on $\xv$ only through the activation $\gv$:
\begin{equation}
    \begin{aligned}
        E_\xi(\xv, \xh)
            &= \frac{1}{\th} \sum_\mu \left( \xh_\mu F^\prime(\xh_\mu) - F(\xh_\mu) \right)  \\
            &\quad - \frac{\coupling}{\th \tv} \sum_{\mu, i} F^\prime(\xh_\mu) \xii{h}{v}_{\mu i} \gv_i(\xv).
    \end{aligned}
\end{equation}

The activation of the visible neurons satisfies $\norm{\gv(\xv)}_{p^\prime}=1$, where $p^\prime$ is the H\"older conjugate of $p$ defined by $1/p + 1/p^\prime = 1$.
Since the energy is independent of the overall scale of $\xv$ (cf.~\sref{sec:partition_function}), the partition function for this family should be defined with the visible integral restricted to a sphere:
\begin{multline}
        Z_\xi(\beta)
            = \int_{\R^{\Nh}} d\xh \, e^{-\frac{\beta}{\th} \sum_\mu \left( \xh_\mu F^\prime(\xh_\mu) - F(\xh_\mu) \right)}  \\
            \times \int_{S} d\Omega(\xx) \exp\left\{ \frac{\beta \coupling}{\th \tv} \sum_{\mu, i} F^\prime(\xh_\mu) \xii{h}{v}_{\mu i} \xx_i \right\},
            \label{eq:partition_function_a_c}
\end{multline}
where
\begin{equation}
    S = \mathbb{S}_{p^\prime}^{\Nv-1}(\Nv^{1/p^\prime}) \coloneq \set{\xx \in \R^{\Nv}  \,\middle|\,  \norm{\xx}_{p^\prime} = \Nv^{1/p^\prime}},
\end{equation}
and $d\Omega$ is the standard measure on this sphere for $1<p<\infty$, while for $p=1, \infty$ the integral reduces to a discrete sum.
The radius of $S$ is chosen such that the components are normalized as $\xx_i = O(1)$: in particular, $S$ is the sphere of radius $\sqrt{\Nv}$ for $p=2$, and the integral reduces to the sum over $\xx \in \set{\pm 1}^{\Nv}$ for $p=1$.

Models A and C correspond to the cases $p=1$ and $p=2$, respectively.
In what follows, we study this family with the hidden Lagrangian specified by $F(x)=x^k/k$ with a positive even integer $k$.%
\footnote{
    The exponent $k$ here is unrelated to the index $k$ of the class $\H_k$.
    Several further notational conflicts occur in what follows, but the meaning should always be clear from the context.
    }
The replica method provides a powerful tool for this purpose.
Since the hidden neurons represent the order parameters of memory retrieval in these systems, we first integrate out the visible neurons and then perform the quenched average over the random patterns.
We focus on the replica symmetric (RS) solutions of these models.

\subsection{Model A}
\label{sec:model_a}

The model-A energy function is given by
\begin{equation}
    \Ea(\xv, \xh) = \Nv \frac{k - 1}{\th k} \sum_\mu \xh_\mu^{k} - \frac{\coupling}{\th \tv} \sum_{\mu, i} \xh_\mu^{k-1} \xii{h}{v}_{\mu i} \sgn(\xv_i),
\end{equation}
where the factor $\Nv$ in the first term is introduced for the extensivity of the energy function.
This modification is equivalent to the renormalizations $\th \to \th/\Nv$ and $\coupling \to \coupling/\Nv$, which fix the correct normalization of $\xh$ and do not affect any other property of Model A as an associative memory.
From \eref{eq:partition_function_a_c}, the partition function for this model reads
\begin{multline}
        \za(\beta)
        = \int d\xh \exp \bigg\{ - \Nv \gamma_k \sum_\mu m_\mu^{k}   \\
        + \sum_{i=1}^{\Nv} \log \bigg( 2\cosh \bigg[ \beta_k \sum_\mu \xii{v}{h}_{i\mu} m_\mu^{k - 1} \bigg] \bigg) \bigg\},
        \label{eq:modelA-partition-function}
\end{multline}
where
\begin{equation}
    \gamma_k = \frac{k - 1}{k} \beta_k,  \quad  \beta_k = \frac{\beta}{\th} \left(\frac{\coupling}{\tv}\right)^k,  \quad  m_\mu = \frac{\xh_\mu}{\coupling/\tv},
\end{equation}
and we dropped an irrelevant overall constant.
Note that the stationarity of \eref{eq:modelA-partition-function} with respect to $m_\mu$ constrains $m_\mu$ to be the thermal average of the overlap between the pattern $\xii{v}{h}_\mu$ and the visible spins $\sgn(\xv_i)$, in accordance with the general role of the hidden neurons discussed in \sref{sec:overview_of_class_h}.

We compute the quenched free energy per visible neuron,
\begin{equation}
    f^{\mathrm{A}}(\beta) = - \lim_{\Nv \to \infty} \frac{1}{\beta \Nv} \EE{\xi}{\log \za(\beta)},
\end{equation}
by the replica method, with the patterns drawn independently as $\xii{h}{v}_{\mu i} = \pm 1$ with equal probability.
We work in the high-load regime of dense associative memories \cite{gardner1987multiconnected,abbott1987storage,baldi1987number},
\begin{equation}
    \Nh = \alpha_k \Nv^{k-1},  
\end{equation}
and consider retrieval states in which a single pattern is condensed, $m_1 = m = O(1)$, while the remaining overlaps are of order $\Nv^{-1/2}$.
For $k>2$, the analysis is performed in the adiabatic limit $\beta/\th \to \infty$ at fixed $\beta_k$, in which the hidden neurons are enslaved to the visible configuration.
The details of the replica computation are presented in \appenref{append:model_a_replica_analysis}.

Under the RS ansatz, we eventually obtain the free energy, up to an additive constant independent of the order parameters,
\begin{multline}
    \beta f^{\mathrm{A}}
        = \gamma_k m^{k} + \frac{\alpha_k \beta_k^2}{2} r (1 - q) + \frac{\alpha_k}{2} \Psi_k(q)  \\
        - \int Dz \log 2\cosh \beta_k \left( m^{k-1} + \sqrt{\alpha_k r}\, z \right),
    \label{eq:modelA-free-energy}
\end{multline}
where $Dz \coloneq dz\, e^{-z^2/2}/\sqrt{2\pi}$ is the standard Gaussian measure, $q$ is the Edwards--Anderson order parameter of the visible spins, $r$ measures the strength of the crosstalk noise from the non-condensed patterns, and the noise entropic term reads
\begin{equation}
    \Psi_k(q) =
        \begin{cases}
            \log \left[ 1 - \beta_k (1 - q) \right] - \dfrac{\beta_k q}{1 - \beta_k (1 - q)},  &k=2,  \\[2mm]
            \beta_k^2 \displaystyle \int_0^q \mathcal{M}_k(s)\, ds,  &k>2,
        \end{cases}
\end{equation}
with $\mathcal{M}_k$ defined in \eref{eq:modelA-Mk} below.
The equations of state for the order parameters follow from the stationarity of the free energy.
We then obtain
\begin{align}
    m &= \int Dz \tanh \beta_k \left( m^{k-1} + \sqrt{\alpha_k r}\, z \right),  \label{eq:modelA-eos-m} \\
    q &= \int Dz \tanh^2 \beta_k \left( m^{k-1} + \sqrt{\alpha_k r}\, z \right),  \label{eq:modelA-eos-q} \\
    r &= \mathcal{M}_k(q),  \label{eq:modelA-eos-r}
\end{align}
where
\begin{equation}
    \mathcal{M}_k(q) =
        \begin{cases}
            \dfrac{q}{\left( 1 - \beta_k (1 - q) \right)^2},  &k=2,  \\[4mm]
            \EE{}{X^{k-1} Y^{k-1}},  &k>2,
        \end{cases}
    \label{eq:modelA-Mk}
\end{equation}
and $(X, Y)$ in the second line denotes a pair of standard Gaussian variables with correlation $\EE{}{XY} = q$.
For $k>2$, $\mathcal{M}_k(q)$ is a polynomial in $q$ with positive coefficients, e.g., $\mathcal{M}_4(q) = 9q + 6q^3$, and $\mathcal{M}_k(1) = (2k-3)!!$.

The two cases in \eref{eq:modelA-Mk} reflect the fate of the Onsager reaction field.
Relative to its bare central-limit scale, the feedback of a non-condensed mode onto itself is of order $\beta_k (1 - q) \Nv^{-(k-2)/2}$.
For $k=2$ this feedback is marginal and must be resummed to all orders, which produces the denominator $1 - \beta_k (1-q)$ of the classical result by Amit, Gutfreund, and Sompolinsky (AGS) \cite{amit1985spin,amit1985storing}.
For $k>2$ it vanishes in the thermodynamic limit, so that the non-condensed overlaps behave as bare Gaussian variables.
A cavity derivation of this dichotomy, together with the closed polynomial form of $\mathcal{M}_k$, is given in \appenref{append:modelA-remarks}.

\ntextbf{Zero-temperature limit and capacity condition.}
Sending $\beta_k \to \infty$, the overlap satisfies $q \to 1$ while the combination $C \coloneq \beta_k (1 - q)$ remains finite, and the equations of state close in the pair $(m, C)$.
The elementary evaluation is given in \appenref{append:modelA-remarks}.
The retrieval overlap obeys
\begin{equation}
    m = \erf \left( \frac{m^{k-1}}{\sqrt{2 \alpha_k r}} \right),
    \label{eq:modelA-zeroT-m}
\end{equation}
the frozen response is
\begin{equation}
    C = \sqrt{\frac{2}{\pi \alpha_k r}}\,
        \exp \left( - \frac{m^{2k-2}}{2 \alpha_k r} \right),
    \label{eq:modelA-zeroT-C}
\end{equation}
and the crosstalk moment \eref{eq:modelA-Mk} becomes
\begin{equation}
    r =
        \begin{cases}
            (1 - C)^{-2},  &k=2,  \\
            (2k-3)!!,  &k>2,
        \end{cases}
    \label{eq:modelA-zeroT-r}
\end{equation}
where the former reproduces the classical AGS zero-temperature equations \cite{amit1985spin,amit1985storing}, and the latter is the $2(k-1)$-th moment of the standard Gaussian.

In terms of the signal-to-noise ratio $t \coloneq m^{k-1}/\sqrt{2 \alpha_k r}$, any retrieval solution with $m>0$ lies on the parametric curve
\begin{equation}
    \alpha_k(t) = \frac{m(t)^{2(k-1)}}{2 t^2\, r(t)},
    \qquad m(t) = \erf(t),
\end{equation}
where $r(t) = (2k-3)!!$ for $k>2$, while for $k=2$ the reaction field enters through $r(t) = [1 - C(t)]^{-2}$ with $C(t) = 2 t e^{-t^2} / [\sqrt{\pi}\, m(t)]$.
Since $\alpha_k(t) \to 0$ both as $t \to 0$ and as $t \to \infty$, the curve attains an interior maximum, and retrieval solutions exist if and only if
\begin{equation}
    \alpha_k \leq \alpha_c \coloneq \max_{t>0} \alpha_k(t),
\end{equation}
where the critical load $\alpha_c$ marks a spinodal at which the stable and unstable retrieval branches merge.
For $k=2$, the maximum occurs at $t_c \simeq 1.51$, yielding $\alpha_c \simeq 0.138$ and $m_c \simeq 0.97$, in agreement with the classical Hopfield result \cite{amit1985spin,amit1985storing}.
For $k>2$, the capacity takes the closed form
\begin{equation}
    \alpha_c = \max_{t>0} \frac{\left[ \erf(t) \right]^{2(k-1)}}{2 t^2\, (2k-3)!!},
    \label{eq:modelA-capacity-closed-form}
\end{equation}
which gives $\alpha_c \simeq 1.32 \times 10^{-2}$ for $k=4$ and $\alpha_c \simeq 1.67 \times 10^{-4}$ for $k=6$.
Note that the roles of the reaction field are opposite in the two cases: for $k=2$, neglecting it ($C \to 0$, $r \to 1$) would overestimate the capacity by more than a factor of four, giving $2/\pi \simeq 0.64$, whereas for $k>2$ the reaction-free expression \eref{eq:modelA-capacity-closed-form} is not an approximation but the leading-order RS result.
The double factorial here is precisely the combinatorial factor that controls the error-free capacity estimate of the dense associative memory \cite{NIPS2016_eaae339c}.
For large $k$, the maximum is attained at $t_c \simeq \sqrt{\ln k}$, so that
\begin{equation}
    \alpha_c \sim \frac{O(1)}{2 t_c^2\, (2k-3)!!}.
\end{equation}
The signal factor $m_c^{2(k-1)}$ remains $O(1)$, and the collapse of the capacity is driven by the factorial growth of the crosstalk variance.

\ntextbf{Critical temperature at $\alpha_k=0$ and steepening of the boundary.}
At $\alpha_k = 0$ the crosstalk noise vanishes and the retrieval overlap satisfies
\begin{equation}
    m = \tanh \left( \beta_k m^{k-1} \right),
\end{equation}
independently of $r$.
The corresponding critical point is determined by the spinodal condition
\begin{equation}
    1 = \beta_k (k-1)\, m^{k-2} \sech^2 \left( \beta_k m^{k-1} \right),
\end{equation}
and the critical value $T_c$ of the effective temperature $T \coloneq \beta_k^{-1}$ decreases with $k$ only slowly (see \tref{tab:modelA-critical-values}).
By contrast, the zero-temperature capacity collapses factorially with $k$, so the retrieval boundary connecting $(\alpha_k, T) = (0, T_c)$ to $(\alpha_c, 0)$ steepens rapidly:
\begin{equation}
    \left| \frac{dT}{d\alpha_k} \right| \sim \frac{T_c}{\alpha_c} \sim (2k-3)!!
\end{equation}
up to factors varying slowly with $k$, and the boundary becomes nearly vertical as $k$ grows.

Physically, the $k$-body interaction sharpens the retrieval landscape through the signal term $m^{k-1}$, while the crosstalk from the non-condensed patterns enters through the $2(k-1)$-th moment of their Gaussian overlaps.
Rare large fluctuations dominate this moment, so the cost of storing one additional pattern grows factorially with $k$.
The model is thus robust against thermal noise at $\alpha_k = 0$ but fragile against pattern loading at $\alpha_k > 0$, an asymmetry that becomes extreme for large $k$.

\begin{table}[t]
    \centering
    \caption{
        Critical values of Model A obtained from the RS equations of state:
        the critical temperature $T_c$ at $\alpha_k = 0$, the zero-temperature critical load $\alpha_c$, and the resulting steepness $T_c/\alpha_c$ of the retrieval boundary.  
        }
    \label{tab:modelA-critical-values}
    \begin{ruledtabular}
    \begin{tabular}{cccc}
        $k$ & $T_c$   & $\alpha_c$            & $T_c / \alpha_c$    \\
        \colrule
        $2$ & $1$     & $0.138$               & $7.2$               \\
        $4$ & $0.496$ & $1.32 \times 10^{-2}$ & $38$                \\
        $6$ & $0.423$ & $1.67 \times 10^{-4}$ & $2.5 \times 10^{3}$ \\
    \end{tabular}
    \end{ruledtabular}
\end{table}

\begin{figure*}[t]
    \centering
    \includegraphics[width=\textwidth]{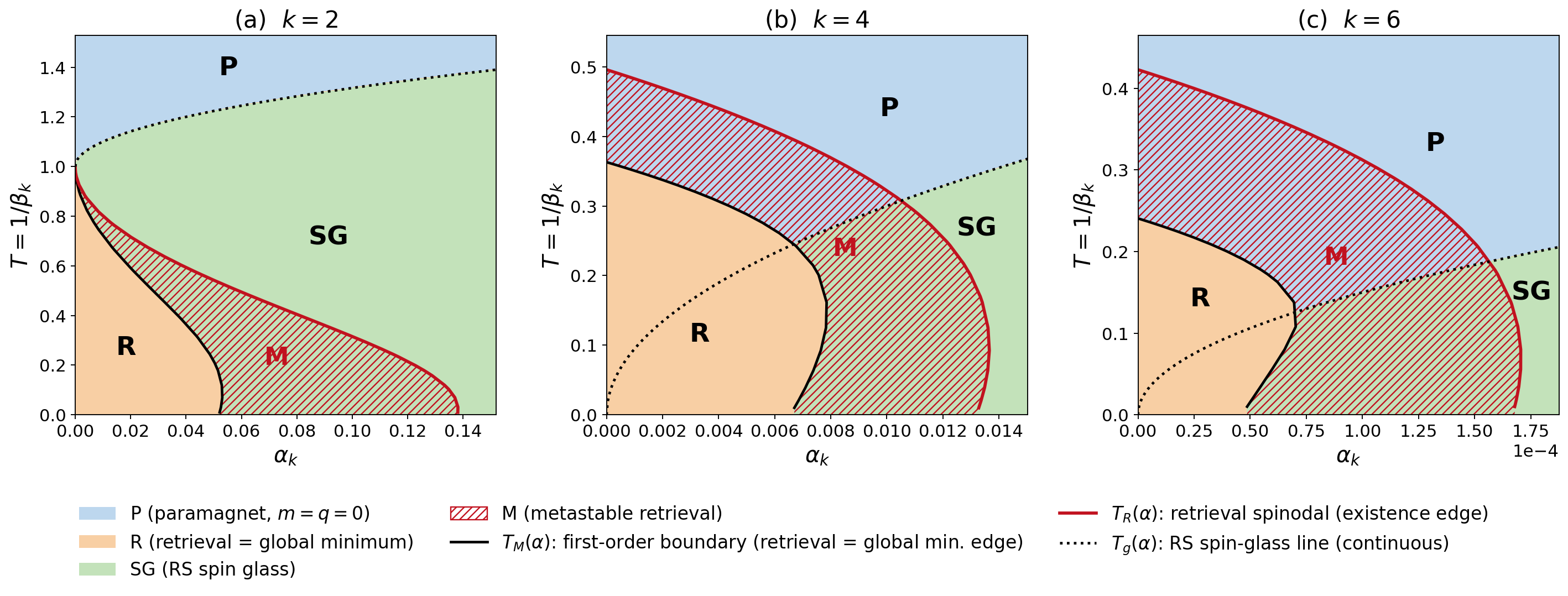}
    \caption{
        RS phase diagram of Model A in the $(\alpha_k, T)$ plane for (a) $k = 2$, (b) $k = 4$, and (c) $k = 6$, where $T = 1/\beta_k$ is the effective temperature.
        In the retrieval phase (R) the retrieval states are the global minima of the free energy.
        In the region M they persist only as metastable states.
        The red solid line is the retrieval spinodal $T_R(\alpha_k)$, beyond which no retrieval solution of \eref{eq:modelA-eos-m}--\eref{eq:modelA-eos-r} exists.
        The black solid line is the first-order boundary $T_M(\alpha_k)$, on which the retrieval free energy crosses that of the $m = 0$ branch.
        The dotted line is the continuous spin-glass transition $T_g(\alpha_k)$ of \eref{eq:modelA-Tg}, separating the paramagnet (P) from the RS spin glass (SG).
        For $k > 2$ the first-order boundary is reentrant, extending to larger loads at intermediate temperatures than at $T = 0$.
        }
    \label{fig:modelA-phase-diagram}
\end{figure*}

\ntextbf{Phase diagram.}
The complete RS phase diagram in the $(\alpha_k, T)$ plane is assembled in \fref{fig:modelA-phase-diagram}.
Besides the retrieval states, the equations of state admit an $m = 0$ solution with $q > 0$, a spin glass sustained by the crosstalk noise alone.
Linearizing \eref{eq:modelA-eos-q} and \eref{eq:modelA-eos-r} at $m = 0$ and small $q$, where $q \simeq \beta_k^2 \alpha_k \mathcal{M}_k(q)$ and $\mathcal{M}_k(q) \simeq [(k-1)!!]^2\, q$ for $k > 2$, shows that this solution bifurcates continuously from the paramagnet at
\begin{equation}
    T_g(\alpha_k) =
        \begin{cases}
            1 + \sqrt{\alpha_k},  &k = 2,  \\[1mm]
            (k-1)!!\, \sqrt{\alpha_k},  &k > 2,
        \end{cases}
    \label{eq:modelA-Tg}
\end{equation}
where the $k = 2$ expression follows from the same expansion with the resummed moment $\mathcal{M}_2$ and reproduces the AGS spin-glass line \cite{amit1987statistical}.
Retrieval solutions exist below the spinodal $T_R(\alpha_k)$, which connects $(0, T_c)$ to $(\alpha_c, 0)$.
They are the global minima of the free energy only below the first-order boundary $T_M(\alpha_k)$, obtained by equating the retrieval free energy \eref{eq:modelA-free-energy} with that of the $m = 0$ branch (the spin glass for $T < T_g$ and the paramagnet above).
Between the two lines retrieval survives as a metastable state.
For $k = 2$ the transition at $\alpha_k = 0$ is continuous, so the metastable band opens only at finite load and closes at $T = 0$ between $\alpha_M \simeq 0.051$ and $\alpha_c \simeq 0.138$ \cite{amit1987statistical}.
For $k > 2$ the transition is first order already at $\alpha_k = 0$, and the band persists down to zero load, where retrieval is metastable against the paramagnet for $T_M(0) < T < T_c$.
The distinct competitor at small load reflects a hierarchy of local stability: a $k$-body coupling exerts no mean field on a disordered configuration (the local field involves a product of $k - 1$ spins), so for $k > 2$ the paramagnet remains locally stable at every temperature, as in the purely ferromagnetic $k$-spin model, and only the accumulated crosstalk of many patterns, with the small-$q$ variance $\alpha_k [(k-1)!!]^2 q$ underlying \eref{eq:modelA-Tg}, can freeze the $m = 0$ sector.
Since $T_g \to 0$ as $\alpha_k \to 0$, at small load retrieval competes directly with the paramagnet on a glass-free background, and the boundaries admit simple estimates: balancing the retrieval energy density $-1/k$ against the paramagnetic entropy $\log 2$, and against the zero-temperature spin-glass energy density $-\sqrt{2 \alpha_k (2k-3)!! / \pi}$, yields
\begin{equation}
    T_M(0) = \frac{1}{k \log 2},
    \qquad
    \alpha_M(0) \simeq \frac{\pi}{2 k^2\, (2k-3)!!},
    \label{eq:modelA-TM-small-load}
\end{equation}
where the former holds up to corrections exponentially small in $k$, and both reproduce the numerical boundaries of \fref{fig:modelA-phase-diagram} within a few percent.
The factorial steepening seen in \tref{tab:modelA-critical-values} is directly visible in the figure: as $k$ grows, the retrieval and metastable regions collapse toward the temperature axis, and the spin-glass phase comes to dominate the diagram.

\ntextbf{Reentrance of the first-order boundary.}
For $k > 2$ the first-order boundary in \fref{fig:modelA-phase-diagram} is visibly reentrant: $\alpha_M(T)$ exceeds its zero-temperature value at intermediate temperatures, by about $20\%$ for $k = 4$ and $54\%$ for $k = 6$.
The mechanism is the entropic asymmetry between the competing states.
In the reentrant window of loads the spin glass has the lower free energy at $T = 0$, but upon heating the $m = 0$ background melts into the paramagnet already at $T_g \propto \sqrt{\alpha_k}$, whereas the retrieval state, protected by the large local field $m^{k-1}$, retains an exponentially small entropy and remains effectively frozen.
Retrieval thereby recovers the global minimum in an intermediate window of temperatures, which closes on the scale $T_M(0)$ of \eref{eq:modelA-TM-small-load}.
Consistently, the reentrance is nearly absent for $k = 2$, where $T_g = 1 + \sqrt{\alpha_k}$ holds the background frozen throughout the retrieval region.
All lines are computed within the RS ansatz.
The spin-glass phase and the low-temperature boundaries acquire corrections from replica symmetry breaking (RSB) \cite{almeida1978stability}, which are known to be small for the retrieval boundaries at $k = 2$ \cite{amit1987statistical} and are expected to remain so for $k > 2$ \cite{gardner1987multiconnected}.
In particular, the far weaker low-temperature reentrance of the spinodal $T_R$ (below four percent for the values of $k$ shown) is the familiar pathology of the RS solution, which RSB removes for $k = 2$ while raising the capacity slightly to $\alpha_c \simeq 0.144$ \cite{crisanti1986saturation}.
The reentrance of $T_M$ instead operates at intermediate temperatures through the melting of the background, although its magnitude may shift under RSB.

\subsection{Model C}
\label{sec:model_c}

Model C is the $p=2$ member of the family, called the spherical memory model in \cite{krotov2020large}: the visible degrees of freedom are continuous variables on the sphere $S = \mathbb{S}^{\Nv-1}(\sqrt{\Nv})$.
The model-C energy function is
\begin{equation}
    \Ec(\xv, \xh) = \Nv \frac{k - 1}{\th k} \sum_\mu \xh_\mu^{k} - \frac{\coupling}{\th \tv} \sum_{\mu, i} \xh_\mu^{k-1} \xii{h}{v}_{\mu i} \xx_i,
\end{equation}
where the factor $\Nv$ in the first term is the same extensivity insertion as in Model A.
The partition function then reads
\begin{multline}
    \zc(\beta)
        = \int d\xh \, e^{- \Nv \gamma_k \sum_\mu m_\mu^{k}} \\
        \times \int_{S} d\Omega(\xx) \exp \left\{ \beta_k \sum_{\mu, i} m_\mu^{k - 1} \xii{h}{v}_{\mu i} \xx_i \right\},
        \label{eq:modelC-partition-function}
\end{multline}
with the same $\gamma_k$, $\beta_k$, and $m_\mu$ as in Model A.
The only change relative to \eref{eq:modelA-partition-function} is that the visible trace runs over the sphere $S$ instead of the hypercube $\set{\pm 1}^{\Nv}$.
In particular, the stationarity with respect to $m_\mu$ again identifies $m_\mu$ with the overlap between the pattern $\xii{h}{v}_\mu$ and the visible configuration $\xx$.

The patterns are now drawn independently and uniformly from the same sphere, $\xii{h}{v}_\mu \sim \mathrm{Unif}(S)$, and we work in the same high-load regime $\Nh = \alpha_k \Nv^{k-1}$ with a single condensed pattern.
By the rotational invariance of the pattern ensemble and of the visible measure, the condensed pattern can be rotated to $\xii{h}{v}_1 = (1, \ldots, 1)$ (the continuous analogue of the gauge choice for binary patterns) exactly at finite $\Nv$, while the remaining patterns are asymptotically Gaussian, with fixed-norm corrections that do not affect the RS equations.
For $k>2$, the analysis is again performed in the adiabatic limit $\beta/\th \to \infty$ at fixed $\beta_k$.
The details of the replica computation are presented in \appenref{append:model_c_replica_analysis}.

In contrast to Model A, the visible trace is a Gaussian integral on the sphere and can be carried out exactly, so no single-site integral survives in the final expressions.
Under the RS ansatz, after eliminating the condensed hidden mode and the Lagrange multiplier of the spherical constraint at their saddle points, we obtain the free energy, up to an additive constant independent of the order parameters,
\begin{equation}
    \begin{aligned}
        \beta f^{\mathrm{C}} =
        &- \frac{\beta_k}{k} m^{k} + \frac{\alpha_k}{2} \Psi_k(q)  \\
        &- \frac{1}{2} \left\{ \log (1 - q) + \frac{q - m^2}{1 - q} \right\},
    \label{eq:modelC-free-energy}
    \end{aligned}
\end{equation}
where $m$ is the overlap between the visible configuration and the condensed pattern, $q$ is the Edwards--Anderson order parameter of the spherical visible state, and $\Psi_k$ is the same noise entropic term as in \eref{eq:modelA-free-energy}.
The equations of state follow from the stationarity of the free energy:
\begin{align}
    0 &= \left( 1 - \beta_k (1 - q)\, m^{k-2} \right) m,  \label{eq:modelC-eos-m} \\
    \frac{q - m^2}{(1 - q)^2} &= \alpha_k \beta_k^2\, \mathcal{M}_k(q),  \label{eq:modelC-eos-q}
\end{align}
where we used $\Psi_k^\prime(q) = \beta_k^2 \mathcal{M}_k(q)$, which holds uniformly in $k$, with the same crosstalk moment $\mathcal{M}_k$ as in \eref{eq:modelA-Mk}.
The crosstalk strength $r = \mathcal{M}_k(q)$ of \eref{eq:modelA-eos-r} thus carries over unchanged.
In the gauge $\xii{h}{v}_1 = (1, \ldots, 1)$, the RS saddle point gives $\ev{\xx_i} = m$ and $\frac{1}{\Nv} \sum_i \ev{\xx_i}^2 = q$.
Since the spherical constraint fixes $\frac{1}{\Nv} \sum_i \ev{\xx_i^2} = 1$, the combination $\beta_k (1 - q)$ is the static susceptibility of the spherical state.
The crosstalk moment is the same as in Model A because the non-condensed overlaps $\Nv^{-1/2} \sum_i \xii{h}{v}_{\mu i} \xx_i$ obey the same central-limit statistics, governed solely by $q$: the crosstalk is universal across the visible ensembles, and the dichotomy of the Onsager reaction field discussed below \eref{eq:modelA-Mk} applies without modification.
The signal equation \eref{eq:modelC-eos-m}, in turn, is of the form familiar from the ferromagnetically biased spherical $p$-spin model \cite{crisanti1992spherical}, the $k$-body signal competing with the entropy of the sphere.

\begin{figure*}[t]
    \centering
    \includegraphics[width=\textwidth]{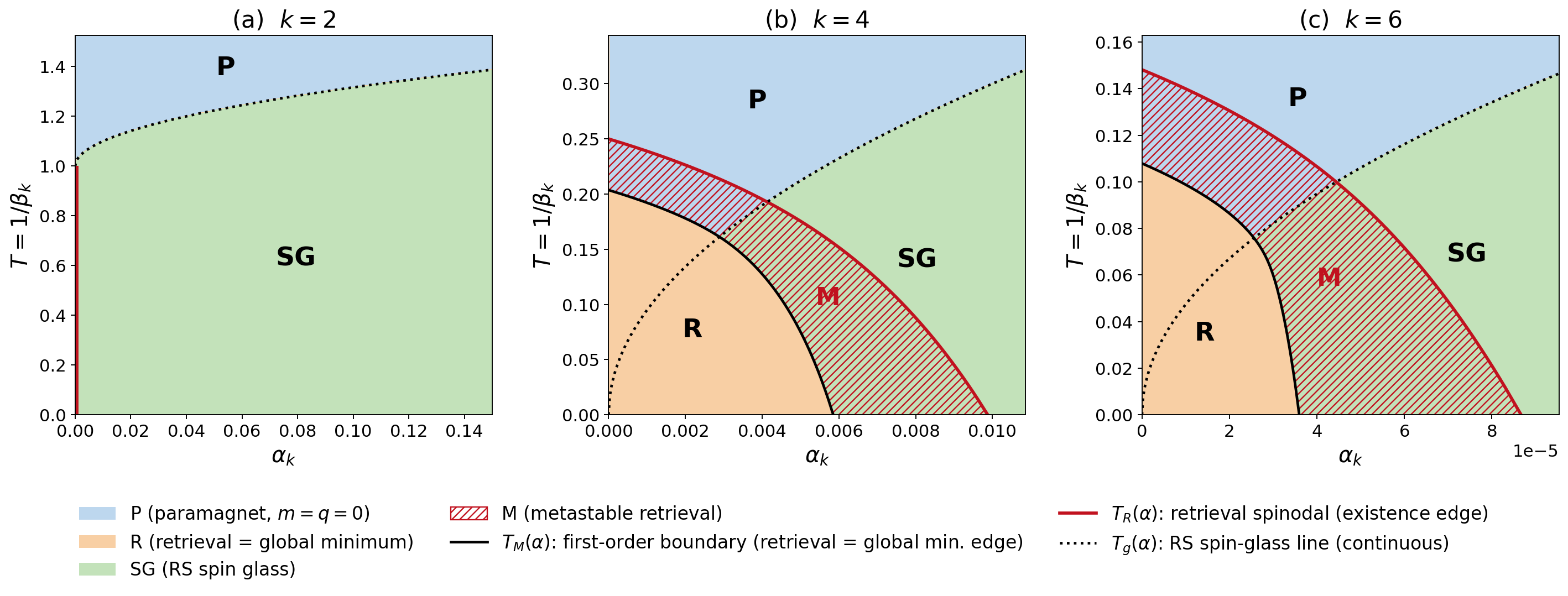}
    \caption{
        RS phase diagram of Model C in the $(\alpha_k, T)$ plane for (a) $k = 2$, (b) $k = 4$, and (c) $k = 6$, in the same conventions as \fref{fig:modelA-phase-diagram}.
        For $k = 2$ retrieval survives only on the segment $\alpha_k = 0$, $T \leq 1$ (red line on the vertical axis), reflecting the zero capacity of the spherical model.
        For $k > 2$ the diagram has the same topology as that of Model A, with uniformly smaller retrieval regions.
        }
    \label{fig:modelC-phase-diagram}
\end{figure*}

\ntextbf{The case $k=2$.}
For $k = 2$ the signal equation degenerates: a retrieval solution $m > 0$ requires $\beta_k (1 - q) = 1$, which is precisely the marginality condition at which the AGS denominator $1 - \beta_k (1 - q)$ of $\mathcal{M}_2$ vanishes.
The right-hand side of \eref{eq:modelC-eos-q} then diverges, so that no RS retrieval solution exists at any $\alpha_k > 0$, including $T = 0$: the quadratic spherical model has zero storage capacity.
This is the known marginality of the spherical Hopfield model \cite{bolle2003spherical}, in which a retrieval phase is recovered only after the energy function is stabilized by an additional quartic term.
At $\alpha_k = 0$, \eref{eq:modelC-eos-q} gives $q = m^2$, and the signal equation yields $m^2 = 1 - T$: retrieval sets in continuously below $T_c = 1$.

\ntextbf{Zero-temperature capacity for $k>2$.}
Sending $\beta_k \to \infty$ with $C = \beta_k (1 - q)$ finite, as in Model A, the signal equation gives $C = m^{2-k}$, while $q \to 1$ and $\mathcal{M}_k(1) = (2k-3)!!$.
The noise equation \eref{eq:modelC-eos-q} then closes algebraically (no error function appears, since the visible integral is Gaussian), and any retrieval solution lies on the curve
\begin{equation}
    \alpha_k(m) = \frac{(1 - m^2)\, m^{2(k-2)}}{(2k-3)!!},
    \label{eq:modelC-capacity-curve}
\end{equation}
whose maximum over $0 < m < 1$ is attained at $m_c^2 = (k-2)/(k-1)$ and yields the capacity in closed form:
\begin{equation}
    \alpha_c = \frac{(k-2)^{k-2}}{(k-1)^{k-1}\, (2k-3)!!},
    \label{eq:modelC-capacity-closed-form}
\end{equation}
again a spinodal at which the stable and unstable retrieval branches merge.
This gives $\alpha_c = 4/405 \simeq 9.88 \times 10^{-3}$ for $k=4$ and $\alpha_c \simeq 8.67 \times 10^{-5}$ for $k=6$.
Note that the retrieval overlap at capacity, $m_c \simeq 0.82$ and $0.89$ respectively, lies visibly below the corresponding values $0.92$ and $0.96$ of Model A: even at zero temperature, the soft spherical spins trade retrieval quality against the crosstalk.

\ntextbf{Critical temperature and retrieval boundary.}
At $\alpha_k = 0$ the equations of state give $q = m^2$ and $1 = \beta_k (1 - m^2)\, m^{k-2}$, whose solvability condition determines, for $k > 2$,
\begin{equation}
    T_c = \frac{2}{k} \left( \frac{k-2}{k} \right)^{(k-2)/2}.
    \label{eq:modelC-Tc-closed-form}
\end{equation}
The transition is again of the spinodal type, with the overlap jumping to $m_c^2 = (k-2)/k$ at $T_c$.
The resulting critical values are collected in \tref{tab:modelC-critical-values}.
As for Model A, the retrieval boundary steepens as $T_c / \alpha_c \sim (2k-3)!!$, driven by the same factorial growth of the crosstalk variance, while every entry is uniformly smaller than its model-A counterpart in \tref{tab:modelA-critical-values}.

\begin{table}[t]
    \centering
    \caption{
        Critical values of Model C obtained from the RS equations of state, in the same conventions as \tref{tab:modelA-critical-values}.
        For $k=2$ retrieval survives only at $\alpha_k = 0$, so that $\alpha_c = 0$ and the steepness is not defined.
        }
    \label{tab:modelC-critical-values}
    \begin{ruledtabular}
    \begin{tabular}{cccc}
        $k$ & $T_c$   & $\alpha_c$            & $T_c / \alpha_c$    \\
        \colrule
        $2$ & $1$     & $0$                   & ---                 \\
        $4$ & $0.25$  & $9.88 \times 10^{-3}$ & $25$                \\
        $6$ & $0.148$ & $8.67 \times 10^{-5}$ & $1.7 \times 10^{3}$ \\
    \end{tabular}
    \end{ruledtabular}
\end{table}

\ntextbf{Phase diagram.}
The resulting RS phase diagram is shown in \fref{fig:modelC-phase-diagram}.
Since the crosstalk moment $\mathcal{M}_k$ is universal, the $m = 0$ sector is governed by the same equations as in Model A, and the continuous spin-glass line $T_g(\alpha_k)$ of \eref{eq:modelA-Tg} carries over unchanged.
The first-order boundary $T_M(\alpha_k)$ again follows by equating the free energy \eref{eq:modelC-free-energy} of the retrieval branch with that of the $m = 0$ branch and, in contrast to Model A, its zero-temperature endpoint is available in closed form: along the capacity curve \eref{eq:modelC-capacity-curve}, the energy balance between the retrieval and spin-glass states reduces to the condition $\sqrt{1 - m^2} = 1/(k-1)$, which yields
\begin{equation}
    \alpha_M(0) = \frac{k^{k-2}\, (k-2)^{k-2}}{(k-1)^{2k-2}\, (2k-3)!!},
    \label{eq:modelC-alphaM-closed-form}
\end{equation}
i.e., $\alpha_M(0) \simeq 5.85 \times 10^{-3}$ for $k = 4$ and $3.60 \times 10^{-5}$ for $k = 6$.
In contrast to Model A, neither boundary of Model C is reentrant: the zero-temperature slope of the spinodal is strictly negative,
\begin{equation}
    \partial_T \alpha_R |_{T=0} \propto (1 - m_c^2)\, \mathcal{M}_k^\prime(1) - \mathcal{M}_k(1) < 0,
\end{equation}
and the first-order boundary is likewise monotone.
The reentrance mechanism of Model A is suppressed by the visible entropy: the spherical retrieval state pays the confinement entropy $-\frac{1}{2} \log (1 - q) \simeq \frac{1}{2} \log \beta_k$ of the condensate, which grows without bound at low temperature, so it melts together with the $m = 0$ background instead of remaining frozen against it.
For $k = 2$ the retrieval phase degenerates to the segment $\alpha_k = 0$, $T \leq 1$, the graphical expression of the marginality discussed above, while the spin-glass line, being common to the two models, is unaffected by the collapse of the retrieval sector.

The comparison between Models A and C, displayed side by side in \fref{fig:modelA-phase-diagram} and \fref{fig:modelC-phase-diagram}, thus isolates the role of the visible degrees of freedom: the crosstalk moment $\mathcal{M}_k$ is universal, so all differences, from the collapse of the $k = 2$ capacity to the absence of the reentrant boundary, originate from the visible entropy.
For $k = 2$, the saturating sign activation of the Ising spins is essential for retrieval: replacing it by the linear spherical activation leaves the retrieval state only marginally confined and destroys the capacity entirely.
For $k > 2$, the $k$-body signal restores a genuine retrieval phase, though with capacity and critical temperature reduced by $O(1)$ factors relative to Model A at each $k$.
The retrieval landscape of this family is therefore shaped jointly by the sharpness of the visible activation and by the order of the hidden nonlinearity, with the latter dominating at large $k$ through the common factorial collapse of \eref{eq:modelA-capacity-closed-form} and \eref{eq:modelC-capacity-closed-form}.

%% file: sec4.tex
\section{Model B}
\label{sec:model_b}

Model B is the attention model of the class $\H$ \cite{krotov2020large}, defined by the Lagrangians
\begin{equation}
    \Lv(\xv) = \frac12 \norm{\xv}^2,  \qquad  \Lh(\xh) = \log \sum_\mu e^{\xh_\mu},
    \label{eq:modelB-lagrangians}
\end{equation}
with the activation functions $\gv(\xv) = \xv$ and $\gh(\xh) = \softmax(\xh)$, where $\softmax(\xh)_\mu \coloneq e^{\xh_\mu} / \sum_\nu e^{\xh_\nu}$.
From \eref{eq:energy-h}, the model-B energy function reads
\begin{equation}
    \begin{aligned}
        \Eb(\xv, \xh)
            &= \frac{1}{2\tv} \norm{\xv}^2 + \frac{1}{\th} \left( \xh^\top \softmax(\xh) - \log \sum_\mu e^{\xh_\mu} \right)  \\
            &\quad - \frac{\coupling}{\th \tv} \softmax(\xh)^\top \xii{h}{v} \xv.
    \end{aligned}
    \label{eq:modelB-energy}
\end{equation}
The hidden Legendre term has an information-theoretic meaning: in terms of the activation $\xf = \softmax(\xh)$ it equals $\sum_\mu \xf_\mu \log \xf_\mu$, the negative Shannon entropy of the probability vector $\xf$.
Since the energy depends on $\xh$ only through $\xf$ (the softmax is invariant under the uniform shift $\xh \to \xh + c\, \onevec$), we may change variables and regard the energy as a function on $\R^{\Nv} \times \simplex^{\Nh}$,
\begin{equation}
    \Eb(\xv, \xf)
        = \frac{\norm{\xv}^2}{2\tv} + \frac{1}{\th} \sum_\mu \xf_\mu \log \xf_\mu - \frac{\coupling}{\th \tv} \xf^\top \xii{h}{v} \xv,
    \label{eq:modelB-energy-f}
\end{equation}
where $\simplex^{\Nh} \coloneq \set{ \xf \in \R^{\Nh} \,\middle|\, \xf_\mu > 0, ~ \sum_\mu \xf_\mu = 1 }$ is the open simplex.
In these variables the hidden sector reduces to the normalized weights $\xf$ that the network assigns to the stored patterns.
These are the attention weights of the modern Hopfield network and of the transformer attention mechanism \cite{NIPS2017_3f5ee243,ramsauer2021hopfield,ota2023attention}, and, in accordance with the general discussion of \sref{sec:overview_of_class_h}, they constitute the order parameter of memory retrieval: retrieval of the pattern $\mu$ corresponds to the concentration of $\xf$ at the vertex $e_\mu$ of the simplex.

Following \sref{sec:partition_function}, the partition function of Model B is defined by%
\footnote{
    The measure $d\xf$ descends from the flat measure $d\xh$ after the zero mode along the uniform shift is factored out.
    This procedure affects only subexponential prefactors and is detailed in \appenref{append:model_b_copy_representation}.}
\begin{equation}
    \zb(\beta)
        = \int_{\R^{\Nv} \times \simplex^{\Nh}} d\xv\, d\xf \exp \left( - \beta \Eb(\xv, \xf) \right).
    \label{eq:modelB-partition-function}
\end{equation}
In the adiabatic limit $\beta/\th \to \infty$, the $\xf$ integral localizes at the stationary point
\begin{equation}
    \xf^\ast = \softmax \left( \frac{\coupling}{\tv} \xii{h}{v} \xv \right),
    \label{eq:modelB-softmax-saddle}
\end{equation}
which is precisely $\gh(\xh_\ast)$ evaluated at the adiabatic saddle point $\xh_\ast$ of \sref{sec:partition_function}, and the partition function reduces to the visible integral \eref{eq:partition-function-adiabatic} with the effective energy
\begin{equation}
    \Eb(\xv, \xh_\ast)
        = \frac{\norm{\xv}^2}{2\tv} - \frac{1}{\th} \log \sum_\mu e^{ \frac{\coupling}{\tv} ( \xii{h}{v} \xv )_\mu }.
    \label{eq:modelB-adiabatic}
\end{equation}
The analysis below depends on the parameters only through the two combinations $\beta/\th$ and $\tilde{\beta} \coloneq (\beta/\tv) (\coupling/\th)^2$.
In parallel with the reduction of the couplings to $\beta_k$ for Models A and C, we fix the normalization $\tv = 1$ and $\th = \coupling$, for which $\tilde{\beta} = \beta$: the effective visible energy in \eref{eq:modelB-adiabatic} becomes $\frac12 \norm{\xv}^2 - \frac{1}{\coupling} \log \sum_\mu e^{\coupling (\xii{h}{v} \xv)_\mu}$, precisely the energy function of the modern Hopfield network \cite{ramsauer2021hopfield} in the form whose exponential storage was analyzed in \cite{lucibello2024exponential}.
In this normalization $\coupling$ controls the sharpness of the attention, while $\beta$ remains the genuine inverse temperature.

As for Models A and C, for Model B the visible sector can be integrated out exactly: the visible Lagrangian is quadratic, so the $\xv$ integral in \eref{eq:modelB-partition-function} is Gaussian.
Carrying it out yields, up to an overall constant,
\begin{align}
    \zb(\beta) &\propto \int_{\simplex^{\Nh}} d\xf\, e^{\Phi(\xf)},  \\
    \Phi(\xf) &= - \frac{\beta}{\coupling} \sum_\mu \xf_\mu \log \xf_\mu + \frac{\beta}{2}\, \xf^\top G \xf,
    \label{eq:modelB-f-representation}
\end{align}
where $G \coloneq \xii{h}{v} \xii{v}{h}$ is the Gram matrix of the patterns, $G_{\mu \nu} = \xii{h}{v}_\mu \cdot \xii{h}{v}_\nu$.
We refer to \eref{eq:modelB-f-representation} as the \defname{$\xf$ representation} of Model B: the entire thermodynamics is expressed by the attention weights alone, as a competition between the attention entropy, which favors delocalized attention, and the positive semi-definite energy $\xf^\top G \xf = \norm{\xii{v}{h} \xf}^2$, which favors concentration.

We draw the patterns independently as Gaussian variables, $\xii{h}{v}_{\mu i} \sim \mathcal{N}(0, 1)$, for which $G_{\mu\mu} = \Nv (1 + O(\Nv^{-1/2}))$ while $G_{\mu\nu} = O(\sqrt{\Nv})$ for $\mu \neq \nu$.
The two terms of $\Phi$ are then of widely different orders.
The energy is extensive: attention concentrated on a single pattern already gains $\frac{\beta}{2} G_{\mu\mu} \approx \frac{\beta}{2} \Nv$.
The entropy, by contrast, is bounded by its value at uniform attention, $\frac{\beta}{\coupling} \log \Nh$: it is of order one per stored pattern rather than per neuron.
A genuine competition between the two therefore requires $\log \Nh \sim \Nv$, and the natural high-load regime of Model B is the exponential load
\begin{equation}
    \Nh = e^{\alpha \Nv},  
    \label{eq:modelB-load}
\end{equation}
in contrast with the polynomial loads $\Nh = \alpha_k \Nv^{k-1}$ of Models A and C.
A second contrast concerns the method: the disorder now enters only through the Gram matrix, whose relevant statistics are large deviations of pattern norms and overlaps, so the analysis below proceeds by direct counting arguments rather than by the replica method.
The relation between the two approaches is clarified in \sref{sec:model_b_copy_representation}.

\subsection{Capacity at zero temperature}
\label{sec:model_b_zero_temperature}

\ntextbf{Vertex condensation.}
The geometry of \eref{eq:modelB-f-representation} dictates the structure of the low-temperature states.
The energy $\xf^\top G \xf = \norm{ \sum_\mu \xf_\mu \xii{h}{v}_\mu }^2$ is a convex function of $\xf$, and a convex function on a compact convex set attains its maximum at an extreme point (Bauer's maximum principle).
On the simplex the extreme points are the vertices.
Explicitly,
\begin{equation}
    \norm{ \sum_\mu \xf_\mu\, \xii{h}{v}_\mu }
        \leq \sum_\mu \xf_\mu \norm{ \xii{h}{v}_\mu }
        \leq \max_\mu \norm{ \xii{h}{v}_\mu },
    \label{eq:modelB-vertex-bound}
\end{equation}
with equality at the vertex of the maximizing pattern.
The entropy opposes this concentration only weakly: spreading the attention uniformly over $M$ patterns reduces the energy gain from $\frac{\beta}{2} \Nv$ to $\frac{\beta}{2} \Nv / M$ at leading order, an extensive loss, whereas the entropy gain is merely $\frac{\beta}{\coupling} \log M$.
For any subexponential load the entropy is thus negligible at leading order, and the Gibbs measure condenses onto the vertices of the simplex, each vertex describing the retrieval state of one pattern with $\xv$ localized near it (cf.~\eref{eq:modelB-softmax-saddle}).
At the exponential load \eref{eq:modelB-load} the competition becomes genuine, but not through interior attention distributions: the measure instead spreads over exponentially many nearly pure vertex states, $\zb \approx \sum_\mu Z_\mu$, and the entropy is the counting entropy of this decomposition.
The capacity question is whether the state condensed on a given typical pattern survives against this exponentially large background.

\ntextbf{Retrieval capacity.}
Fix the retrieved pattern $\xii{h}{v}_1$, of typical norm $\norm{\xii{h}{v}_1}^2 \approx \Nv$.
The stationarity of the effective energy in \eref{eq:modelB-adiabatic} is the fixed-point condition $\xv = \xii{v}{h} \xf^\ast(\xv)$: the visible state is the attention-weighted superposition of the stored patterns.
At zero temperature the retrieval state lies at $\xv \approx \xii{h}{v}_1$, up to corrections controlled by the leak of attention computed below, and each pattern feels the attention field $a_\mu = \coupling\, \xii{h}{v}_\mu \cdot \xv$.
At $\xv = \xii{h}{v}_1$ one has $a_1 = \coupling \norm{\xii{h}{v}_1}^2 \approx \coupling \Nv$, while for $\mu \geq 2$, conditioned on $\xii{h}{v}_1$, the overlaps $\xii{h}{v}_\mu \cdot \xii{h}{v}_1$ are independent centered Gaussians of variance $\norm{\xii{h}{v}_1}^2$, so that $a_\mu = \coupling \sqrt{\Nv}\, \omega_\mu$ with independent standard Gaussian variables $\omega_\mu$.
The condensed attention weight $\mem \coloneq \xf^\ast_1$, the model-B analogue of the retrieval overlap $m$ of Models A and C, then obeys
\begin{equation}
    1 - \mem \leq e^{- a_1} L,
    \qquad
    L \coloneq \sum_{\mu \geq 2} e^{\coupling \sqrt{\Nv}\, \omega_\mu}.
    \label{eq:modelB-leak-sum}
\end{equation}
The sum $L$ is evaluated by the counting argument that underlies extreme value statistics: the number of patterns whose variable $\omega_\mu$ reaches the level $x \sqrt{\Nv}$ is $e^{\Nv (\alpha - x^2/2)}$ to leading exponential order, so the levels up to $x_{\max} = \sqrt{2\alpha}$ are populated while higher levels are empty with high probability.
Retaining the maximal term,
\begin{equation}
    \frac{\log L}{\Nv}
        = \max_{0 \leq x \leq \sqrt{2\alpha}} \big[ \alpha - \tfrac{x^2}{2} + \coupling x \big]
        =
        \begin{cases}
            \alpha + \dfrac{\coupling^2}{2},  & \coupling \leq \sqrt{2\alpha},  \\[2mm]
            \coupling \sqrt{2\alpha},  & \coupling \geq \sqrt{2\alpha}.
        \end{cases}
    \label{eq:modelB-leak-evaluation}
\end{equation}
In the first branch the maximum is attained at the interior point $x^\ast = \coupling$: the leak is carried by exponentially many patterns of moderate overlap, and $L$ is self-averaging and coincides with its annealed average.
In the second branch the maximum is pinned at the boundary $x_{\max}$: the leak is dominated by the $O(1)$ most aligned patterns, and the sum is frozen.
The retrieval state is self-consistent precisely when the leak vanishes, $- \coupling + \Nv^{-1} \log L < 0$.
Combining the two branches yields the zero-temperature capacity
\begin{equation}
    \alpha_c(\coupling) =
        \begin{cases}
            \coupling - \dfrac{\coupling^2}{2},  & \coupling \leq 1,  \\[2mm]
            \dfrac{1}{2},  & \coupling \geq 1,
        \end{cases}
    \label{eq:modelB-capacity}
\end{equation}
the two branches matching continuously at $\coupling = 1$.
Details of these estimates are collected in \appenref{append:model_b_zero_temperature}.

Retrieval fails in physically distinct ways in the two regimes of \eref{eq:modelB-capacity}.
For $\coupling < 1$ the attention is too soft: retrieval is destroyed by the aggregate crosstalk of exponentially many weakly correlated patterns, and sharpening the attention raises the capacity.
For $\coupling > 1$ the crosstalk is dominated by the single most aligned competitor, whose overlap $\max_{\mu \geq 2} \xii{h}{v}_\mu \cdot \xii{h}{v}_1 \approx \sqrt{2\alpha}\, \Nv$ matches the signal $\norm{\xii{h}{v}_1}^2 \approx \Nv$ at $\alpha = 1/2$.
Since $\coupling$ multiplies the signal and the competitor field alike, it cancels from this comparison: no attention sharpness overcomes a competitor as aligned as the signal itself, hence the ceiling.
The capacity \eref{eq:modelB-capacity} is the threshold for retrieving a typical pattern.
Requiring that all $e^{\alpha \Nv}$ patterns be retrievable simultaneously is a stricter demand, met only at lower loads \cite{lucibello2024exponential}.
The value of the ceiling is also specific to the Gaussian ensemble, which enters through the counting rate $x^2/2$: for patterns drawn on the sphere the ceiling disappears and the capacity continues to grow logarithmically in $\coupling$ \cite{lucibello2024exponential}, while for binary patterns exponential capacities were established rigorously in \cite{demircigil2017model}, and the capacity has been computed for more general ensembles, including patterns drawn from a hidden manifold \cite{achilli2025the}.
This sensitivity is characteristic of the exponential regime: at polynomial load the crosstalk is governed by central-limit statistics and is largely insensitive to the pattern ensemble (\sref{sec:models_a_c}), whereas at exponential load it is governed by large deviations, which are not universal.

\ntextbf{Relation to the random energy model.}
The leak sum in \eref{eq:modelB-leak-sum} is precisely the partition function of a random energy model (REM) with $e^{\alpha \Nv}$ independent Gaussian energy levels at inverse temperature $\coupling$ \cite{derrida1981random,mezard2009information}, and \eref{eq:modelB-leak-evaluation} is the standard large-deviation evaluation of its free energy \cite{touchette2009large}.
The two branches are the two phases of the REM: the entropy-dominated phase, in which annealed and quenched averages agree, and the condensed phase below the freezing transition $\coupling = \sqrt{2\alpha}$, in which the measure concentrates on finitely many levels.
This identification makes the agreement with \cite{lucibello2024exponential} structural rather than accidental: there, the zero-temperature energy landscape is split into the signal of the retrieved pattern and a noise term recognized as the free energy of an auxiliary REM, and the resulting typical-pattern capacity $\alpha_1(\coupling)$ coincides exactly with \eref{eq:modelB-capacity}.
The vertex condensation of the Gibbs measure and the landscape analysis are two routes to the same variational problem.

The same extreme value statistics also fixes the status of the retrieval states in the equilibrium ensemble.
At exponential load there exist patterns of atypically large norm, up to $\norm{\xii{h}{v}_{\mu^\ast}}^2 = (1 + \varepsilon_{\max}) \Nv$ with $\varepsilon_{\max} > 0$ determined by the large deviations of the norm, and the state retrieving such a pattern has energy density $-(1 + \varepsilon_{\max})/2$, strictly below the value $-1/2$ of typical retrieval.
Typical retrieval is therefore metastable at any exponential load, and \eref{eq:modelB-capacity} is a spinodal, in the same sense as the capacities of Models A and C.
The equilibrium phase structure built on the condensed extreme patterns is the subject of \sref{sec:model_b_finite_temperature}.

\subsubsection{Copy representation}
\label{sec:model_b_copy_representation}

The $\xf$ representation admits an equivalent discrete formulation, which both fixes its integration measure and clarifies its relation to the replica method.
Consider the adiabatic partition function, the visible integral of the effective energy \eref{eq:modelB-adiabatic}, at the discrete temperatures for which $n \coloneq \beta / \coupling$ is a positive integer.
The logarithm in the exponent then exponentiates into the $n$-th power of $\sum_\mu e^{\coupling\, \xii{h}{v}_\mu \cdot \xv}$, the power expands by the multinomial theorem into a sum over $n$-tuples of pattern indices, and the Gaussian $\xv$ integral gives, up to an overall constant,
\begin{equation}
    \zb(\beta) \propto \sum_{\mu_1, \dots, \mu_n = 1}^{\Nh} \exp \left\{ \frac{\coupling}{2n} \norm{\, \sum_{j=1}^n \xii{h}{v}_{\mu_j} }^2 \right\}.
    \label{eq:modelB-copy-expansion}
\end{equation}
An integer-power representation of this type was introduced in \cite{ota2023attention}.
The hidden-sector fluctuations around the adiabatic saddle point renormalize the exponent as $\beta/\coupling \to \beta/\coupling + \Nh/2$, together with a shift of the summed patterns, and we relegate these corrections to \appenref{append:model_b_copy_representation}.

The expansion \eref{eq:modelB-copy-expansion} describes $n$ ``copies'', each selecting one stored pattern, which interact through the Gram matrix of the selected patterns.
Coarse-graining a copy configuration by its empirical measure $\hat{\xf}_\mu \coloneq n_\mu / n$, with $n_\mu$ the number of copies selecting the pattern $\mu$, the number of configurations in a class is the multinomial coefficient $n! / \prod_\mu n_\mu!$, and Stirling's formula converts \eref{eq:modelB-copy-expansion} into
\begin{equation}
    \zb(\beta) \propto \sum_{\hat{\xf}} \exp \left\{ - \frac{\beta}{\coupling} \sum_\mu \hat{\xf}_\mu \log \hat{\xf}_\mu + \frac{\beta}{2}\, \hat{\xf}^\top G \hat{\xf} \right\}
    \label{eq:modelB-copy-action}
\end{equation}
up to subexponential factors, where the sum runs over the grid $\hat{\xf} \in \simplex^{\Nh} \cap (\Z_{\geq 0} / n)^{\Nh}$.
This is exactly the functional $\Phi$ of \eref{eq:modelB-f-representation}: the copy representation is the $\xf$ representation with its measure made explicit as a counting measure.
This distinction is not innocuous at exponential load, where the simplex has dimension $e^{\alpha \Nv} - 1$ and the volume factors of a continuum measure are a priori uncontrolled.
The counting measure supplied by the model itself carries no such factors, its only extensive entropy being the choice of the selected patterns.

The physical picture behind \eref{eq:modelB-copy-expansion} is transparent.
Reversing the Gaussian integration shows that, given a copy configuration, the visible state is Gaussian with mean the centroid $\frac{1}{n} \sum_j \xii{h}{v}_{\mu_j}$ and variance $1/\beta$ per component: the $n$ copies are parallel retrieval queries sharing the single context $\xv$, and the temperature enters only through the number of queries, $n = \beta/\coupling$.
Retrieval is the fully aligned configuration in which all copies select the same pattern, and the elementary excitation above it is a single copy defecting to a competitor, which carries the attention quantum $1/n = \coupling T$.
As $T \to 0$ the number of copies diverges, the aligned configurations reproduce the vertex condensation described above, and the stability of alignment against single-copy defection reproduces the capacity \eref{eq:modelB-capacity} (see \appenref{append:model_b_finite_temperature_copy}).

The copy representation also locates the present analysis relative to the replica method: in the replica approach the patterns are averaged first, at a formal number of replicas continued to zero, whereas here the thermal variable $\xv$ is integrated out first, at a physical, integer number of copies, so the thermal and quenched averages exchange roles.
The effective $\log\det$ interaction that the pattern average generates among replicas reappears here as the counting entropy of the pattern choices, its Legendre transform (\appenref{append:model_b_copy_representation}).
Likewise, the symmetric structure of the order parameter, an ansatz in the replica computation, arises here as the exact maximizer within each class of copy configurations, and the frozen branch of \eref{eq:modelB-leak-evaluation} plays the role that one-step RSB (1RSB) plays in the REM.
The construction realizes the clone method of Monasson \cite{monasson1995structural}, with the number of clones set by the physical temperature rather than introduced as an auxiliary parameter.

\subsection{Phase diagram at finite temperature}
\label{sec:model_b_finite_temperature}

The copy representation reduces the finite-temperature problem to a combinatorial one.
The temperature enters only through the copy number $n = \beta / \coupling$, and a configuration is specified by the partition of the $n$ copies into groups occupying distinct patterns.
Since the weight in \eref{eq:modelB-copy-expansion} depends on the occupied patterns only through their Gram matrix, the disorder average reduces to a counting problem: the number of choices of $M$ distinct patterns with prescribed normalized Gram matrix $Q_{jl} = G_{\mu_j \mu_l} / \Nv$ is $e^{\Nv ( M \alpha - I_M(Q) )}$ to leading exponential order, where
\begin{equation}
    I_M(Q) = \frac{1}{2} \left[ \Tr Q - \log \det Q - M \right]
    \label{eq:modelB-gram-rate}
\end{equation}
is the large-deviation rate function of the Gram matrix, the relative entropy between the centered Gaussian ensembles of covariance $Q$ and of unit covariance \cite{touchette2009large}.
For $M \alpha < I_M(Q)$ no such choice exists with high probability.
Maximizing the counting factor times the weight under this existence constraint extends the two branches of \eref{eq:modelB-leak-evaluation} to every configuration class: an annealed branch, attained at the typical Gram matrix of an exponentially tilted ensemble, and a frozen branch, pinned to the most extreme configurations actually present.
The optimization over the partition classes is elementary and is carried out in \appenref{append:model_b_finite_temperature_copy}.
Here we summarize the results.

\begin{figure*}[t]
    \centering
    \includegraphics[width=\textwidth]{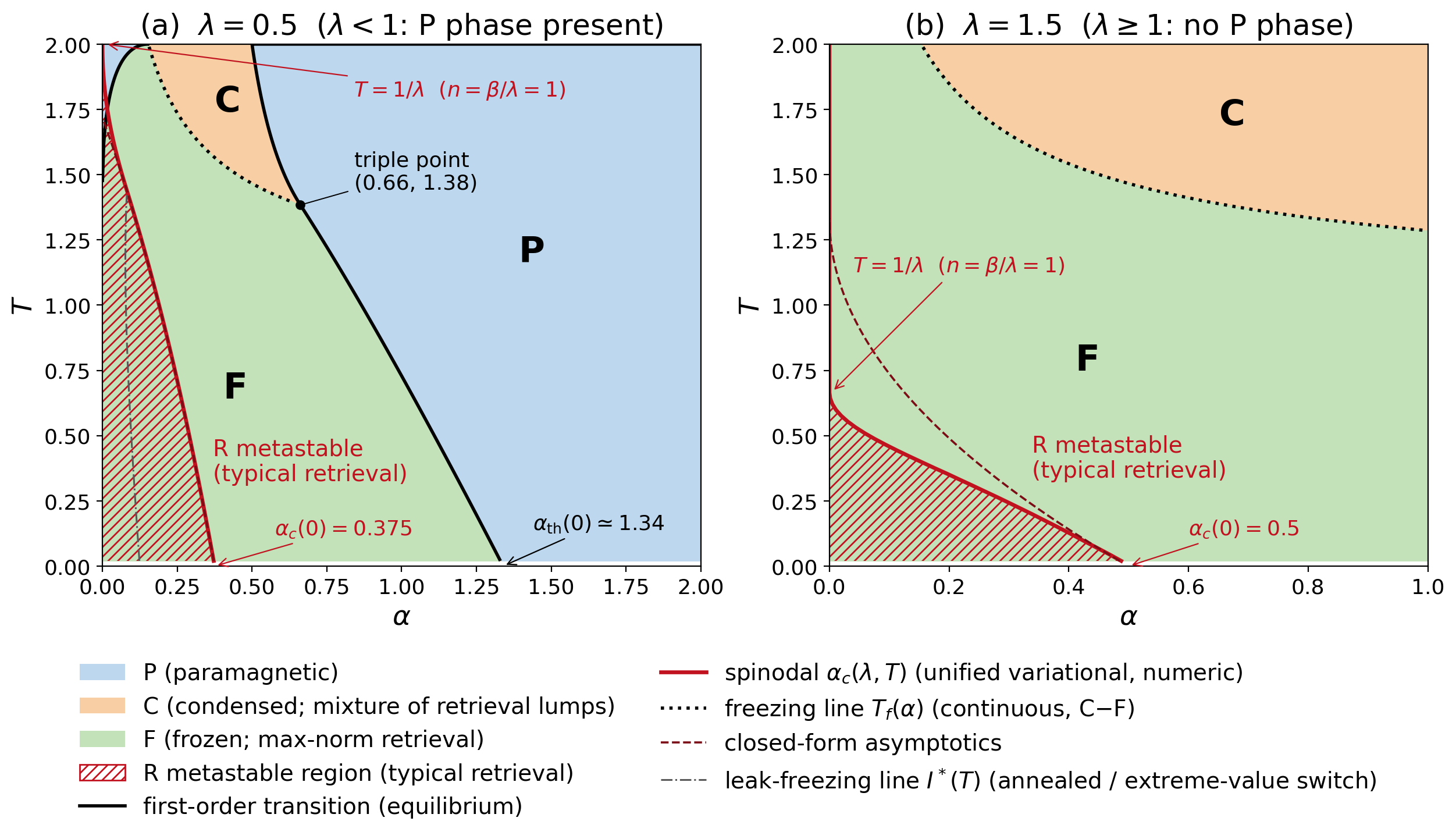}
    \caption{
        Finite-temperature phase diagram of Model B in the $(\alpha, T)$ plane at (a) $\coupling = 0.5$ and (b) $\coupling = 1.5$.
        Black solid lines are the first-order boundaries between the paramagnetic phase (P) and the condensed sector, obtained by equating the free energies \eref{eq:modelB-branch-free-energies}.
        For $\coupling \geq 1$ the paramagnetic phase is absent, and in (a) the P--F boundary meets $T = 0$ at $\alpha_{\mathrm{th}}(0)$ determined by $2 \alpha / \coupling = 1 + \varepsilon_{\max}(\alpha)$, see \appenref{append:model_b_zero_temperature}.
        The dotted line is the freezing line \eref{eq:modelB-freezing-line}, the continuous transition on which the entropy of the condensed phase (C) vanishes and the measure freezes onto the maximum-norm patterns (F).
        It is independent of $\coupling$ and hence identical in the two panels.
        The thick red line is the retrieval spinodal, evaluated numerically from the one-quantum criterion of \appenref{append:complete_phase_diagram}, whose annealed and frozen branches are \eref{eq:modelB-spinodal} and \eref{eq:modelB-ceiling}.
        The dashed line shows its closed-form asymptotics, and the dash-dotted line is the switch line $I^\ast(T)$ between the annealed and extreme-value evaluations of the defection destination.
        The hatched region is the metastable retrieval region, which terminates at $T = 1/\coupling$, where the attention quantum $\coupling T$ reaches the full attention weight ($n = 1$).
        }
    \label{fig:modelB-phase-diagram}
\end{figure*}

\ntextbf{Equilibrium phases.}
Only the two extreme partitions survive the optimization, all copies on separate patterns or all copies on a single pattern.
Measuring the free energy per visible neuron relative to the Gaussian reference $\int d\xv\, e^{-\beta \norm{\xv}^2 / 2}$, the resulting branches are
\begin{equation}
    \begin{aligned}
        f_{\mathrm{P}} &= - \frac{\alpha}{\coupling} + \frac{T}{2} \log (1 - \coupling),  \\
        f_{\mathrm{C}} &= - T \alpha + \frac{T}{2} \log (1 - \beta),  \\
        f_{\mathrm{F}} &= - \frac{1 + \varepsilon_{\max}(\alpha)}{2},
    \end{aligned}
    \label{eq:modelB-branch-free-energies}
\end{equation}
and the equilibrium phase at given $(\alpha, \coupling, T)$ is the branch of lowest free energy.
In the paramagnetic phase (P) each copy selects its own pattern and contributes the selection entropy $\alpha$.
The visible state is a weak thermal condensate, $\ev{\norm{\xv}^2} / \Nv = T / (1 - \coupling)$, which vanishes as $T \to 0$.
The selection entropy itself, however, does not vanish: it enters in proportion to the copy number $n = \beta / \coupling$ and leaves the temperature-independent term $- \alpha / \coupling$ in $f_{\mathrm{P}}$, so that the paramagnet survives down to $T = 0$ and remains the equilibrium phase at loads beyond the zero-temperature intercept $\alpha_{\mathrm{th}}(0)$ of the P--F boundary in \fref{fig:modelB-phase-diagram}(a): at exponential load, the entropy of pattern selection acts as an energy.
In the condensed phase (C) all $n$ copies align on a single pattern, thermally tilted toward the norm $(1 - \beta)^{-1} \Nv$, and the measure is an ergodic mixture over the exponentially many retrieval lumps of \sref{sec:model_b_zero_temperature}, with entropy density $\alpha - \kappa(\beta)$, where $\kappa(\beta) \coloneq I_1((1 - \beta)^{-1}) = \frac{1}{2} [ \beta / (1 - \beta) + \log (1 - \beta) ]$ is the counting cost of the tilted norm.
This entropy vanishes on the freezing line
\begin{equation}
    T_f(\alpha) = 1 + \frac{1}{\varepsilon_{\max}(\alpha)},
    \label{eq:modelB-freezing-line}
\end{equation}
where $\varepsilon_{\max}(\alpha)$, defined by $I_1(1 + \varepsilon_{\max}) = \alpha$, makes precise the maximal norm excess introduced in \sref{sec:model_b_zero_temperature}.
Below $T_f$ the mixture freezes onto the $O(1)$ patterns of maximal norm: this frozen phase (F) is the retrieval state of the maximum-norm pattern, its free energy is temperature independent, and the transition at $T_f$ is the continuous freezing transition of the REM \cite{derrida1981random,mezard2009information}.
The existence conditions of the two annealed branches, $\coupling < 1$ for P and $T > 1$ for C, are two instances of a single criterion: a state with attention spread over $n / s$ patterns has inverse participation ratio $\norm{\xf}^2 = s / n$, and its visible Gaussian fluctuations are stable only while $\beta \norm{\xf}^2 < 1$, the instability at $\beta \norm{\xf}^2 = 1$ being the divergence of the visible integral in the $\xf$ representation \eref{eq:modelB-f-representation}.
In particular, for $\coupling \geq 1$ the paramagnetic phase is absent altogether.
The first-order boundaries between P and the condensed sector follow by equating the branches of \eref{eq:modelB-branch-free-energies}, and the resulting phase diagram is shown in \fref{fig:modelB-phase-diagram}.
For $\coupling < 1$ the paramagnetic and the two condensed phases meet at a triple point, located at $(\alpha, T) \simeq (0.66, 1.38)$ for $\coupling = 0.5$.

\ntextbf{Retrieval branch.}
Typical retrieval appears in this landscape as a constrained branch.
Fixing the attention weight $\mem = \hat{\xf}_1$ condensed on a typical pattern as a reaction coordinate, and optimizing over the destinations and the Gram matrix of the $n (1 - \mem)$ remaining copies, yields the Landau function
\begin{equation}
    f_{\mathrm{R}}(\mem)
        = - \frac{\alpha}{\coupling} (1 - \mem)
        + \frac{T}{2} \log A
        - \frac{\mem^2}{2 A},
    \label{eq:modelB-landau-function}
\end{equation}
where $A \coloneq 1 - \coupling (1 - \mem)$.
The three terms are the selection entropy of the leaked attention, the determinant of the visible fluctuations softened by the leak, and the signal energy.
The leaked copies respond linearly to the retrieval field and align weakly with the retrieved pattern, which amplifies the visible overlap to $m = \mem / A \geq \mem$.
The branch interpolates between the paramagnet, $f_{\mathrm{R}}(0) = f_{\mathrm{P}}$, and pure retrieval, $f_{\mathrm{R}}(1) = - \frac12$, the zero-temperature retrieval energy density of \sref{sec:model_b_zero_temperature}.
Its interior stationary point is a maximum: a barrier between the two endpoints, not a phase.
Interior attention distributions therefore never become states at any temperature, which is the finite-temperature form of vertex condensation.
The construction is a Franz--Parisi potential \cite{franz1995recipes} whose reference configuration is the stored pattern itself, and the retrieval state at $\mem = 1$ is a metastable state in the restricted-ensemble sense \cite{penrose1971rigorous}.

\ntextbf{Finite-temperature capacity.}
The stability of the retrieval endpoint is not governed by the slope of \eref{eq:modelB-landau-function} at $\mem = 1$.
The attention weight moves on the lattice $\mem = 1 - k \coupling T$ with $k \in \Z_{\geq 0}$: the elementary excitation is the defection of a single copy, which carries the attention quantum $\coupling T$, and retrieval survives as long as a single defection raises the free energy.
Optimizing the destination of the defecting copy over the patterns present in the ensemble yields the finite-temperature spinodal
\begin{equation}
    \alpha_c(\coupling, T)
        = \frac{\coupling (2 - \coupling - \coupling T)}{2 (1 - \coupling^2 T)}
        + \frac{1}{2} \log \left( 1 - \coupling^2 T \right)
    \label{eq:modelB-spinodal}
\end{equation}
in the annealed regime, in which destinations of the optimal overlap and norm exist in exponential number.
For sharp attention, $\coupling^2 \gtrsim 2 \alpha$, they do not, and the defection freezes onto the single most aligned pattern present, as in the frozen branch of \eref{eq:modelB-leak-evaluation}.
Two effects of order $T$ then favor the defection: the condensate left behind is thinner by one copy, and an atypically large norm of the destination pattern becomes an energy gain in its own right.
The zero-temperature ceiling accordingly bends downward,
\begin{equation}
    \alpha_{\mathrm{ceil}}(\coupling, T)
        = \frac{1}{2} \left( 1 - \frac{\coupling T}{2} \right)^{2} + O(T^2),
    \label{eq:modelB-ceiling}
\end{equation}
and the finite-temperature capacity follows \eref{eq:modelB-spinodal} in the annealed regime and \eref{eq:modelB-ceiling} in the frozen regime, with the branch switch at $\coupling^2 \approx 2 \alpha$.
As $T \to 0$ the two branches reduce to the zero-temperature capacity \eref{eq:modelB-capacity}.
In the opposite direction, the metastable region terminates at $T = 1/\coupling$, where the attention quantum $\coupling T$ reaches the full attention weight and a single defection already empties the condensate (see \fref{fig:modelB-phase-diagram}).
The approach to this endpoint is controlled by the frozen branch: at small $\alpha$ destinations of the optimal overlap cease to exist, so the annealed expression \eref{eq:modelB-spinodal}, whose zero lies below $1/\coupling$, underestimates the stability of retrieval, and the defection onto the most favorable pattern actually present sustains a narrow metastable strip up to $T = 1/\coupling$.
A continuum treatment of $\mem$ would overestimate the capacity at any $T > 0$: it resolves barriers thinner than one attention quantum and counts the selection entropy per infinitesimal attention weight rather than per copy.
The thermal destabilization of retrieval thus proceeds by discrete reassignments of attention, not by a smooth erosion of the overlap.

The results above are derived at the integer temperature points $n \in \Z_{>0}$, but they are not tied to this lattice: replacing the multinomial expansion of \sref{sec:model_b_copy_representation} by the generalized binomial expansion of the same integrand extends the sector decomposition, the attention quantum $\coupling T$, and all the formulas above unchanged to arbitrary real $\beta$.
The complete real-temperature construction of the phase diagram is presented in \appenref{append:complete_phase_diagram}.

The phase diagram fixes the status of retrieval at exponential load.
Since $\varepsilon_{\max} > 0$ at any $\alpha > 0$, the frozen phase lies below pure retrieval, $f_{\mathrm{F}} < f_{\mathrm{R}}(1) = - \frac12$, at all temperatures: for Gaussian patterns, typical retrieval never becomes the equilibrium phase, and the associative memory operates throughout the hatched region of \fref{fig:modelB-phase-diagram} as a metastable state, with an escape time exponentially large in $\Nv$, governed in the Arrhenius sense by the one-quantum barrier of \appenref{append:complete_phase_diagram}, mapped over the metastable region in \fref{fig:appB-phase-diagram-full}.
This sharpens the zero-temperature statement of \sref{sec:model_b_zero_temperature} and contrasts with Models A and C in two respects.
First, the mechanism of thermal destabilization is different: there the retrieval overlap erodes smoothly through the equations of state, while here the temperature acts solely through the number of copies, and retrieval is undone by quantized reassignments of attention.
Second, the scale is different: the retrieval boundaries of Models A and C steepen factorially with $k$ (\tref{tab:modelA-critical-values} and \tref{tab:modelC-critical-values}), whereas the model-B retrieval region retains an extent of order unity in both $\alpha$ and $T$.
Finally, the equilibrium dominance of the frozen phase is a norm-fluctuation effect of the Gaussian ensemble, driven by the same large deviations that set the capacity ceiling: for patterns drawn on the sphere, $\varepsilon_{\max} \equiv 0$, the frozen phase loses its advantage, and typical retrieval can compete as a genuine equilibrium phase.
The ensemble sensitivity noted below \eref{eq:modelB-capacity} thus extends from the capacity to the entire equilibrium structure.

%% file: sec6.tex
\section{Discussion}
\label{sec:discussion}

We have investigated the statistical mechanics of the class-$\H$ associative memories \cite{krotov2020large} through the hidden neurons, the order parameter of memory retrieval.
For Models A and C (\sref{sec:models_a_c}), the replica method at polynomial load yields the RS phase diagrams and closed-form capacities.
Since the crosstalk moment is universal, the model dependence, foremost the zero capacity of the quadratic spherical model, rests on the visible entropy, while the factorial growth of the crosstalk variance collapses both capacities at large $k$.
For Model B (\sref{sec:model_b}), the hidden sector reduces to the attention weights, the load is exponential, and the copy representation maps the thermodynamics onto REM counting, with paramagnetic, condensed, and frozen phases and typical retrieval metastable, undone thermally by quantized reassignments of attention.
The two regimes differ in their crosstalk statistics, central-limit and largely insensitive to the pattern ensemble at polynomial load, large-deviation and ensemble-dependent at exponential load.
Retrieval thus separates into the two roles carried by the Lagrangians of the class $\H$: the visible Lagrangian fixes, through the visible entropy, the stability of retrieval under a given crosstalk, while the hidden Lagrangian fixes the storage scale and with it the character of the crosstalk statistics.
The models analyzed here are positions on these two axes rather than separate theories.

Two limitations deserve mention.
First, all results for Models A and C are derived within the RS ansatz, the capacities being the spinodals of the RS free energy.
The de~Almeida--Thouless (AT) stability of the RS saddle points \cite{almeida1978stability} has not been examined, and the spin-glass phases and the low-temperature boundaries acquire corrections from RSB, known to be small for the retrieval boundaries at $k = 2$ \cite{amit1987statistical,crisanti1986saturation} and expected to remain so at $k > 2$ \cite{gardner1987multiconnected}.
Second, for $k > 2$ and for Model B the hidden sector is treated in the adiabatic limit $\beta / \th \to \infty$, with the hidden neurons enslaved to the visible configuration.
At finite $\beta / \th$ the class $\H$ is a genuinely two-temperature system, whose phase diagrams away from the adiabatic limit we do not address.
The equal-temperature end of this interpolation is the bipartite Gibbs measure of a restricted Boltzmann machine, whose equivalence with the Hopfield model \cite{barra2012equivalence} and whose phase diagram for general hidden priors \cite{barra2018phase} are known.

These limitations mark the first direction for future work, completing the phase diagrams under RSB.
For Models A and C this amounts to locating the AT lines and computing the one-step corrections to the spin-glass phase and the retrieval boundaries, for which the techniques developed for the Hopfield model \cite{amit1987statistical,crisanti1986saturation}, dense networks \cite{albanese2022replica,albanese2024replica}, and bipartite spin glasses \cite{hartnett2018replica} apply directly.
For Model B the corresponding step is the stability of the sector decomposition against fluctuations between sectors, together with a test of the predicted Ruelle statistics of the attention weights in the frozen phase \cite{ruelle1987mathematical} (\appenref{append:complete_phase_diagram}).

A second direction concerns the hidden-sector corrections of Model B.
The Gaussian fluctuations of the hidden neurons around the adiabatic saddle point produce the exponent shift $\beta / \coupling \to \beta / \coupling + \Nh / 2$ and an additive shift of the summed patterns (\appenref{append:model_b_copy_representation}), neither of which is small at exponential load.
Whether they can be absorbed into a redefinition of the reference measure, leaving the rate-level phase diagram intact, is the main structural question left open by our analysis (\appenref{append:complete_phase_diagram}).
Since both originate from the zero mode of the softmax, their resolution would clarify how the normalization of the attention weights, shared by the transformer attention and the modern Hopfield network \cite{NIPS2017_3f5ee243,ramsauer2021hopfield,ota2023attention}, affects the thermodynamics beyond the leading order.

The third direction is more open-ended.
Regarded as design variables rather than properties of given models, the two axes suggest constructing new Lagrangians, or selecting them with prescribed retrieval properties, in the spirit of recent programs on the design of energy-based architectures \cite{krotov2023new,hoover2024energy,krotov2025modern}.
The family generating Models A and C (\appenref{append:family_of_models_a_c}) is the smallest setting in which both axes can be varied, and the same question extends to the class $\H_k$ (\appenref{append:class_hk}), where additional layers of hidden neurons implement the hierarchical associative memory \cite{krotov2021hierarchical,karakida2024hierarchical}.
Identifying the order parameters of the deeper hidden layers, and the phases they support, would extend the present analysis to genuinely hierarchical models, for which exponential capacities from distributed hidden representations \cite{kafraj2026biologically} and emergent computations in assemblies of networks \cite{agliari2026networks} have recently been reported.

%% file: appendix.tex
\section{Overview of the class-$\H_k$ associative memories}
\label{append:class_hk}

The class-$\H_k$ associative memories are a slight modification of Krotov's hierarchical associative memory \cite{krotov2021hierarchical},%
\footnote{
    $\H_k$ stands for ``\textbf{H}ierarchical Associative Memory with $\bm{k}$ layers'' (HAM$_{k}$, cf.~\cite{hoover2022universal}), namely the class of associative memories with $k$ defining Lagrangians.
    One can also think of it as an extended \textbf{K}rotov--\textbf{H}opfield model \cite{krotov2020large}, which in this paper we call the class $\H$, a special case of $\H_k$ with $k=2$.}
the modification being that the energy function carries explicit coupling constants $\coupling_{A+1, A}$ and layer-wise weights $1/\tau_A$.
The class $\H_k$ consists of $k ~ (\geq 2)$ layers, with $N_A$ neurons in layer $A$ ($A=1,\dots,k$).
This system is a generalization of the model discussed in \cite{krotov2020large} and is specified by the following components:
the neuron states $x^A(t)\in \R^{N_A}$ in each layer, interaction matrices between the adjacent layers, $\xii{A+1}{A}\in \R^{N_{A+1}\times N_{A}}$, and activation functions
\begin{equation}
    g^A: \R^{N_A} \to \R^{N_A},  \quad  A = 1, \dots, k,
\end{equation}
determined through Lagrangians $L^A: \R^{N_A}\to \R$ such that $g^A = \nabla L^A$.
Note that the interaction matrices are required to be ``symmetric'': $\xii{A}{A+1} \coloneq (\xii{A+1}{A})^\top$.
(For more details, see \cite[Sec.~3]{krotov2021hierarchical}.)

Let us now define an energy function on $T^\ast (\R^{\sum_{A=1}^k N_A}) \simeq \prod_{A=1}^k (\R^{N_A} \times \R^{N_A})$ by
\begin{equation}
    E(\set{y^A}, \set{x^A})
        = \sum_{A=1}^k \frac{1}{\tau_A} \left( (x^A)^\top y^A - L^A(x^A) \right)
        - \sum_{A=1}^{k-1} \frac{\coupling_{A+1, A}}{\tau_{A+1} \tau_{A}} (y^{A+1})^\top \xii{A+1}{A} y^A,
\end{equation}
where $\coupling_{A+1, A}$ are coupling constants between the $(A+1)$- and $A$-th layers and are assumed to be symmetric, $\coupling_{A+1, A} = \coupling_{A, A+1}$.
This energy function may be regarded as a Hamiltonian function on the phase space, with the variables $y^A$ playing the role of the positions and the neuron states $x^A$ that of the conjugate momenta.
The first sum in $E$ is then the Legendre transform of the Lagrangians.
The physical configurations lie on the Legendre constraint surface $\Sigma \coloneq \set{y^A = g^A(x^A)}_{A=1}^k$, which is precisely the locus where the gradients $\nabla_{x^A} E = (y^A - g^A(x^A))/\tau_A$ vanish, namely, where the position sector of Hamilton's canonical equations, $dy^A/dt = \nabla_{x^A} E$, becomes stationary.
We then define the dynamical equations of the system as the momentum sector of the canonical equations restricted to the constraint surface $\Sigma$, with the convention $\coupling_{1,0}\equiv 0$ and $\coupling_{k+1,k}\equiv 0$:
\begin{align}
    \dd{x^A(t)}{t}
    &\coloneq - \nabla_{y^A} E(\set{y^A}, \set{x^A}) \big|_{\set{(y^A, x^A) = (g^A(x^A(t)), x^A(t))}_{A=1}^k}\\
    &= \frac{1}{\tau_A} \left( \frac{\coupling_{A, A-1}}{\tau_{A-1}} \xii{A}{A-1}g^{A-1}\left(x^{A-1}(t)\right) + \frac{\coupling_{A, A+1}}{\tau_{A+1}} \xii{A}{A+1}g^{A+1}\left(x^{A+1}(t)\right) - x^A(t) \right),
\end{align}
where $\nabla_{y^A}$ are the gradient operators with respect to $y^A$, and we used the fact that $(\xii{A+1}{A})^\top = \xii{A}{A+1}$.
The energy function of the class $\H_k$, which serves as a Lyapunov function of the dynamical system, is defined by
\begin{equation}
    \begin{aligned}
        E_{\H_k}(\set{x^A})
        &\coloneq E(\set{y^A}, \set{x^A}) \big|_{\set{y^A = g^A(x^A)}_{A=1}^k}  \\
        &= \sum_{A=1}^k \frac{1}{\tau_A} \left( (x^A)^\top g^A(x^A) - L^A(x^A) \right)
        - \sum_{A=1}^{k-1} \frac{\coupling_{A+1, A}}{\tau_{A+1} \tau_{A}} \big(g^{A+1}(x^{A+1})\big)^\top \xii{A+1}{A} g^A(x^A).
    \end{aligned}
\end{equation}

Since the position sector of the canonical equations is replaced by the constraint $y^A = g^A(x^A)$, the flow is not symplectic, and the energy is not conserved along the trajectory.
Instead, provided that the Hessians of the Lagrangians are positive (semi-)definite, this energy function monotonically decreases along the solution trajectory of the dynamical equations,
\begin{equation}
    \dd{E_{\H_k}(\set{x^A(t)})}{t}
        = - \sum_{A=1}^k \left( \dd{x^A(t)}{t} \right)^\top \hess L^A(x^A(t)) \, \dd{x^A(t)}{t}
        \leq 0,
\end{equation}
which exhibits the dissipative nature of the system.
If, in addition, the overall energy function is bounded from below, the trajectory is guaranteed to converge to a fixed-point attractor state, which corresponds to one of the local minima of the energy function.
Such fixed points may be identified with the stored memories, and the convergence toward them with memory retrieval.
Setting $k=2$ and identifying $x^1=\xv$ and $x^2=\xh$ reproduces the system discussed in the main text.
For the two-layer ($k=2$) case, as in the main text, we drop the subscript $k$ and refer to the model as the class-$\H$ associative memories.

\section{Details on Models A and C}
\label{append:models_a_c_computation_details}

\subsection{A family of associative memories that generates Models A and C}
\label{append:family_of_models_a_c}

In this appendix we characterize the family of visible Lagrangians underlying the construction of \sref{sec:models_a_c} and explain in what sense Models A and C are its distinguished members.
Throughout, the hidden sector is kept general, and it reenters only in the final remark.

\ntextbf{Euler's identity and scale invariance.}
In the energy function \eref{eq:energy-h}, the visible neurons enter both through the Legendre term $\Ev(\xv) = \xv^\top \gv(\xv) - \Lv(\xv)$ and through the activation $\gv(\xv)$ in the interaction term.
The bare visible neurons drop out of the energy precisely when the Legendre term vanishes identically,
\begin{equation}
    \xv^\top \nabla \Lv(\xv) - \Lv(\xv) = 0,
    \label{eq:appAC-euler}
\end{equation}
in which case the energy depends on $\xv$ only through $\gv(\xv)$, as observed in \sref{sec:models_a_c}.
Equation (\ref{eq:appAC-euler}) is Euler's identity, and by Euler's theorem on homogeneous functions it holds on an open cone on which $\Lv$ is differentiable if and only if $\Lv$ is positively homogeneous of degree one there \cite{courant1989introduction},
\begin{equation}
    \Lv(t \xv) = t\, \Lv(\xv),  \qquad  t > 0.
    \label{eq:appAC-homogeneity}
\end{equation}
Differentiating \eref{eq:appAC-homogeneity} with respect to $\xv$ shows that the activation is then homogeneous of degree zero, $\gv(t \xv) = \gv(\xv)$: it is a pure readout of the direction of $\xv$, insensitive to its scale.
This is the family-wide origin of the scale invariance of the energy noted in \sref{sec:models_a_c}.

Degree-one homogeneity fixes $\Lv$ along each ray from the origin once its value on a single cross-section is given.
Choosing the Euclidean unit sphere as the cross-section yields the representation
\begin{equation}
    \Lv(\xv) = \norm{\xv}\, \phi \left( \frac{\xv}{\norm{\xv}} \right),
    \qquad
    \phi \colon \mathbb{S}^{\Nv - 1} \to \R,
    \label{eq:appAC-polar}
\end{equation}
where $\norm{\cdot}$ is the Euclidean norm.
Conversely, every function $\phi$ on the unit sphere defines a positively homogeneous $\Lv$ through \eref{eq:appAC-polar}.
No differentiability is needed for the representation itself.
Smoothness enters only through the correspondence that $\Lv$ is $C^1$ away from the origin exactly when $\phi \in C^1(\mathbb{S}^{\Nv-1})$, which is what defines the activation $\gv = \nabla \Lv$ there.
In particular, $\phi$ need not be an elementary function, so the family is genuinely infinite-dimensional.
In two dimensions, \eref{eq:appAC-polar} is simply $\Lv = r \phi(\theta)$ in polar coordinates, and the convexity requirement introduced next takes the classical form $\phi(\theta) + \phi^{\prime\prime}(\theta) \geq 0$ of the support-function condition \cite{schneider2014convex}.

\ntextbf{Convexity and support functions.}
Within the class $\H$, the Lagrangians are not arbitrary: the monotonic decrease of the energy (the defining property of an associative memory) requires positive semi-definite Hessians, i.e., the Lagrangians must be convex, though not necessarily strictly.
For the nondifferentiable members below, convexity itself is the appropriate formulation of the same requirement.
A convex, everywhere finite, positively degree-one homogeneous function is precisely a sublinear function, and every sublinear function is the support function of a uniquely determined compact convex set $K \subset \R^{\Nv}$ \cite[Sec.~13]{rockafellar1970convex},
\begin{equation}
    \Lv(\xv) = h_K(\xv) \coloneq \max_{u \in K} u^\top \xv,
    \qquad
    K = \partial \Lv(0),
    \label{eq:appAC-support}
\end{equation}
where $\partial \Lv(0)$ denotes the subdifferential at the origin.
The family underlying Models A and C is thus classified by a convex body: the choice of $K$ determines the model.
Moreover, wherever $\Lv$ is differentiable, the gradient is the unique point of $K$ at which the linear function $u \mapsto u^\top \xv$ attains its maximum \cite[Sec.~25]{rockafellar1970convex}, so that
\begin{equation}
    \gv(\xv) = \nabla h_K(\xv) \in \partial K.
    \label{eq:appAC-gradient-image}
\end{equation}
The effective visible degree of freedom is therefore the activation $u = \gv(\xv)$, which lives on the boundary of $K$: the direction of $\xv$ determines the point of $\partial K$ at which the hyperplane with normal $\xv$ supports $K$.

Since the energy is exactly constant along each open ray $\set{t \xv : t > 0}$, the radial direction is a flat direction (an exact zero mode) of $\beta E_\xi$ for every member of the family, and the naive visible integral in \eref{eq:partition-function} diverges in proportion to the infinite radial volume.
This divergence is an overall factor, independent of the patterns and of all order parameters, so it factors out of every observable.
Removing it in the standard collective-coordinate manner leaves an integral over the space of rays.
Any cross-section of the rays represents this quotient, and the natural choice is provided by the activation itself: the visible trace becomes an integral over $u = \gv(\xv) \in \partial K$, rescaled so that the components are $O(1)$ as in the main text.
This is the general form of the prescription anticipated in \sref{sec:partition_function} and adopted for the family in \eref{eq:partition_function_a_c}.
One point deserves emphasis: the quotient fixes the domain of the visible trace but not its measure, which must be supplied as part of the model definition.
For polytopes (including $p = 1, \infty$ below) the activation takes finitely many values and the counting measure is canonical, and for $p = 2$ the rotation-invariant measure on the sphere is singled out by symmetry.
For the other members inequivalent natural choices coexist (e.g., the surface measure and the cone measure on the dual sphere differ by an explicit density \cite{barthe2005probabilistic}), and in \eref{eq:partition_function_a_c} the standard surface measure is understood.

\ntextbf{The $\ell_p$ family.}
For $\Lv = \norm{\xv}_p$ with $1 \leq p \leq \infty$, the maximum in \eref{eq:appAC-support} is H\"older's inequality $u^\top \xv \leq \norm{u}_{p^\prime} \norm{\xv}_p$ together with its equality condition, attained precisely at the activation given in \sref{sec:models_a_c}, and the body is the unit ball of the dual norm,
\begin{equation}
    K = B_{p^\prime} \coloneq \set{u \in \R^{\Nv} \,\middle|\, \norm{u}_{p^\prime} \leq 1},
    \qquad
    \frac{1}{p} + \frac{1}{p^\prime} = 1,
    \label{eq:appAC-dual-ball}
\end{equation}
whose boundary carries the constraint $\norm{\gv}_{p^\prime} = 1$ of \sref{sec:models_a_c}.
Three cases deserve mention.
(i) $p = 2$: $K$ is the Euclidean ball, which is self-dual.
The activation $\gv(\xv) = \xv / \norm{\xv}$ sweeps the whole unit sphere, the visible degrees of freedom are spherical spins, and the member is Model C.
(ii) $p = 1$: $K = [-1, 1]^{\Nv}$ is the hypercube, the dual ball of $\ell_\infty$.
Off the coordinate hyperplanes, the activation is $\gv(\xv) = \sgn(\xv)$ componentwise, whose image is the finite vertex set $\set{\pm 1}^{\Nv}$.
The visible degrees of freedom are Ising spins, and the member is Model A, a discrete spin model arising from continuous neurons through a piecewise linear Lagrangian.
(iii) $p = \infty$: $K = \mathrm{conv} \set{\pm e_1, \ldots, \pm e_{\Nv}}$ is the cross-polytope, with $e_i$ the standard basis vectors.
Wherever the largest component $\abs{\xv_{i^\ast}}$ is unique, $\gv(\xv) = \sgn(\xv_{i^\ast})\, e_{i^\ast}$, so the visible layer collapses to a single winner-take-all (one-hot) degree of freedom with $2 \Nv$ states.

Cases (ii) and (iii) illustrate a general rule.
Abbreviating the local field acting on the visible layer in \eref{eq:partition_function_a_c} by $c_i \coloneq \frac{\beta \coupling}{\th \tv} \sum_\mu F^\prime(\xh_\mu)\, \xii{h}{v}_{\mu i}$, the visible trace takes the form $\int_S d\Omega(\xx)\, e^{c^\top \xx}$: a Laplace-type transform of the chosen measure on $\partial K$.
If $K$ is a polytope with vertices $a_1, \ldots, a_M$, then $h_K(\xv) = \max_{1 \leq l \leq M} a_l^\top \xv$, the activation is piecewise constant with values in the vertex set, and the visible trace is a finite exponential sum,
\begin{equation}
    \int_S d\Omega(\xx)\, e^{c^\top \xx}
    \;\longrightarrow\;
    \sum_{l=1}^{M} w_l\, e^{c^\top a_l},
    \label{eq:appAC-polytope-sum}
\end{equation}
with weights $w_l$ given by the counting measure and the vertices rescaled according to the radius convention of \eref{eq:partition_function_a_c}.
Discrete spin models thus arise within the class $\H$ as the polyhedral shapes $K$, with no need to postulate discrete variables at the outset.
For the hypercube the sum factorizes over the sites, $\sum_{\xx \in \set{\pm 1}^{\Nv}} e^{c^\top \xx} = \prod_i 2 \cosh c_i$, which is the elementary identity behind \eref{eq:modelA-partition-function}.
For the cross-polytope it is a single sum over sites, $\sum_i (e^{c_i} + e^{-c_i})$ up to the radius normalization.
For $p = 2$, on the other hand, the trace is rotation invariant and is evaluated exactly by Gaussian methods in \appenref{append:model_c_replica_analysis}.
Within the $\ell_p$ family, $p = 1$ and $p = 2$ are therefore the two members whose visible trace closes in elementary terms at every $\Nv$.
This is the quantitative content of the footnote in \sref{sec:models_a_c} and the reason the main text focuses on Models A and C.

For the remaining members, general $1 < p < \infty$, no elementary closed form of the visible trace is known, but three standard representations organize the available results.
(i) \emph{Probabilistic representation.}
If $Y = (Y_1, \ldots, Y_{\Nv})$ has i.i.d.~components with density proportional to $e^{-\abs{t}^{p^\prime}}$, then $Y / \norm{Y}_{p^\prime}$ is distributed according to the cone measure on the unit dual sphere $\partial B_{p^\prime}$ and is independent of $\norm{Y}_{p^\prime}$ \cite{schechtman1990volume}:
\begin{equation}
    u \overset{\mathrm{d}}{=} \frac{Y}{\norm{Y}_{p^\prime}},
    \qquad
    Y_i \sim \frac{e^{-\abs{t}^{p^\prime}}}{2 \Gamma(1 + 1/p^\prime)}\, dt
    \quad \text{i.i.d.},
    \label{eq:appAC-sz}
\end{equation}
and the same holds for the surface measure after inserting an explicit density \cite{barthe2005probabilistic}.
This trades the hard constraint for $\Nv$ independent single-site variables coupled only through the norm $\norm{Y}_{p^\prime}$ (the $\ell_{p^\prime}$ analogue of representing the uniform measure on the sphere by a Gaussian vector conditioned on its radius), and is the natural starting point for a saddle-point evaluation of the trace at large $\Nv$.
(ii) \emph{Mellin--Barnes representation.}
Resolving the residual radial constraint in \eref{eq:appAC-sz} by a Mellin--Barnes contour integral expresses the trace as a single contour integral over products of Gamma functions, i.e., a representation of Fox-$H$ type, from which asymptotics can be extracted by shifting the contour \cite{prudnikov1990integrals}.
(iii) \emph{Confluent-hypergeometric special case.}
When the local field has a single non-vanishing component, $c \parallel e_i$, the trace collapses to the one-dimensional Beta-family integral $\int_{-1}^{1} e^{c t} (1 - \abs{t}^{p^\prime})^{(\Nv - 1)/p^\prime - 1}\, dt$ (the one-coordinate marginal of the cone measure), whose term-by-term expansion is a confluent series of Kummer type.
For $p^\prime = 2$ it is precisely a confluent hypergeometric function ${}_1F_1$, equivalently a modified Bessel function \cite[Ch.~13]{olver2010nist}, and for rational $p^\prime$ it reduces to finite combinations of generalized hypergeometric functions \cite{prudnikov1990integrals}.
None of these yields an elementary closed form at a generic field: Models A and C are the two exactly solvable members of the family, between which the other $\ell_p$ members interpolate.

\ntextbf{Remarks.}
Three structural comments conclude this subsection.
(i) \emph{The origin is necessarily singular.}
If $\Lv$ is differentiable at the origin, the subdifferential $K = \partial \Lv(0)$ reduces to the single point $\nabla \Lv(0)$ \cite[Sec.~25]{rockafellar1970convex}, and \eref{eq:appAC-support} gives $\Lv(\xv) = \nabla \Lv(0)^\top \xv$: the Lagrangian is linear, the activation is a constant vector, and the interaction energy no longer depends on the visible configuration: no memory is stored.
Every nontrivial member of the family is therefore non-differentiable at $\xv = 0$: the kinks of $\norm{\cdot}_1$ on the coordinate hyperplanes and the conical point of $\norm{\cdot}_2$ at the origin are structural features of scale-free activations rather than accidental features of the two examples.
Physically, $\xv = 0$ carries no direction information, and a pure direction readout cannot be continuously defined there.
(ii) \emph{Norms and beyond.}
The support function $h_K$ is a norm exactly when $K$ is centrally symmetric with the origin in its interior \cite{schneider2014convex}.
This is the case relevant to the $\pm$-symmetric patterns of the main text.
General members allow asymmetric bodies, e.g., $h_K(\xv) = \max_l a_l^\top \xv$ with a generic vertex set, for which $\Lv(-\xv) \neq \Lv(\xv)$ and patterns and anti-patterns are treated asymmetrically.
The family is in one-to-one correspondence with compact convex bodies, of which the $\ell_p$ balls form a one-parameter slice.
This is the precise content of the footnote in \sref{sec:models_a_c}.
(iii) \emph{The hidden layer.}
The criterion \eref{eq:appAC-euler} applies equally to $\Lh$: a degree-one homogeneous hidden Lagrangian would remove the bare hidden neurons from \eref{eq:energy-h} as well.
The choice $\Lh = \sum_\mu F(\xh_\mu)$ with $F(x) = x^k / k$ is homogeneous of degree $k \neq 1$, so the hidden Legendre term survives (it is precisely the confining term $\Nv \gamma_k \sum_\mu m_\mu^k$ of \eref{eq:modelA-partition-function}), and the bare hidden neurons remain in the energy, which allows them to act as the order parameters of memory retrieval.

\subsection{Model A replica analysis}
\label{append:model_a_replica_analysis}

In this appendix we derive the RS free energy \eref{eq:modelA-free-energy} and the equations of state \eref{eq:modelA-eos-m}--\eref{eq:modelA-eos-r} of Model A.
We follow the standard procedure of the replica method \cite{mezard1987spin,nishimori2001statistical}: after averaging the replicated partition function over the patterns, the crosstalk noise is characterized by an overlap matrix $R$ and its conjugate $\hat{R}$, and the hidden-sector integral factorizes over the non-condensed modes.
For $k=2$ the hidden sector is Gaussian, the computation closes exactly, and it reproduces the AGS theory of the Hopfield model \cite{amit1985spin,amit1985storing,amit1987statistical}.
For $k>2$, in contrast, we show that the non-condensed hidden integral admits no consistent evaluation at finite $\beta_k$ (a genuine pathology of the continuous hidden variables at the load $\Nh = \alpha_k \Nv^{k-1}$, not a technical artifact), so that the analysis must be based on the adiabatic reduction of \sref{sec:partition_function}, for which we then carry out the corrected computation.

\subsubsection{Replica setup and order parameters}
\label{append:modelA-replica-setup}

Throughout we work with the normalized hidden variables $m_\mu = \xh_\mu / (\coupling/\tv)$ introduced in the main text, and the Jacobian of this change of variables only contributes an irrelevant constant.
Introducing the replica index $a = 1, \ldots, n$ and averaging \eref{eq:modelA-partition-function} over the patterns $\xii{h}{v}_{\mu i} \sim \mathrm{Unif}(\set{\pm 1})$, which are independent across the sites $i$, we have
\begin{equation}
    \EE{\xi}{(\za)^n}
        = \int \prod_{a=1}^n dm^a \, e^{-\Nv \gamma_k \sum_{\mu, a} (m^a_\mu)^k}
        \prod_{i=1}^{\Nv} \EE{\xi_i}{\prod_{a=1}^n 2\cosh \beta_k \bigg( \sum_\mu \xii{v}{h}_{i\mu} (m^a_\mu)^{k-1} \bigg)},
    \label{eq:appA-replicated}
\end{equation}
where $\xi_i$ denotes the $i$-th row of the pattern matrix.
Since the pattern distribution is invariant under $\xii{v}{h}_{i\mu} \to \xii{v}{h}_{i1} \xii{v}{h}_{i\mu}$ for each $i$, we may set $\xii{v}{h}_{i1} = 1$ for all $i$ without loss of generality.
To describe the retrieval phase, we assume that only the first pattern is condensed,
\begin{equation}
    m^a \coloneq m^a_1 = O(1),  \qquad  m^a_\mu = O(\Nv^{-1/2}) \quad (\mu \geq 2),
\end{equation}
and collect the non-condensed contributions to the local field into the crosstalk noise
\begin{equation}
    u_i^a \coloneq \sum_{\mu \geq 2} \xii{v}{h}_{i\mu} (m^a_\mu)^{k-1}.
\end{equation}
For fixed $\set{m^a_\mu}$, the vectors $u_i = (u_i^a)_{a=1}^n$ are i.i.d.~across the sites with zero mean, and by the central limit theorem they become Gaussian at large $\Nv$,
\begin{equation}
    u_i \xrightarrow{\mathrm{d}} \mathcal{N}(0, R),
    \qquad
    R^{ab} \coloneq \sum_{\mu \geq 2} (m^a_\mu)^{k-1} (m^b_\mu)^{k-1},
    \label{eq:appA-noise-covariance}
\end{equation}
where the covariance matrix $R$ is of order $\Nh \times \Nv^{-(k-1)} = \alpha_k = O(1)$ on the assumed scaling of the non-condensed modes.
The site average in \eref{eq:appA-replicated} thus becomes
\begin{equation}
    \EE{\xi_i}{\prod_a 2\cosh \beta_k \big( (m^a)^{k-1} + u_i^a \big)}
        \to \Psi(m, R) \coloneq \EE{z \sim \mathcal{N}(0, R)}{\prod_a 2\cosh \beta_k \big( (m^a)^{k-1} + z^a \big)}.
    \label{eq:appA-Psi}
\end{equation}

We next promote $R^{ab}$ to independent integration variables by inserting the identity
\begin{equation}
    \begin{aligned}
        1 &= \int \prod_{a, b} dR^{ab} \, \delta \bigg( R^{ab} - \sum_{\mu \geq 2} (m^a_\mu)^{k-1} (m^b_\mu)^{k-1} \bigg)  \\
          &\propto \int \prod_{a, b} dR^{ab} d\hat{R}^{ab} \exp \bigg\{ - \frac{\beta_k^2 \Nv}{2} \sum_{a, b} \hat{R}^{ab} \bigg( R^{ab} - \sum_{\mu \geq 2} (m^a_\mu)^{k-1} (m^b_\mu)^{k-1} \bigg) \bigg\},
    \end{aligned}
    \label{eq:appA-conjugate-insertion}
\end{equation}
where the conjugate variables $\hat{R}^{ab}$ run along the imaginary axis and take real values at the saddle point, and the prefactor $\beta_k^2 \Nv / 2$ is chosen for later convenience.
The integrals over the non-condensed modes then factorize,
\begin{equation}
    \prod_{\mu = 2}^{\Nh} \int \prod_a dm^a_\mu \exp \bigg[ - \Nv \gamma_k \sum_a (m^a_\mu)^k + \frac{\beta_k^2 \Nv}{2} \sum_{a, b} \hat{R}^{ab} (m^a_\mu)^{k-1} (m^b_\mu)^{k-1} \bigg]
        = \mathcal{I}_k(\hat{R})^{\Nh - 1},
\end{equation}
with the single-mode integral
\begin{equation}
    \mathcal{I}_k(\hat{R}) \coloneq \int_{\R^n} \prod_a dx^a \exp \bigg[ - \Nv \gamma_k \sum_a (x^a)^k + \frac{\beta_k^2 \Nv}{2} \sum_{a, b} \hat{R}^{ab} (x^a)^{k-1} (x^b)^{k-1} \bigg],
    \label{eq:appA-Ik}
\end{equation}
and the replicated partition function reads
\begin{equation}
    \EE{\xi}{(\za)^n}
        = \int \prod_a dm^a \int \prod_{a, b} dR^{ab} d\hat{R}^{ab}
        \exp \bigg[ - \Nv \gamma_k \sum_a (m^a)^k - \frac{\beta_k^2 \Nv}{2} \sum_{a, b} \hat{R}^{ab} R^{ab} + \Nv \log \Psi(m, R) \bigg] \, \mathcal{I}_k(\hat{R})^{\Nh - 1}.
    \label{eq:appA-Zn}
\end{equation}

We now impose the RS ansatz,
\begin{equation}
    m^a = m,
    \qquad
    R^{aa} = \alpha_k r_d, \quad R^{ab} = \alpha_k r ~(a \neq b),
    \qquad
    \hat{R}^{aa} = \hat{r}_d, \quad \hat{R}^{ab} = \hat{r} ~(a \neq b),
\end{equation}
where the factors $\alpha_k$ are inserted so that $r$ matches the noise parameter of the main text.
To evaluate $\Psi$, we realize the Gaussian vector $z \sim \mathcal{N}(0, R)$ as
\begin{equation}
    z^a = \sqrt{\alpha_k r} \, z + \sqrt{\alpha_k (r_d - r)} \, w^a,
    \qquad
    z, w^a \sim \mathcal{N}(0, 1) ~\text{i.i.d.},
\end{equation}
which decomposes the noise into a component $z$ frozen across the replicas and thermal components $w^a$ independent between the replicas.
Using $\int Dw \, 2\cosh(A + Bw) = 2 e^{B^2/2} \cosh A$ and expanding to first order in $n$, we obtain
\begin{equation}
    \log \Psi(m, R)
        = n \bigg[ \frac{\beta_k^2 \alpha_k}{2} (r_d - r) + \int Dz \log 2\cosh \beta_k \big( m^{k-1} + \sqrt{\alpha_k r} \, z \big) \bigg] + O(n^2).
    \label{eq:appA-logPsi}
\end{equation}
Similarly, the conjugate term in \eref{eq:appA-Zn} becomes
\begin{equation}
    - \frac{\beta_k^2 \Nv}{2} \sum_{a, b} \hat{R}^{ab} R^{ab}
        = - \frac{\beta_k^2 \Nv \alpha_k}{2} \big[ n \hat{r}_d r_d + n(n-1) \hat{r} r \big].
\end{equation}

Two of the saddle-point equations can already be read off, because $\mathcal{I}_k$ depends only on the conjugate variables.
Stationarity of the exponent with respect to $r_d$ gives, from \eref{eq:appA-logPsi} and the conjugate term,
\begin{equation}
    \frac{\beta_k^2 \alpha_k}{2} - \frac{\beta_k^2 \alpha_k}{2} \hat{r}_d = 0
    \qquad \Longrightarrow \qquad
    \hat{r}_d = 1.
    \label{eq:appA-rhat-d}
\end{equation}
Substituting $\hat{r}_d = 1$ back, the two $r_d$-dependent terms cancel identically, so that $r_d$ and $\hat{r}_d$ disappear from the free energy altogether.
The value of $r_d$ itself is fixed by stationarity with respect to $\hat{r}_d$, which identifies $\alpha_k r_d$ with the average of $\sum_{\mu \geq 2} (m^a_\mu)^{2(k-1)}$ in the hidden sector, but it plays no further role.
Stationarity with respect to $r$ gives, using Gaussian integration by parts $\int Dz \, z \tanh \beta_k(m^{k-1} + \sqrt{\alpha_k r} z) = \beta_k \sqrt{\alpha_k r} \int Dz \sech^2 \beta_k(m^{k-1} + \sqrt{\alpha_k r} z)$,
\begin{equation}
    \hat{r} = \int Dz \tanh^2 \beta_k \big( m^{k-1} + \sqrt{\alpha_k r} \, z \big) = q.
    \label{eq:appA-rhat-q}
\end{equation}
The conjugate variable $\hat{r}$ is therefore precisely the Edwards--Anderson order parameter $q$ of the visible spins $s_i = \sgn(\xv_i)$: the integrand of \eref{eq:appA-rhat-q} is the squared thermal average $\ev{s_i}^2$ in the effective single-site measure.
Note that \eref{eq:appA-rhat-d} and \eref{eq:appA-rhat-q} hold for every $k$.
What distinguishes $k=2$ from $k>2$ is the remaining hidden-sector factor $\mathcal{I}_k(\hat{R})$, to which we now turn.

\subsubsection{The case $k=2$: Gaussian hidden sector and the AGS theory}
\label{append:modelA-k2}

For $k=2$ we have $\gamma_2 = \beta_k/2$, and the single-mode integral \eref{eq:appA-Ik} is Gaussian:
\begin{equation}
    \mathcal{I}_2(\hat{R})
        = \int_{\R^n} \prod_a dx^a \exp \bigg[ - \frac{\beta_k \Nv}{2} \sum_{a, b} x^a \big( \delta^{ab} - \beta_k \hat{R}^{ab} \big) x^b \bigg]
        \propto \det \big( I_n - \beta_k \hat{R} \big)^{-1/2},
\end{equation}
up to an irrelevant constant.
Under the RS ansatz, $\hat{R}$ has the eigenvalue $\hat{r}_d + (n-1)\hat{r}$ (non-degenerate) and $\hat{r}_d - \hat{r}$ with degeneracy $n - 1$, so that
\begin{equation}
    \log \det \big( I_n - \beta_k \hat{R} \big)
        = (n - 1) \log \big( 1 - \beta_k (\hat{r}_d - \hat{r}) \big) + \log \big( 1 - \beta_k (\hat{r}_d - \hat{r}) - n \beta_k \hat{r} \big).
\end{equation}
Expanding to first order in $n$ and inserting $\hat{r}_d = 1$, $\hat{r} = q$, we obtain
\begin{equation}
    \log \mathcal{I}_2(\hat{R})
        = - \frac{n}{2} \bigg[ \log \big( 1 - \beta_k (1 - q) \big) - \frac{\beta_k q}{1 - \beta_k (1 - q)} \bigg] + O(n^2)
        = - \frac{n}{2} \Psi_2(q) + O(n^2),
    \label{eq:appA-logI2}
\end{equation}
which, multiplied by $\Nh - 1 \simeq \alpha_k \Nv$, yields precisely the noise entropic term $-\frac{n \Nv \alpha_k}{2} \Psi_2(q)$ of the main text.
Collecting \eref{eq:appA-logPsi}--\eref{eq:appA-logI2} in \eref{eq:appA-Zn} and using the replica trick
\begin{equation}
    f^{\mathrm{A}}(\beta) = - \frac{1}{\beta \Nv} \lim_{n \to 0} \frac{1}{n} \Big( \EE{\xi}{(\za)^n} - 1 \Big),
\end{equation}
we arrive at the RS free energy \eref{eq:modelA-free-energy} with $\Psi_k = \Psi_2$.
In the assembly, the diagonal terms combine as $\frac{\beta_k^2 \alpha_k}{2} (r_d - \hat{r}_d r_d) = 0$, which is the cancellation of $r_d$ anticipated above, while the off-diagonal terms produce $\frac{\alpha_k \beta_k^2}{2} r (1 - q)$ upon using $\hat{r} = q$.
Finally, stationarity with respect to $\hat{r}$ picks up contributions from the conjugate term and from $\log \mathcal{I}_2$:
\begin{equation}
    \frac{\beta_k^2 \alpha_k}{2} r
        = - \alpha_k \frac{\partial}{\partial \hat{r}} \lim_{n\to 0} \frac{1}{n} \log \mathcal{I}_2
        = \frac{\alpha_k}{2} \frac{\beta_k^2 \hat{r}}{\big( 1 - \beta_k (\hat{r}_d - \hat{r}) \big)^2}
    \qquad \Longrightarrow \qquad
    r = \frac{q}{\big( 1 - \beta_k (1 - q) \big)^2} = \mathcal{M}_2(q),
    \label{eq:appA-M2}
\end{equation}
reproducing the AGS equations of state of the Hopfield model near saturation \cite{amit1985spin,amit1985storing,amit1987statistical}.
We emphasize that for $k=2$ no adiabatic assumption has been made: the hidden variables enter quadratically, their Hessian is field-independent, and the Gaussian integration is exact at any $\beta/\th$.

\subsubsection{The case $k>2$: breakdown of the hidden-sector integral and the adiabatic reduction}
\label{append:modelA-breakdown}

For $k>2$ the integral \eref{eq:appA-Ik} does not admit an analogous evaluation, for a reason that is structural rather than technical.
Rescaling $x^a = \Nv^{c} y^a$, the two terms in the exponent of $\mathcal{I}_k$ scale as
\begin{equation}
    \Nv \gamma_k \sum_a (x^a)^k \sim \Nv^{1 + kc},
    \qquad
    \frac{\beta_k^2 \Nv}{2} \sum_{a, b} \hat{R}^{ab} (x^a)^{k-1} (x^b)^{k-1} \sim \Nv^{1 + 2(k-1)c}.
    \label{eq:appA-two-scales}
\end{equation}
For $k = 2$ the two exponents coincide for any $c$, and the choice $c = -1/2$ makes both $O(1)$.
The $\Nh - 1 \simeq \alpha_k \Nv$ modes then sum up to an extensive contribution, which is the calculation of \appenref{append:modelA-k2}.
For $k > 2$, however, no choice of $c$ balances the two terms at a nontrivial order: at the central-limit scale $c = -1/2$, which underlies the ansatz $R = O(1)$ in \eref{eq:appA-noise-covariance}, both exponents are negative and the integrand of $\mathcal{I}_k$ becomes flat, so the integral is not confined to this scale.
Instead, $\mathcal{I}_k$ is dominated by the bare confinement scale $c = -1/k$, on which the coupling term still vanishes as $\Nv^{(2-k)/k}$.
This has two consequences.
First, the typical amplitude of a non-condensed mode is $|m_\mu| \sim \Nv^{-1/k} \gg \Nv^{-1/2}$, so that the diagonal noise covariance evaluates to
\begin{equation}
    R^{aa} = \sum_{\mu \geq 2} (m^a_\mu)^{2(k-1)} \sim \alpha_k \Nv^{k-1} \cdot \Nv^{-2(k-1)/k} = \alpha_k \Nv^{(k-1)(k-2)/k} \longrightarrow \infty,
    \label{eq:appA-runaway}
\end{equation}
in contradiction with $R = O(1)$: the crosstalk variance diverges, and the continuous hidden variables cannot sustain the load $\Nh = \alpha_k \Nv^{k-1}$ at any finite $\beta_k$.
Second, and equivalently, the $\hat{R}$-dependent part of $(\Nh - 1) \log \mathcal{I}_k$ is of order $\Nv^{(k^2 - 2k + 2)/k}$, which is superextensive for $k > 2$, so no extensive saddle-point structure exists for the conjugate variables.
We also note that this pathology cannot be removed by any limit of the time-scale parameters: restoring the original variables via $\xh = (\coupling/\tv) m$ shows that the equilibrium partition function \eref{eq:partition_function_a_c} depends on $\beta/\th$ and $\coupling/\tv$ only through the combination $\beta_k = (\beta/\th)(\coupling/\tv)^k$, so that ``$\beta/\th \to \infty$ at fixed $\beta_k$'' leaves the equilibrium measure unchanged.

For $k > 2$ we therefore \emph{define} Model A through the adiabatic reduction of \sref{sec:partition_function}, in which the hidden neurons are replaced by their stationary values given the visible configuration, as in \eref{eq:partition-function-adiabatic}.
This definition is motivated by the dynamics: in the adiabatic regime $\tv \gg \th$ the hidden neurons deterministically track the visible configuration, and their thermal fluctuations (the source of the runaway \eref{eq:appA-runaway}) are switched off.
Concretely, writing the partition function with the visible spins $s_i \in \set{\pm 1}$ unintegrated and eliminating each mode $m_\mu$ by its stationarity condition, $k \gamma_k m_\mu^{k-1} = (k-1) \beta_k m_\mu^{k-2} \hat{m}_\mu$ with $\hat{m}_\mu(s) \coloneq \frac{1}{\Nv} \sum_i \xii{h}{v}_{\mu i} s_i$, we obtain $m_\mu = \hat{m}_\mu(s)$ and
\begin{equation}
    - \Nv \gamma_k m_\mu^k + \beta_k \Nv m_\mu^{k-1} \hat{m}_\mu \Big|_{m_\mu = \hat{m}_\mu}
        = \frac{\beta_k}{k} \Nv \hat{m}_\mu^k,
    \qquad
    Z^{\mathrm{A,ad}}_\xi \coloneq \sum_{s \in \set{\pm 1}^{\Nv}} \exp \bigg[ \frac{\beta_k}{k} \Nv \sum_\mu \hat{m}_\mu(s)^k \bigg],
    \label{eq:appA-Zad}
\end{equation}
which is the dense associative memory with polynomial energy \cite{NIPS2016_eaae339c}, also known as the multiconnected network \cite{gardner1987multiconnected}.
For the condensed mode this substitution coincides with the exact Laplace evaluation of its integral, whose exponent is $O(\Nv)$ with an $O(1)$ saddle.
For $k = 2$ it agrees with the exact Gaussian integration up to a constant, as noted above.
The definition is therefore consistent across $k$, and the pathology is confined to the thermal fluctuations of the non-condensed continuous modes at $k > 2$.

\subsubsection{The case $k>2$: noise sector from the cumulant expansion}
\label{append:modelA-k-gt-2}

We now carry out the replica analysis of \eref{eq:appA-Zad}.
Replicating and averaging over the patterns, the condensed and non-condensed sectors factorize:
\begin{equation}
    \EE{\xi}{(Z^{\mathrm{A,ad}}_\xi)^n}
        = \sum_{\set{s^a}} \exp \bigg[ \frac{\beta_k}{k} \Nv \sum_a \hat{m}_1(s^a)^k \bigg]
        \prod_{\mu \geq 2} \EE{\xi_\mu}{\exp \bigg( \varepsilon \sum_a (\hat{y}^a_\mu)^k \bigg)},
    \qquad
    \varepsilon \coloneq \frac{\beta_k}{k} \Nv^{1 - k/2},
    \label{eq:appA-Zad-replicated}
\end{equation}
where we used the gauge $\xii{v}{h}_{i1} = 1$ and introduced the rescaled non-condensed overlaps
\begin{equation}
    \hat{y}^a_\mu \coloneq \frac{1}{\sqrt{\Nv}} \sum_i \xii{h}{v}_{\mu i} s_i^a = \sqrt{\Nv} \, \hat{m}_\mu(s^a).
\end{equation}
For fixed spin configurations, the vector $(\hat{y}^a_\mu)_{a=1}^n$ is a normalized sum of i.i.d.~bounded random variables and becomes jointly Gaussian at large $\Nv$ by the central limit theorem,
\begin{equation}
    (\hat{y}^a_\mu)_a \xrightarrow{\mathrm{d}} \mathcal{N}(0, Q),
    \qquad
    Q_{ab} = q_{ab} \coloneq \frac{1}{\Nv} \sum_i s_i^a s_i^b,
    \qquad
    q_{aa} = 1,
\end{equation}
with the replica-overlap matrix of the spins as its covariance.
Since $\varepsilon \to 0$ for $k > 2$, the per-pattern average in \eref{eq:appA-Zad-replicated} is evaluated by the cumulant expansion
\begin{equation}
    \log \EE{\xi_\mu}{e^{\varepsilon \sum_a (\hat{y}^a)^k}}
        = \varepsilon \sum_a \EE{}{(\hat{y}^a)^k}
        + \frac{\varepsilon^2}{2} \sum_{a, b} \mathrm{Cov} \big( (\hat{y}^a)^k, (\hat{y}^b)^k \big)
        + O(\varepsilon^3).
    \label{eq:appA-cumulant}
\end{equation}
The first term is independent of the spin configurations: all single-replica cumulants of $\hat{y}^a$ are spin-independent, because $\kappa_j(\hat{y}^a) = \Nv^{-j/2} \kappa_j(\xi) \sum_i (s_i^a)^j$ vanishes for odd $j$ (symmetry of $\xi$) and reduces to $\Nv^{1 - j/2} \kappa_j(\xi)$ for even $j$ ($s_i^2 = 1$).
It therefore contributes a state-independent constant (superextensive, of order $\Nv^{k/2}$ after summing over the patterns, but a pure constant) relative to which we define the free energy (see also Remark~2 below).
The second term is governed by the covariance kernel
\begin{equation}
    \Phi_k(q) \coloneq \mathrm{Cov} \big( X^k, Y^k \big),
    \qquad
    (X, Y) ~\text{standard Gaussian with correlation } \EE{}{XY} = q,
    \label{eq:appA-Phik}
\end{equation}
evaluated at $q = q_{ab}$.
Note that $\Phi_k(0) = 0$, and the diagonal terms $a = b$ contribute the $q$-independent constant $n \Phi_k(1)$, which we drop.
Summing over the $\Nh - 1 \simeq \alpha_k \Nv^{k-1}$ patterns, the noise sector contributes the extensive action
\begin{equation}
    (\Nh - 1) \, \frac{\varepsilon^2}{2} \sum_{a \neq b} \Phi_k(q_{ab})
        = \frac{\alpha_k \beta_k^2}{2 k^2} \Nv \sum_{a \neq b} \Phi_k(q_{ab}),
    \label{eq:appA-noise-action}
\end{equation}
while the third and higher cumulants are subextensive for even $k \geq 4$ (Remark~2).
In this representation the collapse of the noise-sector order parameters is manifest: the covariance $R^{ab}$ of \eref{eq:appA-noise-covariance}, evaluated on the enslaved modes $m_\mu = \hat{m}_\mu(s)$, self-averages by the law of large numbers to
\begin{equation}
    R^{ab}
        = \Nv^{-(k-1)} \sum_{\mu \geq 2} (\hat{y}^a_\mu)^{k-1} (\hat{y}^b_\mu)^{k-1}
        \longrightarrow \alpha_k \EE{}{X^{k-1} Y^{k-1}} \Big|_{q = q_{ab}} = \alpha_k \mathcal{M}_k(q_{ab}),
    \label{eq:appA-R-slaved}
\end{equation}
so that, in contrast to the $k=2$ computation, the pair $(R, \hat{R})$ carries no independent degrees of freedom: the crosstalk noise is entirely slaved to the spin overlap $q_{ab}$.
In particular the diagonal element is finite, $R^{aa} \to \alpha_k \mathcal{M}_k(1) = \alpha_k (2k-3)!!$, in contrast with the divergence \eref{eq:appA-runaway} of the continuous model.

The remaining steps are standard.
We introduce the condensed overlap and the spin overlaps with their conjugates,
\begin{equation}
    1 = \int \prod_a dm^a \, \delta \Big( \Nv m^a - \sum_i s_i^a \Big),
    \quad
    1 = \int \prod_{a < b} dq_{ab} \, \delta \Big( \Nv q_{ab} - \sum_i s_i^a s_i^b \Big),
\end{equation}
represented with conjugate variables $\tilde{m}^a$ and $\hat{q}_{ab}$ as in \eref{eq:appA-conjugate-insertion}, upon which the spin trace factorizes over the sites.
Under the RS ansatz $m^a = m$, $q_{ab} = q$, $\hat{q}_{ab} = \hat{q}$ ($a < b$), the conjugate $\tilde{m}^a = \tilde{m}$ is eliminated by its saddle $\tilde{m} = \beta_k m^{k-1}$, and the single-site trace gives the familiar
\begin{equation}
    \lim_{n \to 0} \frac{1}{n} \log \mathrm{Tr}_{s} \exp \bigg[ \frac{\hat{q}}{2} \Big( \sum_a s^a \Big)^2 - \frac{n \hat{q}}{2} + \tilde{m} \sum_a s^a \bigg]
        = - \frac{\hat{q}}{2} + \int Dz \log 2\cosh \big( \beta_k m^{k-1} + \sqrt{\hat{q}} \, z \big).
\end{equation}
Stationarity with respect to $\hat{q}$ returns $q = \int Dz \tanh^2 (\beta_k m^{k-1} + \sqrt{\hat{q}} z)$, while stationarity with respect to $q$ ties the conjugate to the noise kernel:
\begin{equation}
    \hat{q} = \frac{\alpha_k \beta_k^2}{k^2} \Phi_k^\prime(q).
    \label{eq:appA-qhat}
\end{equation}
The derivative of the kernel is evaluated by Price's theorem \cite{price1958useful}: for jointly Gaussian $(X, Y)$ with correlation $q$,
\begin{equation}
    \frac{\partial}{\partial q} \EE{}{X^k Y^k} = k^2 \, \EE{}{X^{k-1} Y^{k-1}}.
    \label{eq:appA-price}
\end{equation}
A short proof follows from the Hermite expansion: writing $x^m = \sum_j c^{(m)}_j \mathrm{He}_j(x)$ with the probabilists' Hermite polynomials $\mathrm{He}_j$ and using the orthogonality $\EE{}{\mathrm{He}_i(X) \mathrm{He}_j(Y)} = \delta_{ij} \, j! \, q^j$, one has $\EE{}{X^m Y^m} = \sum_j (c^{(m)}_j)^2 j! \, q^j$.
Differentiating term by term and using $k c^{(k-1)}_{j-1} = j c^{(k)}_j$, which is the Hermite-coefficient form of $\frac{d}{dx} x^k = k x^{k-1}$ together with $\mathrm{He}_j^\prime = j \mathrm{He}_{j-1}$, yields \eref{eq:appA-price}.
Combining \eref{eq:appA-qhat} and \eref{eq:appA-price} and defining the noise parameter as in \eref{eq:appA-R-slaved},
\begin{equation}
    r \coloneq \mathcal{M}_k(q) = \EE{}{X^{k-1} Y^{k-1}},
    \qquad
    \hat{q} = \alpha_k \beta_k^2 \, \mathcal{M}_k(q) = \alpha_k \beta_k^2 r,
\end{equation}
the local field becomes $\beta_k m^{k-1} + \sqrt{\hat{q}} z = \beta_k (m^{k-1} + \sqrt{\alpha_k r} z)$, matching the main text.
Assembling all the terms at $n \to 0$, the $\Phi_k$ contribution enters the free energy as $\frac{\alpha_k \beta_k^2}{2 k^2} \Phi_k(q)$, and since $\Phi_k(0) = 0$ and $\Phi_k^\prime = k^2 \mathcal{M}_k$,
\begin{equation}
    \frac{\alpha_k \beta_k^2}{2 k^2} \Phi_k(q) = \frac{\alpha_k \beta_k^2}{2} \int_0^q \mathcal{M}_k(s) \, ds = \frac{\alpha_k}{2} \Psi_k(q),
\end{equation}
which is precisely the noise entropic term of \eref{eq:modelA-free-energy} for $k > 2$.
The terms $\gamma_k m^k$ (from $\frac{\beta_k}{k} m^k - \tilde{m} m$ at $\tilde{m} = \beta_k m^{k-1}$) and $\frac{\alpha_k \beta_k^2}{2} r (1 - q)$ (from $\frac{\hat{q}}{2}(1 - q)$) complete the RS free energy, up to the additive constant $-\frac{\alpha_k \beta_k^2}{2 k^2} \Phi_k(1)$ and the state-independent constants discussed above.
The equations of state \eref{eq:modelA-eos-m}--\eref{eq:modelA-eos-r} then follow from stationarity, as verified directly in the main text.

\subsubsection{Consistency checks and remarks}
\label{append:modelA-remarks}

\ntextbf{Remark 1: the Onsager reaction field.}
The two cases of $\mathcal{M}_k$ can be cross-checked by a cavity argument that does not rely on replicas.
In the effective spin model \eref{eq:appA-Zad}, adding a single non-condensed mode $\mu$ shifts the local field on the spin $s_i$ by $\beta_k \xii{v}{h}_{i\mu} \hat{m}_\mu^{k-1}$, and hence its thermal average by $\delta \ev{s_i} = \beta_k (1 - \ev{s_i}^2) \, \xii{v}{h}_{i\mu} \hat{m}_\mu^{k-1}$ in linear response.
Feeding this back into the overlap of the same mode gives the self-feedback (Onsager reaction)
\begin{equation}
    \delta \hat{m}_\mu = \frac{1}{\Nv} \sum_i \xii{h}{v}_{\mu i} \, \delta \ev{s_i} = \beta_k (1 - q) \, \hat{m}_\mu^{k-1},
    \qquad \text{i.e.} \qquad
    \delta \hat{y}_\mu = \beta_k (1 - q) \, \Nv^{-(k-2)/2} \, \hat{y}_\mu^{k-1}.
\end{equation}
For $k = 2$ the feedback is marginal, $O(1)$, and must be resummed to all orders: with the frozen cavity field $\sqrt{q} z$, the self-consistent solution is $\ev{\hat{y}}_z = \sqrt{q} z / (1 - \beta_k (1 - q))$, so that $r = \int Dz \ev{\hat{y}}_z^2 = q / (1 - \beta_k(1-q))^2 = \mathcal{M}_2(q)$.
The geometric resummation is the origin of the AGS denominator, in agreement with \appenref{append:modelA-k2}.
For $k > 2$ the feedback vanishes in the thermodynamic limit, the non-condensed overlaps remain bare Gaussian variables, and $r = \mathcal{M}_k(q)$ is their bare moment, in agreement with \appenref{append:modelA-k-gt-2}.
This is the content of the scaling statement in the main text.

\ntextbf{Remark 2: validity of the cumulant expansion.}
The truncation of \eref{eq:appA-cumulant} at second order is controlled as follows.
(i) The third cumulant contributes $O(\varepsilon^3)$ per pattern, hence $O(\Nv^{2 - k/2})$ in total after multiplication by $\Nh$, which is subextensive for $k \geq 4$.
Since $k$ is even in our setting, this covers all $k > 2$, and higher cumulants are smaller still.
(ii) The corrections to the joint Gaussianity of $(\hat{y}^a)_a$ are of relative order $\Nv^{-1}$ (Edgeworth), and, for $\pm 1$ spins and symmetrically distributed patterns, the single-replica corrections are exactly state-independent, as noted below \eref{eq:appA-cumulant}.
The state-dependent cross-replica corrections contribute only at $O(1)$ in total.
(iii) The first cumulant produces a state-independent constant of order $\Nv^{k/2}$, which is superextensive but common to all configurations and all phases, and drops out of every order-parameter equation and free energy difference.
(iv) The evenness of $k$ is used twice: it bounds the hidden potential from below, and it guarantees the vanishing of the odd pattern cumulants used in (i) and (ii).

\ntextbf{Remark 3: closed form of the noise kernel.}
The Hermite expansion provides the explicit polynomial form of $\mathcal{M}_k$.
Writing $x^{k-1} = \sum_j c_{k,j} \mathrm{He}_j(x)$ with
\begin{equation}
    c_{k,j} = \frac{(k-1)!}{j! \, 2^{(k-1-j)/2} \left( \frac{k-1-j}{2} \right)!},
    \qquad
    j \equiv k - 1 ~(\mathrm{mod}~2),
    \quad
    0 \leq j \leq k - 1,
\end{equation}
the orthogonality $\EE{}{\mathrm{He}_i(X) \mathrm{He}_j(Y)} = \delta_{ij} \, j! \, q^j$ gives
\begin{equation}
    \mathcal{M}_k(q) = \sum_{j} c_{k,j}^2 \, j! \, q^j,
    \qquad
    \mathcal{M}_k(1) = \EE{}{X^{2(k-1)}} = (2k-3)!!,
\end{equation}
a polynomial with positive coefficients, for instance $\mathcal{M}_4(q) = 9q + 6q^3$ and $\mathcal{M}_6(q) = 225 q + 600 q^3 + 120 q^5$.
The same polynomial kernel appears as the noise covariance in the dynamical mean-field theory of dense associative memories with the polynomial (Krotov--Hopfield-type) energy \cite{mimura2025dynamical,sumikawa2026testing}, with the static correspondence between the equal-time correlation and the replica overlap.
The monomial kernel $\propto q^{k-1}$ familiar from $p$-spin models arises instead for the diagonal-free (Abbott--Arian-type) variant of the interaction \cite{abbott1987storage,sumikawa2026testing}, which is a different model from ours.

\ntextbf{Remark 4: the zero-temperature limit.}
The zero-temperature equations quoted in \sref{sec:model_a} follow from \eref{eq:modelA-eos-m}--\eref{eq:modelA-eos-r} by standard manipulations.
As $\beta_k \to \infty$ at fixed $C = \beta_k (1 - q)$, the hyperbolic tangent in \eref{eq:modelA-eos-m} reduces to the sign of its argument, and $\int Dz\, \sgn ( m^{k-1} + \sqrt{\alpha_k r}\, z ) = \erf ( m^{k-1} / \sqrt{2 \alpha_k r} )$ yields \eref{eq:modelA-zeroT-m}.
For the frozen response, one combines $1 - q = \int Dz \sech^2 \beta_k ( m^{k-1} + \sqrt{\alpha_k r}\, z )$ with $\beta_k \sech^2 (\beta_k x) \to 2 \delta(x)$.
The delta function picks up the Gaussian density at the zero $z_0 = -m^{k-1} / \sqrt{\alpha_k r}$ of the local field, which gives \eref{eq:modelA-zeroT-C}.
Finally, $q \to 1$ at fixed $C$ reduces the crosstalk moment \eref{eq:modelA-Mk} to \eref{eq:modelA-zeroT-r}: for $k = 2$ the denominator survives as $1 - \beta_k (1 - q) \to 1 - C$, while for $k > 2$ the polynomial is simply evaluated at $q = 1$.

\subsection{Model C replica analysis}
\label{append:model_c_replica_analysis}

In this appendix we derive the RS free energy \eref{eq:modelC-free-energy} and the equations of state \eref{eq:modelC-eos-m}--\eref{eq:modelC-eos-q} of Model C.
The hidden sector of Model C coincides with that of Model A: after the conjugate insertion, the non-condensed hidden modes factorize into the same single-mode integral \eref{eq:appA-Ik}, so the dichotomy established in \appenref{append:model_a_replica_analysis} carries over: the hidden sector closes exactly for $k=2$ and admits no consistent evaluation for $k>2$, where the model is defined by the adiabatic reduction.
We exploit this from the outset by working with the formulation in which the hidden variables are eliminated in favor of the pattern overlaps.
This formulation is \emph{exact} for $k=2$ and \emph{definitional} for $k>2$, and it leads directly to the two-parameter free energy quoted in the main text.
The computation that retains the hidden variables explicitly is presented in \appenref{append:modelC-direct} for completeness.
What is genuinely new compared with Model A is the visible sector: the spherical trace is Gaussian and is carried out exactly, so that no single-site integral remains in the final expressions.

\subsubsection{Overlap formulation and replica setup}
\label{append:modelC-setup}

We start from the partition function \eref{eq:modelC-partition-function}.
The flat radial direction of the visible variables has already been removed by the sphere-restricted definition of \eref{eq:partition_function_a_c}, and no further regularization is needed.
The basic objects are the pattern overlaps of a visible configuration $\xx \in S$,
\begin{equation}
    \hat{m}_\mu(\xx) \coloneq \frac{1}{\Nv} \sum_i \xii{h}{v}_{\mu i} \xx_i .
\end{equation}
For $k=2$ the hidden variables can be eliminated exactly: $\gamma_2 = \beta_k / 2$, each $m_\mu$ appears quadratically in \eref{eq:modelC-partition-function}, and the Gaussian integral gives, up to an overall constant,
\begin{equation}
    \int dm_\mu \exp \Big( - \frac{\beta_k}{2} \Nv m_\mu^2 + \beta_k \Nv \hat{m}_\mu(\xx) \, m_\mu \Big)
        \propto \exp \Big( \frac{\beta_k}{2} \Nv \hat{m}_\mu(\xx)^2 \Big).
    \label{eq:appC-gauss-elimination}
\end{equation}
For $k>2$ the direct evaluation of the hidden sector fails in exactly the manner analyzed in \appenref{append:modelA-breakdown}: the non-condensed modes factorize into the single-mode integral $\mathcal{I}_k(\hat{R})$ of \eref{eq:appA-Ik}, which admits no extensive saddle-point structure, and the crosstalk variance of the continuous modes suffers the runaway \eref{eq:appA-runaway}.
We return to this formulation in \appenref{append:modelC-direct}.
As for Model A, for $k>2$ we therefore \emph{define} Model C through the adiabatic reduction of \sref{sec:partition_function}.
The enslaved value $m_\mu = \hat{m}_\mu(\xx)$ follows from the same per-mode stationarity algebra as in \eref{eq:appA-Zad}, and we work with
\begin{equation}
    Z^{\mathrm{C,ad}}_\xi \coloneq \int_S d\Omega(\xx) \, \exp \bigg[ \frac{\beta_k}{k} \Nv \sum_\mu \hat{m}_\mu(\xx)^k \bigg],
    \label{eq:appC-Zeff}
\end{equation}
which is exact for $k=2$ by \eref{eq:appC-gauss-elimination} and is the adiabatic definition for $k>2$, so that the analysis below is uniform in $k$.

Replicating \eref{eq:appC-Zeff} and averaging over the patterns, we use the rotational invariance of the spherical ensemble to rotate the condensed pattern to $\xii{h}{v}_1 = (1, \ldots, 1)$.
This gauge choice is exact at finite $\Nv$ (Remark~1 below), and the condensed term becomes a function of the visible magnetizations,
\begin{equation}
    \hat{m}_1(\xx^a) = \frac{1}{\Nv} \sum_i \xx^a_i \eqcolon m^a .
\end{equation}
For the non-condensed patterns we introduce the rescaled overlaps
\begin{equation}
    \hat{y}^a_\mu \coloneq \frac{1}{\sqrt{\Nv}} \sum_i \xii{h}{v}_{\mu i} \xx^a_i = \sqrt{\Nv} \, \hat{m}_\mu(\xx^a).
    \label{eq:appC-yhat}
\end{equation}
Since the spherical ensemble has $\EE{}{\xii{v}{h}_{i\mu} \xii{v}{h}_{j\mu}} = \delta_{ij}$, the covariance of the overlaps is exactly the replica-overlap matrix of the visible variables,
\begin{equation}
    \EE{\xi_\mu}{\hat{y}^a_\mu \hat{y}^b_\mu} = \frac{1}{\Nv} \sum_i \xx^a_i \xx^b_i \eqcolon q_{ab},
    \qquad
    q_{aa} = 1,
    \label{eq:appC-covariance}
\end{equation}
where the diagonal is fixed by the spherical constraint, the counterpart for continuous spins of $s_i^2 = 1$ in Model A.
At large $\Nv$ the vector $(\hat{y}^a_\mu)_a$ becomes jointly Gaussian with covariance matrix $Q = (q_{ab})$.
Moreover, by rotational invariance the exact per-pattern average depends on the visible configurations only through $Q$, so that all finite-$\Nv$ corrections are automatically functions of the order parameters (Remark~1).

\subsubsection{Noise sector}
\label{append:modelC-noise}

The average over one non-condensed pattern produces the noise factor
\begin{equation}
    \EE{\xi_\mu}{\exp \bigg( \varepsilon \sum_a (\hat{y}^a_\mu)^k \bigg)},
    \qquad
    \varepsilon = \frac{\beta_k}{k} \Nv^{1 - k/2},
    \label{eq:appC-noise-factor}
\end{equation}
of the same form as in \eref{eq:appA-Zad-replicated}.

For $k = 2$ we have $\varepsilon = \beta_k / 2 = O(1)$, and the Gaussian asymptotics of $(\hat{y}^a_\mu)_a$ gives the closed form
\begin{equation}
    \EE{\xi_\mu}{e^{\frac{\beta_k}{2} \sum_a (\hat{y}^a_\mu)^2}}
        \to \det \big( I_n - \beta_k Q \big)^{-1/2},
    \label{eq:appC-det}
\end{equation}
convergent for $\beta_k \lambda_{\max}(Q) < 1$, with $\lambda_{\max}$ the largest eigenvalue of $Q$.
Under the RS ansatz $q_{ab} = q$ ($a \neq b$), the eigenvalues of $Q$ are $1 - q$ with degeneracy $n - 1$ and $1 + (n-1) q$, both of which tend to $1 - q$ in the replica limit, so that
\begin{equation}
    \log \det \big( I_n - \beta_k Q \big)
        = (n - 1) \log \big( 1 - \beta_k (1 - q) \big) + \log \big( 1 - \beta_k (1 - q) - n \beta_k q \big)
        = n \, \Psi_2(q) + O(n^2),
    \label{eq:appC-det-RS}
\end{equation}
which is the same algebra as \eref{eq:appA-logI2} evaluated at $(\hat{r}_d, \hat{r}) = (1, q)$.
Multiplying by $-\frac{1}{2}(\Nh - 1) \simeq -\frac{\alpha_k \Nv}{2}$, the noise sector contributes $-\frac{n \alpha_k \Nv}{2} \Psi_2(q)$ to the replicated exponent, i.e., $+\frac{\alpha_k}{2} \Psi_2(q)$ to $\beta f^{\mathrm{C}}$.

For $k > 2$ we have $\varepsilon \to 0$, and \eref{eq:appC-noise-factor} is evaluated by the cumulant expansion \eref{eq:appA-cumulant}, whose structure carries over intact.
The first cumulant is state-independent because it depends only on the diagonal $q_{aa} = 1$, now enforced by the spherical constraint.
The second cumulant is governed by the same covariance kernel $\Phi_k(q_{ab})$ of \eref{eq:appA-Phik}.
The third and higher cumulants are subextensive for even $k \geq 4$, with the Model C-specific aspects of the power counting collected in Remark~2.
Summing over the $\Nh - 1$ patterns, the noise sector contributes the extensive action \eref{eq:appA-noise-action}, which, by Price's theorem \eref{eq:appA-price}, integrates to $\frac{\alpha_k}{2} \Psi_k(q)$ in the free energy, exactly as in \appenref{append:modelA-k-gt-2}.

In both cases, therefore, the noise sector enters $\beta f^{\mathrm{C}}$ as $\frac{\alpha_k}{2} \Psi_k(q)$, with the uniform derivative $\Psi_k^\prime(q) = \beta_k^2 \mathcal{M}_k(q)$ quoted in the main text (for $k=2$ directly from \eref{eq:appC-det-RS}, for $k>2$ from Price's theorem).
Note also that the noise factors depend on the visible configurations only through the overlap matrix $Q$, which is precisely the quantity fixed by the conjugate insertions of the next subsection.

\subsubsection{Visible Gaussian sector and the reduced free energy}
\label{append:modelC-RS}

The remaining trace over the visible variables is organized by introducing the condensed overlaps and the replica overlaps with their conjugates,
\begin{equation}
    1 = \int \prod_a dm^a \, \delta \Big( \Nv m^a - \sum_i \xx^a_i \Big),
    \qquad
    1 = \int \prod_{a < b} dq_{ab} \, \delta \Big( \Nv q_{ab} - \sum_i \xx^a_i \xx^b_i \Big),
\end{equation}
represented with conjugate variables $\tilde{m}^a$ and $\hat{q}_{ab}$ as in \eref{eq:appA-conjugate-insertion}, together with the spherical constraints
\begin{equation}
    1 = \int_{c - \ii \infty}^{c + \ii \infty} \prod_a \frac{du^a}{4\pi \ii} \exp \bigg[ - \frac{u^a}{2} \Big( \sum_i (\xx^a_i)^2 - \Nv \Big) \bigg],
    \qquad c > 0.
    \label{eq:appC-spherical-multiplier}
\end{equation}
The multiplier $u^a$ is precisely the conjugate of the diagonal overlap $q_{aa}$: it absorbs the entire diagonal sector, playing the role of the pair $(r_d, \hat{r}_d)$ in Model A, where we found $\hat{r}_d = 1$ and a complete cancellation of $r_d$ (\eref{eq:appA-rhat-d} and below).
After these insertions the visible integral factorizes over the sites.
Under the RS ansatz $m^a = m$, $\tilde{m}^a = \tilde{m}$, $u^a = u$, $q_{ab} = q$, $\hat{q}_{ab} = \hat{q}$, the single-site measure is an $n$-dimensional Gaussian.
Decoupling the replica coupling with a frozen Gaussian field, $e^{\hat{q} \sum_{a<b} \xx^a \xx^b} = e^{-\frac{n\hat{q}}{2}} \int Dz \, e^{\sqrt{\hat{q}} z \sum_a \xx^a}$, each site contributes
\begin{equation}
    \int Dz \prod_a \int d\xx^a \exp \Big[ - \frac{D}{2} (\xx^a)^2 + \big( \tilde{m} + \sqrt{\hat{q}} \, z \big) \xx^a \Big],
    \qquad
    D \coloneq u + \hat{q},
\end{equation}
whose logarithm is, expanding to first order in $n$ and dropping constants,
\begin{equation}
    \lim_{n \to 0} \frac{1}{n} \log (\text{site factor})
        = - \frac{1}{2} \log D + \frac{\tilde{m}^2 + \hat{q}}{2 D}.
    \label{eq:appC-single-site}
\end{equation}
Collecting the condensed term, the conjugate terms, the constraint terms, the noise sector of \appenref{append:modelC-noise}, and \eref{eq:appC-single-site}, we obtain the variational free energy, up to additive constants,
\begin{equation}
    \beta f^{\mathrm{C}}
        = - \frac{\beta_k}{k} m^k + \tilde{m} m - \frac{\hat{q} q}{2} + \frac{\alpha_k}{2} \Psi_k(q)
        - \frac{D - \hat{q}}{2} + \frac{1}{2} \log D - \frac{\tilde{m}^2 + \hat{q}}{2 D},
    \label{eq:appC-variational}
\end{equation}
where we traded the multiplier $u = D - \hat{q}$ for $D$.

The stationarity conditions of \eref{eq:appC-variational} are elementary.
Variations with respect to $\tilde{m}$ and $D$ give
\begin{equation}
    \tilde{m} = D m,
    \qquad
    \frac{1}{D} + \frac{\tilde{m}^2 + \hat{q}}{D^2} = 1,
    \label{eq:appC-linear-saddles}
\end{equation}
the latter being the spherical constraint $\frac{1}{\Nv} \sum_i \ev{\xx_i^2} = 1$, while variation with respect to $\hat{q}$ identifies
\begin{equation}
    q = \frac{\tilde{m}^2 + \hat{q}}{D^2} = m^2 + \frac{\hat{q}}{D^2}.
    \label{eq:appC-q-saddle}
\end{equation}
This exhibits $q$ as the Edwards--Anderson order parameter: in the single-site measure, $\ev{\xx} = (\tilde{m} + \sqrt{\hat{q}} z) / D$, so that $q = \int Dz \ev{\xx}^2$ decomposes into the condensed part $m^2$ and the glassy part $\hat{q} / D^2$.
Solving \eref{eq:appC-linear-saddles}--\eref{eq:appC-q-saddle},
\begin{equation}
    D = \frac{1}{1 - q},
    \qquad
    \tilde{m} = \frac{m}{1 - q},
    \qquad
    \hat{q} = \frac{q - m^2}{(1 - q)^2},
    \label{eq:appC-solved}
\end{equation}
so that $1 - q = 1/D$ is the single-site thermal variance left after the condensed and glassy components are removed, and $\beta_k (1 - q)$ is the static susceptibility of the spherical state.
The remaining two conditions are the equations of state.
Variation with respect to $q$ gives, using $\Psi_k^\prime = \beta_k^2 \mathcal{M}_k$,
\begin{equation}
    \hat{q} = \alpha_k \beta_k^2 \, \mathcal{M}_k(q)
    \qquad \Longrightarrow \qquad
    \frac{q - m^2}{(1 - q)^2} = \alpha_k \beta_k^2 \, \mathcal{M}_k(q),
    \label{eq:appC-qhat}
\end{equation}
which is \eref{eq:modelC-eos-q}, and variation with respect to $m$ gives $\tilde{m} = \beta_k m^{k-1}$, which combined with $\tilde{m} = m / (1 - q)$ yields the signal equation \eref{eq:modelC-eos-m}.

Finally, we eliminate the auxiliary variables.
Substituting $D$ and $\tilde{m}$ from \eref{eq:appC-solved} into \eref{eq:appC-variational}, the $\hat{q}$-dependent terms cancel identically,
\begin{equation}
    - \frac{\hat{q} q}{2} + \frac{\hat{q}}{2} - \frac{\hat{q}}{2} (1 - q) = 0,
\end{equation}
(the three terms coming from the conjugate, constraint, and single-site terms, respectively, so that $\hat{q}$ need not even be substituted), while $\tilde{m} m - \tilde{m}^2 / (2D) = m^2 / (2 (1 - q))$, and we arrive at
\begin{equation}
    \beta f^{\mathrm{C}}
        = - \frac{\beta_k}{k} m^k + \frac{\alpha_k}{2} \Psi_k(q)
        - \frac{1}{2} \left[ \log (1 - q) + \frac{q - m^2}{1 - q} \right] - \frac{1}{2},
    \label{eq:appC-reduced}
\end{equation}
which is the free energy \eref{eq:modelC-free-energy} of the main text, up to the additive constant $-1/2$.
Because the eliminated variables were removed at their exact stationary points, the stationarity of \eref{eq:appC-reduced} in $(m, q)$ reproduces \eref{eq:modelC-eos-m}--\eref{eq:modelC-eos-q}, as stated in the main text.

\subsubsection{The case \texorpdfstring{$k=2$}{k=2}: marginality and zero capacity}
\label{append:modelC-k2}

For $k = 2$ the two expressions for the conjugate field obtained above, $\tilde{m} = \beta_k m$ and $\tilde{m} = m / (1 - q)$, are compatible only if
\begin{equation}
    \big[ 1 - \beta_k (1 - q) \big] \, m = 0 .
    \label{eq:appC-k2-degeneracy}
\end{equation}
A retrieval state thus requires the marginality condition $\beta_k (1 - q) = 1$, and when it holds the magnitude of $m$ is left undetermined at this order: the condensed direction is a flat direction (zero mode) of the quadratic spherical energy rather than a genuine minimum.
At $\alpha_k = 0$ the flatness is lifted by the spherical constraint itself: \eref{eq:appC-qhat} gives $q = m^2$, and $\beta_k (1 - m^2) = 1$ determines $m^2 = 1 - T$ below $T_c = 1$, as quoted in the main text.

The noise sector exhibits the same marginality from a complementary viewpoint.
The exact $k=2$ noise factor \eref{eq:appC-det} is convergent only for $\beta_k \lambda_{\max}(Q) < 1$, which under the RS ansatz becomes $\beta_k (1 - q) < 1$ in the replica limit.
A retrieval state lies precisely at the boundary of this normalizability region, where $\det (I_n - \beta_k Q) \to 0$: the fluctuations of the non-condensed overlaps soften and condense, which is the standard condensation mechanism of spherical models \cite{berlin1952spherical,kosterlitz1976spherical}, and the divergence of $\mathcal{M}_2(q) = q / (1 - \beta_k (1 - q))^2$ is its precursor.
Consequently the noise equation \eref{eq:modelC-eos-q} admits no solution with $m > 0$ at any $\alpha_k > 0$, including $T = 0$, and the quadratic spherical model has zero storage capacity, in agreement with the marginal behavior of the spherical Hopfield model \cite{bolle2003spherical}.
There, a retrieval phase is restored by augmenting the Hamiltonian with a quartic term.
Within the class $\H$, the same lifting of the flat direction is provided by the hidden nonlinearity with $k > 2$.

\subsubsection{Direct computation with continuous hidden variables}
\label{append:modelC-direct}

For completeness, we sketch the computation that retains the hidden variables explicitly, parallel to \appenref{append:modelA-replica-setup}, and show where it connects to the overlap formulation above.
Averaging the replicated \eref{eq:modelC-partition-function} over the non-condensed patterns gives the noise coupling $\frac{\beta_k^2}{2} \sum_{a,b} R^{ab} \sum_i \xx^a_i \xx^b_i$ with the same covariance matrix $R^{ab} = \sum_{\mu \geq 2} (m^a_\mu)^{k-1} (m^b_\mu)^{k-1}$ as in \eref{eq:appA-noise-covariance}.
Inserting the conjugate pair $(R, \hat{R})$ exactly as in \eref{eq:appA-conjugate-insertion}, the non-condensed hidden modes factorize into the single-mode integral $\mathcal{I}_k(\hat{R})$ of \eref{eq:appA-Ik}.
With the spherical constraints \eref{eq:appC-spherical-multiplier} inserted, the visible trace is an unconstrained Gaussian integral,
\begin{equation}
    \int \prod_{a, i} d\xx^a_i \exp \bigg[ - \frac{1}{2} \sum_i \bm{\xx}_i^\top S \, \bm{\xx}_i + \beta_k \sum_{a, i} h_a^{k-1} \xx^a_i \bigg]
        = (2\pi)^{n \Nv / 2} (\det S)^{-\Nv / 2} \exp \bigg( \frac{\Nv \beta_k^2}{2} \bm{b}^\top S^{-1} \bm{b} \bigg),
    \label{eq:appC-visible-gaussian}
\end{equation}
where $\bm{\xx}_i = (\xx^a_i)_a$, $h_a \coloneq m^a_1$ is the condensed hidden mode, $\bm{b} = (h_a^{k-1})_a$, and $S \coloneq \mathrm{diag}(u^a) - \beta_k^2 R$.
Under the RS ansatz ($h_a = h$, $u^a = u$, $R^{aa} = \alpha_k r_d$, $R^{ab} = \alpha_k r$, $\hat{R}^{aa} = \hat{r}_d$, $\hat{R}^{ab} = \hat{r}$),
\begin{equation}
    S = \tilde{D} \, I_n - \alpha_k \beta_k^2 r \, \bm{1} \bm{1}^\top,
    \qquad
    \tilde{D} \coloneq u - \alpha_k \beta_k^2 (r_d - r),
\end{equation}
whose eigenvalues are $\tilde{D}$ (with degeneracy $n - 1$) and $\tilde{D} - n \alpha_k \beta_k^2 r$, and the Sherman--Morrison formula gives
\begin{equation}
    \lim_{n \to 0} \frac{1}{n} \log \det S = \log \tilde{D} - \frac{\alpha_k \beta_k^2 r}{\tilde{D}},
    \qquad
    \lim_{n \to 0} \frac{1}{n} \bm{b}^\top S^{-1} \bm{b} = \frac{h^{2k-2}}{\tilde{D}}.
    \label{eq:appC-S-RS}
\end{equation}
The stationarity conditions with respect to $r_d$ and $u$ read
\begin{equation}
    \hat{r}_d = \frac{1}{\tilde{D}} + \frac{\beta_k^2 (\alpha_k r + h^{2k-2})}{\tilde{D}^2},
    \qquad
    \frac{1}{\tilde{D}} + \frac{\beta_k^2 (\alpha_k r + h^{2k-2})}{\tilde{D}^2} = 1,
\end{equation}
so that $\hat{r}_d = 1$, and all remaining $r_d$-dependence then cancels via the shift $u = \tilde{D} + \alpha_k \beta_k^2 (r_d - r)$, in complete parallel with Model A (\eref{eq:appA-rhat-d} and below).
The stationarity condition with respect to $r$ gives
\begin{equation}
    \hat{r} = \frac{\beta_k^2 (\alpha_k r + h^{2k-2})}{\tilde{D}^2} = m^2 + \frac{\alpha_k \beta_k^2 r}{\tilde{D}^2} = q,
    \qquad
    m \coloneq \frac{\beta_k h^{k-1}}{\tilde{D}},
    \label{eq:appC-rhat-q}
\end{equation}
the Edwards--Anderson overlap of the visible variables, as in \eref{eq:appA-rhat-q}: per site, $m = \ev{\xx_i}$ is the mean of the visible Gaussian and $\alpha_k \beta_k^2 r / \tilde{D}^2 = (S^{-1})_{ab}$ ($a \neq b$) is its frozen covariance, so that $\hat{r} = \ev{\xx^a_i} \ev{\xx^b_i} + (S^{-1})_{ab}$.
For $k = 2$, the hidden factor $\mathcal{I}_2(\hat{R})$ is evaluated exactly as in \appenref{append:modelA-k2} and yields the noise term $\frac{\alpha_k}{2} \Psi_2(q)$ together with $r = \mathcal{M}_2(q)$ (\eref{eq:appA-logI2} and \eref{eq:appA-M2}).
Assembling all terms and eliminating $(h, u)$ reproduces \eref{eq:appC-reduced}, with the condensed hidden mode tied to the visible overlap by $m = \beta_k (1 - q) h^{k-1}$.
Combined with the $h$-saddle, this relation gives $m = h$ on the retrieval branch: the condensed hidden neuron equals the overlap it detects, in accordance with the general role of the hidden neurons discussed in \sref{sec:overview_of_class_h}.
For $k > 2$, this computation terminates at the same obstruction as in Model A: $\mathcal{I}_k$ admits no extensive evaluation (\appenref{append:modelA-breakdown}), and one must pass to the adiabatic formulation \eref{eq:appC-Zeff}, upon which the pair $(R, \hat{R})$ collapses onto the visible overlap, $R^{ab} \to \alpha_k \mathcal{M}_k(q_{ab})$ as in \eref{eq:appA-R-slaved}, and the computation reduces to the one performed in \appenref{append:modelC-setup}--\appenref{append:modelC-RS}.

\subsubsection{Remarks}
\label{append:modelC-remarks}

\ntextbf{Remark 1: exactness of the gauge rotation and the Gaussian replacement.}
The rotation that maps the condensed pattern to $(1, \ldots, 1)$ is an orthogonal transformation of $\R^{\Nv}$ under which both the spherical pattern ensemble and the visible measure $d\Omega(\xx)$ are invariant.
The gauge choice is therefore exact at any finite $\Nv$, just as the site-wise sign gauge of Model A.
The only asymptotic step in the pattern average is the replacement of the remaining spherical patterns by Gaussian vectors.
By rotational invariance, the exact average of any function of the overlaps $(\hat{y}^a_\mu)_a$ depends on the visible configurations only through the Gram matrix $Q$.
The finite-$\Nv$ corrections are therefore automatically functions of the order parameters, and they are suppressed by $O(\Nv^{-1})$ relative to the Gaussian leading term, leaving the RS equations unaffected.

\ntextbf{Remark 2: validity of the cumulant expansion.}
The power counting of Remark~2 in \appenref{append:modelA-remarks} applies with two modifications.
First, the overlaps are bounded, $|\hat{y}^a_\mu| \leq \sqrt{\Nv}$ by the Cauchy--Schwarz inequality with $\norm{\xii{h}{v}_\mu}_2 = \norm{\xx^a}_2 = \sqrt{\Nv}$, so the per-pattern average \eref{eq:appC-noise-factor} and all its cumulants are finite.
Note that naively replacing $\hat{y}$ by an exactly Gaussian variable \emph{inside} the exponential would produce a divergent average for $k > 2$.
The correct order of operations is to expand in cumulants first and to evaluate each cumulant by the Gaussian asymptotics.
The divergence of the naive replacement is precisely the $\mathcal{I}_k$ pathology of \appenref{append:modelA-breakdown} in another guise.
Second, the role played by $s_i^2 = 1$ in Model A (the state-independence of the first cumulant) is now played by the spherical constraint, which fixes $q_{aa} = 1$ exactly.
The third cumulant is $O(\Nv^{2 - k/2})$ by the same counting as in Model A, subextensive for even $k \geq 4$.

\ntextbf{Remark 3: the Onsager reaction field.}
The cavity argument of Remark~1 in \appenref{append:modelA-remarks} carries over with $s_i \to \xx_i$: the linear response of a soft spin to the field shift of one non-condensed mode is controlled by its single-site thermal variance, whose average $\frac{1}{\Nv} \sum_i (\ev{\xx_i^2} - \ev{\xx_i}^2) = 1 - q$ follows from the spherical constraint.
The self-feedback of a non-condensed overlap is again of relative order $\beta_k (1 - q) \Nv^{-(k-2)/2}$: marginal at $k = 2$, where its geometric resummation produces the denominator of $\mathcal{M}_2$ (equivalently, the determinant \eref{eq:appC-det} resums it exactly), and vanishing for $k > 2$, where $r = \mathcal{M}_k(q)$ is the bare Gaussian moment.
This confirms, by an argument independent of replicas, that the crosstalk moment of Model C coincides with that of Model A, as stated in the main text.

\section{Details on Model B}
\label{append:model_b_computation_details}

This appendix collects the computations behind \sref{sec:model_b}.
\appenref{append:model_b_zero_temperature} carries out the zero-temperature estimates underlying the capacity \eref{eq:modelB-capacity}.
\appenref{append:model_b_copy_representation} derives the copy representation exactly, including the hidden-sector zero mode and the shifts quoted in \sref{sec:model_b_copy_representation}, and makes the duality with the replica method quantitative.
\appenref{append:model_b_finite_temperature_copy} performs the optimization over copy configurations behind the finite-temperature phases, the retrieval branch, and the capacity of \sref{sec:model_b_finite_temperature}.
\appenref{append:complete_phase_diagram} reconstructs the phase diagram at arbitrary real temperature and settles the analytic continuation off the integer-temperature lattice.
Throughout we use the normalization of \sref{sec:model_b}, $\tv = 1$ and $\th = \coupling$, for which $\tilde{\beta} = \beta$, with Gaussian patterns at the exponential load \eref{eq:modelB-load}.
Rates of partition functions are denoted $\varphi \coloneq \lim_{\Nv \to \infty} \Nv^{-1} \log \zb$, measured relative to the Gaussian reference $\int d\xv\, e^{-\beta \norm{\xv}^2 / 2}$ as in \sref{sec:model_b_finite_temperature}, and are related to the free energy densities of the main text by $f = - T \varphi$.

\subsection{Vertex condensation and the zero-temperature capacity}
\label{append:model_b_zero_temperature}

\ntextbf{Vertex condensation.}
The Gram matrix is positive semi-definite, $x^\top G x = \norm{\xii{v}{h} x}^2 \geq 0$, so the energy $Q(\xf) \coloneq \xf^\top G \xf$ is convex on the simplex, and a convex function on a compact convex set attains its maximum at an extreme point \cite[Sec.~32]{rockafellar1970convex}.
Together with the elementary bound \eref{eq:modelB-vertex-bound}, this places the maximum at the vertex of the pattern of maximal norm, a deterministic fact independent of the pattern ensemble.
The entropic competition quoted in \sref{sec:model_b_zero_temperature} is quantified as follows.
For the uniform mixture $\xf_{\mathrm{mix}} = \frac{1}{M} \sum_{\mu \in S} e_\mu$ over a set $S$ of $M$ patterns,
\begin{equation}
    Q(\xf_{\mathrm{mix}})
        = \frac{1}{M^2} \Big[ \sum_{\mu \in S} \norm{\xii{h}{v}_\mu}^2 + \sum_{\substack{\mu \neq \nu \in S}} \xii{h}{v}_\mu \cdot \xii{h}{v}_\nu \Big]
        = \frac{\Nv}{M} + O\big(\tfrac{\sqrt{\Nv}}{M}\big),
    \label{eq:appB-mixture-energy}
\end{equation}
since the $M (M-1)$ zero-mean cross terms, each of size $O(\sqrt{\Nv})$, add up to $O(M \sqrt{\Nv})$ typically.
Hence
\begin{equation}
    \Phi(e_\mu) - \Phi(\xf_{\mathrm{mix}})
        \approx \frac{\beta}{2} \Nv \left( 1 - \frac{1}{M} \right) - \frac{\beta}{\coupling} \log M,
    \label{eq:appB-mixture-competition}
\end{equation}
which is positive unless $\log M \gtrsim \Nv$: the entropy of an interior mixture can never offset its extensive energy cost at subexponential load, and the competition becomes genuine only at the load \eref{eq:modelB-load}.
Even then the competitor is not an interior point of the simplex.
The Gibbs measure decomposes into lumps attached to the vertices,
\begin{equation}
    \zb \approx \sum_{\mu = 1}^{\Nh} Z_\mu,
    \qquad
    Z_\mu \asymp e^{\frac{\beta}{2} \norm{\xii{h}{v}_\mu}^2},
    \label{eq:appB-lump-decomposition}
\end{equation}
and the entropy is gained by spreading the measure over exponentially many, individually almost pure, lumps.
The distinction between a mixed configuration and a mixture of pure states is the same as in spin-glass theory, and the counting of the lumps is precisely the entropy evaluated below.
A final caution concerns the continuum measure: the simplex has dimension $\Nh - 1 = e^{\alpha \Nv} - 1$, so the flat measure carries volume factors of order $\log \mathrm{Vol}(\simplex^{\Nh}) \sim - \Nh \log \Nh$, doubly exponential in $\Nv$, which would overwhelm any $e^{O(\Nv)}$ energy unless they cancel exactly.
The statements above are therefore formulated either through the adiabatic saddle point \eref{eq:modelB-softmax-saddle}, for which vertex condensation is the single-term dominance of the log-sum-exp, or through the discrete copy representation of \appenref{append:model_b_copy_representation}, whose counting measure carries no volume factors.

\ntextbf{Norm statistics and freezing.}
The single-pattern rate function quoted throughout the main text follows from Cram\'er's theorem \cite{touchette2009large,dembo1998large}.
For $\norm{\xii{h}{v}_\mu}^2 = \sum_{i} (\xii{h}{v}_{\mu i})^2$ a sum of $\Nv$ i.i.d.~squared Gaussians, the cumulant generating function per component is
\begin{equation}
    \Lambda(t) \coloneq \log \E\, e^{t \xi^2} = - \frac{1}{2} \log (1 - 2t),
    \qquad
    t < \frac{1}{2},
    \label{eq:appB-chi2-cgf}
\end{equation}
and the Legendre transform $\sup_t [ t x - \Lambda(t) ]$, whose stationary point is $t^\ast = \frac12 (1 - 1/x)$, yields
\begin{equation}
    \P \left( \frac{\norm{\xii{h}{v}_\mu}^2}{\Nv} \approx x \right) \asymp e^{- \Nv I_1(x)},
    \quad
    I_1(x) = \frac{x - 1 - \log x}{2},
    \label{eq:appB-chi2-rate}
\end{equation}
the $M = 1$ case of \eref{eq:modelB-gram-rate}.
The rate $I_1$ is convex, vanishes at $x = 1$, and has the bounded derivative $I_1^\prime(x) = \frac12 (1 - 1/x) < \frac12$.
Writing $\norm{\xii{h}{v}_\mu}^2 = (1 + \varepsilon_\mu) \Nv$, the number of patterns at norm excess $\varepsilon$ is
\begin{equation}
    \mathcal{N}(\varepsilon) \asymp \Nh\, e^{- \Nv I_1(1 + \varepsilon)} = e^{\Nv [ \alpha - I_1(1 + \varepsilon) ]}.
    \label{eq:appB-level-counting}
\end{equation}
For $\alpha > I_1(1 + \varepsilon)$ the occupation numbers of independent patterns concentrate on this value ($\mathrm{Var}/\mathrm{Mean}^2 \asymp e^{-\Nv [\alpha - I_1]} \to 0$ by the second-moment method), while for $\alpha < I_1(1 + \varepsilon)$ the level is empty with high probability by Markov's inequality.
The maximal norm excess is therefore $\varepsilon_{\max}(\alpha)$ with $I_1(1 + \varepsilon_{\max}) = \alpha$, as stated in \sref{sec:model_b_finite_temperature}, and $\varepsilon_{\max} \approx 2 \sqrt{\alpha}$ for small $\alpha$.
The lump sum \eref{eq:appB-lump-decomposition} then follows by the maximal-term principle,
\begin{equation}
    \frac{1}{\Nv} \log \sum_\mu e^{\frac{\beta}{2} \norm{\xii{h}{v}_\mu}^2}
        = \frac{\beta}{2} + \max_{0 \leq \varepsilon \leq \varepsilon_{\max}} \left[ \alpha - I_1(1 + \varepsilon) + \frac{\beta}{2} \varepsilon \right].
    \label{eq:appB-lump-sum}
\end{equation}
The interior stationary point $I_1^\prime(1 + \varepsilon^\ast) = \beta/2$ gives $\varepsilon^\ast = \beta / (1 - \beta)$, which exists for $\beta < 1$ and lies in the populated range while $I_1(1 + \varepsilon^\ast) = \kappa(\beta) \leq \alpha$.
Substitution collapses \eref{eq:appB-lump-sum} to $\alpha - \frac12 \log(1 - \beta)$, in exact agreement with the annealed average $\E\, e^{\frac{\beta}{2} \chi^2_{\Nv}} = (1 - \beta)^{-\Nv/2}$ per pattern: the sum is carried by exponentially many typical terms and self-averages.
Since $I_1^\prime < \frac12$, for $\beta \geq 1$ the bracket in \eref{eq:appB-lump-sum} increases monotonically, and for $\kappa(\beta) > \alpha$ the stationary point leaves the populated range.
In either case the maximum is pinned at the boundary $\varepsilon_{\max}$, where the counting exponent vanishes and
\begin{equation}
    \frac{1}{\Nv} \log \sum_\mu e^{\frac{\beta}{2} \norm{\xii{h}{v}_\mu}^2}
        = \frac{\beta}{2} \left( 1 + \varepsilon_{\max}(\alpha) \right):
    \label{eq:appB-lump-frozen}
\end{equation}
the sum is dominated by the $O(1)$ patterns of maximal norm, and the quenched value falls below the annealed one, which diverges altogether for $\beta \geq 1$.
This is the freezing mechanism of the REM, for whose rigorous treatment see \cite{bovier2006statistical}.
These two branches are the free energies $f_{\mathrm{C}}$ and $f_{\mathrm{F}}$ of \eref{eq:modelB-branch-free-energies}, and the threshold $\beta = 1$ reappears in \appenref{append:model_b_finite_temperature_copy} as the stability criterion of the visible fluctuations.

\ntextbf{Leak sum and capacity.}
The stationary points of the effective energy \eref{eq:modelB-adiabatic} obey the fixed-point equation $\xv = \xii{v}{h} \xf^\ast(\xv)$ of \sref{sec:model_b_zero_temperature}.
For the retrieval ansatz we evaluate the fields at $\xv = \xii{h}{v}_1$ and verify self-consistency afterwards.
Conditioned on $\xii{h}{v}_1$, the overlaps $\xii{h}{v}_\mu \cdot \xii{h}{v}_1$ for $\mu \geq 2$ are exactly i.i.d.~centered Gaussians of variance $\norm{\xii{h}{v}_1}^2 \approx \Nv$, which is the statement $a_\mu = \coupling \sqrt{\Nv}\, \omega_\mu$ used in \eref{eq:modelB-leak-sum}.
The leak sum $L$ is a Boltzmann sum over the linear random energies $\coupling \sqrt{\Nv}\, \omega_\mu$, so the counting dichotomy above applies with the Gaussian rate $x^2/2$ in place of $I_1$: levels $\omega_\mu \approx x \sqrt{\Nv}$ are populated by $e^{\Nv (\alpha - x^2/2)}$ patterns up to $x_{\max} = \sqrt{2 \alpha}$ and empty beyond, which is the content of \eref{eq:modelB-leak-evaluation}, with the interior branch again matching the annealed average of $L$ and the boundary branch frozen on the $O(1)$ most aligned patterns.
Retrieval is self-consistent when $1 - \mem \leq e^{-\coupling \Nv} L \to 0$, i.e.~$\Nv^{-1} \log L < \coupling$.
On the frozen branch this reads $\coupling \sqrt{2 \alpha} < \coupling$, i.e.~$\alpha < \frac12$, while on the annealed branch it reads $\alpha + \coupling^2 / 2 < \coupling$, i.e.~$\alpha < \coupling - \coupling^2/2$.
The branch applicable at the threshold is determined by comparing $\coupling$ with $\sqrt{2 \alpha_c}$, which reduces to $\coupling \gtrless 1$, and the two conditions combine into the capacity \eref{eq:modelB-capacity}, continuous at $\coupling = 1$.
The correction to the ansatz is controlled by the same leak: $\delta \coloneq \bar{\xv} - \xii{h}{v}_1 = \sum_{\mu \geq 2} \xf^\ast_\mu \xii{h}{v}_\mu$ obeys $\norm{\delta} \leq (1 - \mem) \max_\mu \norm{\xii{h}{v}_\mu}$, exponentially small whenever the leak exponent is negative, so the fixed point survives in a neighborhood of $\xii{h}{v}_1$ throughout the retrieval region.

\ntextbf{Zero-temperature thermodynamics.}
At the retrieval saddle the effective energy \eref{eq:modelB-adiabatic} is $E(\xii{h}{v}_1) \approx \frac{\Nv}{2} - \frac{1}{\coupling} \cdot \coupling \Nv = - \frac{\Nv}{2}$, the leak contributing only $e^{-O(\Nv)}$ corrections.
The point $\xv = 0$ is also stationary, $\nabla E |_0 = - \Nh^{-1} \sum_\mu \xii{h}{v}_\mu = O(\sqrt{\Nv / \Nh})$, with energy $E(0) = - \frac{1}{\coupling} \log \Nh = - \frac{\alpha}{\coupling} \Nv$.
Typical retrieval therefore lies below this paramagnetic point only for $\alpha < \coupling / 2$, and for $\coupling \leq 1$ the window $\coupling/2 < \alpha < \coupling - \coupling^2/2$ supports retrieval only as a metastable state.
Neither of these is the true zero-temperature equilibrium, however: the retrieval state of the maximal-norm pattern has energy density $- (1 + \varepsilon_{\max})/2$, below both, and its leak condition is satisfied wherever the comparison is relevant because the signal is enhanced by the factor $1 + \varepsilon_{\max}$.
Equating it with the paramagnetic value yields the zero-temperature equilibrium boundary $2 \alpha / \coupling = 1 + \varepsilon_{\max}(\alpha)$, whose small-$\coupling$ solution is $\alpha \approx \coupling/2 + \coupling^{3/2}/\sqrt{2}$.
This is the $T \to 0$ limit of the P--F boundary derived in \appenref{append:model_b_finite_temperature_copy}, and the extreme-value enhancement is precisely the amount by which the equilibrium boundary exceeds the naive comparison $\alpha = \coupling/2$.

\subsection{Copy representation and hidden-sector corrections}
\label{append:model_b_copy_representation}

\ntextbf{Zero mode of the hidden Hessian and the shifts.}
The adiabatic evaluation \eref{eq:partition-function-adiabatic} presumes a positive-definite Hessian of the hidden Lagrangian.
For Model B this fails in exactly one direction.
At any point of the hidden space,
\begin{equation}
    \hess \big( \Lh(\xh) \big)
        = \diag(\xf) - \xf \xf^\top,
    \qquad
    \xf = \softmax(\xh),
    \label{eq:appB-hidden-hessian}
\end{equation}
and the normalization $\sum_\mu \xf_\mu = 1$ makes the all-one vector an exact zero mode,
\begin{equation}
    \big( \diag(\xf) - \xf \xf^\top \big)\, \onevec
        = \xf - \xf \sum_\mu \xf_\mu = 0.
    \label{eq:appB-zero-mode}
\end{equation}
This is not an accident of the saddle point but the gauge direction of the softmax, which is invariant under the uniform shift $\xh \to \xh + c\, \onevec$ noted in \sref{sec:model_b}.
The energy is exactly constant along $\onevec$, not merely flat to quadratic order.
The remaining spectrum is well behaved: \eref{eq:appB-hidden-hessian} is a rank-one downdate of a positive diagonal matrix, its eigenvalues interlace the softmax weights, and exactly one eigenvalue vanishes while the others remain positive.
Its pseudo-determinant (the product of the nonzero eigenvalues, denoted $\detp$) follows from the matrix determinant lemma with an $\epsilon$ regularization,
\begin{equation}
    \begin{aligned}
        \det \big( \hess + \epsilon I \big)
            &= \Big( 1 - \sum_\mu \frac{\xf_\mu^2}{\xf_\mu + \epsilon} \Big) \prod_\mu (\xf_\mu + \epsilon)  \\
            &= \big[ \Nh\, \epsilon + O(\epsilon^2) \big] \prod_\mu \xf_\mu\, \big( 1 + O(\epsilon) \big),
    \end{aligned}
    \label{eq:appB-detp-regularization}
\end{equation}
so that
\begin{equation}
    \detp \hess \big( \Lh(\xh) \big)
        = \lim_{\epsilon \to 0} \frac{\det ( \hess + \epsilon I )}{\epsilon}
        = \Nh \prod_{\mu=1}^{\Nh} \xf_\mu.
    \label{eq:appB-detp}
\end{equation}
Expanding the $\xh$ integral around the adiabatic saddle point along the Hessian eigenbasis, the flat direction contributes a divergent volume $V_0$ that is independent of $\xv$ and of the patterns.
Dividing it out and performing the remaining $\Nh - 1$ Gaussian modes yields the corrected version of \eref{eq:partition-function-adiabatic},
\begin{equation}
    \zb(\beta) / V_0
        = \left( \frac{2\pi}{\beta / \coupling} \right)^{\!(\Nh - 1)/2}
        \int_{\R^{\Nv}} d\xv\; e^{- \beta \Eb(\xv, \xh_\ast)} \left( \detp \hess ( \Lh(\xh_\ast) ) \right)^{-1/2}.
    \label{eq:appB-adiabatic-corrected}
\end{equation}
The exponent $(\Nh - 1)/2$ in place of $\Nh/2$, together with the quotient by $V_0$, is the measure prescription stated in the footnote of \sref{sec:model_b}: the flat measure $d\xf$ on the simplex is the flat measure $d\xh$ modulo the gauge orbit.

The determinant factor in \eref{eq:appB-adiabatic-corrected} is itself an exponential of the fields.
With $a \coloneq \coupling\, \xii{h}{v} \xv$ and $\xf^\ast_\nu = \softmax(a)_\nu = e^{a_\nu} / \sum_\mu e^{a_\mu}$,
\begin{equation}
    \left( \detp \hess \right)^{-1/2}
        = \Nh^{-1/2}\, e^{- \frac12 \sum_\nu a_\nu} \Big( \sum_\mu e^{a_\mu} \Big)^{\Nh / 2},
    \label{eq:appB-detp-rewrite}
\end{equation}
so the integrand of \eref{eq:appB-adiabatic-corrected} combines with the power $(\sum_\mu e^{a_\mu})^{\beta/\coupling}$ of the effective energy \eref{eq:modelB-adiabatic} into
\begin{equation}
    \exp \Big[ - \frac{\beta}{2} \norm{\xv}^2 - \frac12 \sum_\nu a_\nu \Big] \Big( \sum_\mu e^{a_\mu} \Big)^{\beta/\coupling + \Nh/2}.
    \label{eq:appB-combined-power}
\end{equation}
Whenever the combined exponent $\bar{n} \coloneq \beta/\coupling + \Nh/2$ is a positive integer, the multinomial theorem linearizes the power into a sum over $\bar{n}$-tuples, each term is Gaussian in $\xv$, and completing the square component by component gives, up to the same prefactors,
\begin{equation}
    \zb(\beta) / V_0
        \propto \sum_{\mu_1, \dots, \mu_{\bar{n}}} \exp \Big[ \frac{\coupling^2}{2 \beta} \norm{ \tilde{\xi}_{\set{\mu}} }^2 \Big],
    \quad
    \tilde{\xi}_{\set{\mu}} \coloneq - \frac12 \sum_{\nu=1}^{\Nh} \xii{h}{v}_\nu + \sum_{j=1}^{\bar{n}} \xii{h}{v}_{\mu_j}.
    \label{eq:appB-shifted-copy}
\end{equation}
Relative to the representation \eref{eq:modelB-copy-expansion}, which coincides with the visible-only computation of \cite{ota2023attention}, the hidden-sector fluctuations thus produce exactly the two corrections quoted in \sref{sec:model_b_copy_representation}: the exponent shift $\beta/\coupling \to \beta/\coupling + \Nh/2$ and the additive pattern shift $- \frac12 \sum_\nu \xii{h}{v}_\nu$.
Two features of \eref{eq:appB-shifted-copy} deserve emphasis at exponential load.
First, positivity of the temperature imposes $\beta / \coupling = \bar{n} - \Nh/2 > 0$: the admissible integers satisfy $\bar{n} > \Nh / 2$, so the small copy numbers, in particular $\bar{n} = 1$, are excluded once $\Nh$ is large.
Second, the shift is not a perturbation: the components of $\frac12 \sum_\nu \xii{h}{v}_\nu$ are of order $\sqrt{\Nh}$, so at $\Nh = e^{\alpha \Nv}$ the shifted sum is dominated by the correction term itself.
The analysis of the main text is therefore formulated for the adiabatic partition function proper, the leading saddle-point term for which the unshifted expansion \eref{eq:modelB-copy-expansion} is an exact identity at $n = \beta/\coupling \in \Z_{>0}$.
The corrections \eref{eq:appB-shifted-copy} constitute the complete Gaussian fluctuation content of the hidden sector around that limit, and their consistent treatment at exponential load remains open (see the remarks closing \appenref{append:complete_phase_diagram}).

\ntextbf{Alignment versus dispersion.}
Reversing the Gaussian integration that produced \eref{eq:modelB-copy-expansion} shows that, conditioned on a copy configuration, the visible state is Gaussian around the centroid of the selected patterns,
\begin{equation}
    \xv \,\big|\, \set{\mu_j}
        \;\sim\; \mathcal{N} \Big( \frac{1}{n} \sum_{j=1}^n \xii{h}{v}_{\mu_j},\; \beta^{-1} I_{\Nv} \Big),
    \label{eq:appB-conditional-gaussian}
\end{equation}
which is the parallel-query picture of \sref{sec:model_b_copy_representation} and will be used repeatedly below to read off visible observables.
Expanding the energy of \eref{eq:modelB-copy-expansion},
\begin{equation}
    \norm{ \sum_j \xii{h}{v}_{\mu_j} }^2
        = \sum_j \norm{\xii{h}{v}_{\mu_j}}^2
        + \sum_{j \neq l} \Big[ \Nv\, \delta_{\mu_j \mu_l} + O(\sqrt{\Nv}) \Big],
    \label{eq:appB-potts}
\end{equation}
the copies form an $n$-site system with $\Nh$ states per site, a ferromagnetic gain of order $\Nv$ for every pair of aligned copies, and random couplings of order $\sqrt{\Nv}$ otherwise.
The zero-temperature equilibrium of \appenref{append:model_b_zero_temperature} can be recovered from this picture by pure counting.
The fully aligned configurations ($\mu_j \equiv \mu$, with $\Nh$ choices) carry the exponent $\frac{\coupling}{2n} n^2 \Nv + \alpha \Nv = ( \frac{\beta}{2} + \alpha ) \Nv$, while the fully dispersed ones ($n_\mu \in \set{0, 1}$, with $\approx e^{n \alpha \Nv}$ choices) retain only the diagonal energy, $( \frac{\coupling}{2} + n \alpha ) \Nv$.
The difference,
\begin{equation}
    \Big( \frac{\beta}{2} + \alpha \Big) - \Big( \frac{\coupling}{2} + n \alpha \Big)
        = (n - 1) \Big( \frac{\coupling}{2} - \alpha \Big),
    \label{eq:appB-alignment-dispersion}
\end{equation}
shows that alignment is favored precisely for $\alpha < \coupling/2$, reproducing the zero-temperature equilibrium threshold of \appenref{append:model_b_zero_temperature} without any reference to the $f$ representation.
The frozen phase corresponds, in the same language, to the co-condensation of all copies onto the few patterns of extreme norm.

\ntextbf{Legendre duality and one-step RSB.}
The qualitative dictionary of \sref{sec:model_b_copy_representation} becomes a precise convex duality.
On the replica side one computes $\EE{\xi}{(\zb)^r}$ at a formal replica number $r$.
The patterns enter the replicated exponent through $\sum_{i} \xi_i^\top \Theta\, \xi_i$ with the tilting matrix $\Theta \coloneq \frac{\beta}{2} \sum_{a=1}^r \xf^a (\xf^a)^\top$, where $\xi_i \in \R^{\Nh}$ collects the $i$-th components of all patterns, and the Gaussian average per visible component is the matrix version of \eref{eq:appB-chi2-cgf},
\begin{equation}
    \E\, e^{\xi^\top \Theta \xi}
        = \det ( I - 2 \Theta )^{-1/2}
        \eqcolon e^{\Lambda(\Theta)},
    \label{eq:appB-matrix-cgf}
\end{equation}
valid for $I - 2\Theta \succ 0$.
By Sylvester's identity, $\det(I_{\Nh} - \beta \sum_a \xf^a (\xf^a)^\top) = \det(I_r - \beta R)$ with the replica overlap matrix $R^{ab} = \xf^a \cdot \xf^b$, so the disorder average generates the effective replica coupling $- \frac{\Nv}{2} \log \det (I_r - \beta R)$.
On the copy side the disorder statistics appear instead as the counting rate $I_M(Q)$ of \eref{eq:modelB-gram-rate}.
The two are a Legendre--Fenchel pair \cite{rockafellar1970convex,touchette2009large},
\begin{equation}
    \begin{aligned}
        I_M(Q) &= \sup_\Theta \big[ \Tr (\Theta Q) - \Lambda(\Theta) \big],  \\
        \Lambda(\Theta) &= \sup_Q \big[ \Tr (\Theta Q) - I_M(Q) \big],
    \end{aligned}
    \label{eq:appB-legendre-pair}
\end{equation}
with stationarity $Q^\ast = (I - 2\Theta)^{-1}$: the optimal Gram matrix is the covariance of the exponentially tilted Gaussian ensemble.
The annealed branches of \appenref{append:model_b_finite_temperature_copy} realize the second line of \eref{eq:appB-legendre-pair} at the rank-one tilts generated by the copy energy, and the conjugate order parameter that a replica computation introduces through the Fourier representation of $\delta$ functions is precisely the tilting matrix $\Theta$: the field of the exponential change of measure that selects which distortion of the pattern statistics dominates.
Freezing is the point where the two routes diverge.
The Gaussian average \eref{eq:appB-matrix-cgf} runs over an unbounded ensemble, whereas only $e^{\alpha \Nv}$ patterns exist.
The correct variational problem is the Fenchel transform constrained to the populated set,
\begin{equation}
    \sup_{Q}\; \big[ \Tr(\Theta Q) - I_M(Q) \big]
    \quad \text{subject to} \quad
    I_M(Q) \leq M \alpha .
    \label{eq:appB-constrained-fenchel}
\end{equation}
When the constraint is active, the Karush--Kuhn--Tucker condition with multiplier $\eta \geq 0$ reads $\Theta = (1 + \eta) \nabla I_M(Q)$, i.e.
\begin{equation}
    Q = \Big( I - \frac{2 \Theta}{1 + \eta} \Big)^{-1}:
    \label{eq:appB-kkt}
\end{equation}
the tilt is renormalized by the factor $(1 + \eta)^{-1} \in (0, 1]$.
In the scalar case relevant to the frozen phase (tilt $\beta/2$, boundary condition $I_1(Q) = \alpha$, i.e.~$Q = 1 + \varepsilon_{\max}$), \eref{eq:appB-kkt} gives $\beta / (1 + \eta) = \varepsilon_{\max} / (1 + \varepsilon_{\max}) = \beta_f$, so that
\begin{equation}
    \frac{1}{1 + \eta} = \frac{\beta_f}{\beta} = \frac{T}{T_f},
    \label{eq:appB-parisi}
\end{equation}
precisely the temperature dependence of the one-step Parisi parameter of the REM \cite{derrida1981random,mezard2009information,bovier2006statistical}: the frozen phase is pinned at the edge of the populated set and behaves as if held at the freezing temperature, which is the origin of the temperature independence of $f_{\mathrm{F}}$.

The identification of the freezing prescription with one-step RSB can be checked exactly on the lump sum \eref{eq:appB-lump-sum}, which is a REM in its own right.
Computing $\E \tilde{Z}^r$ for $\tilde{Z} = \sum_\mu e^{\frac{\beta}{2} \norm{\xii{h}{v}_\mu}^2}$ with the one-step ansatz, $r/x$ blocks of $x$ replicas co-condensing on distinct patterns, each block contributes $e^{\alpha \Nv}\, \E\, e^{\frac{x \beta}{2} \chi^2_{\Nv}} = e^{\Nv [ \alpha - \frac12 \log (1 - x \beta) ]}$, so per replica
\begin{equation}
    \varphi_{\mathrm{1RSB}}(x)
        = \frac{1}{x} \Big[ \alpha - \frac12 \log (1 - x \beta) \Big].
    \label{eq:appB-minirem}
\end{equation}
Extremizing over the block size continued to $x \in (0, 1]$ yields $\alpha = \kappa(x \beta)$, whose solution is $x^\ast \beta = \beta_f$, i.e.~$x^\ast = T/T_f$, in agreement with the multiplier \eref{eq:appB-parisi}.
Substituting back, $\alpha - \frac12 \log(1 - \beta_f) = \alpha + \frac12 \log (1 + \varepsilon_{\max}) = \varepsilon_{\max}/2$ and
\begin{equation}
    \varphi_{\mathrm{1RSB}}(x^\ast)
        = \frac{\beta}{\beta_f} \cdot \frac{\varepsilon_{\max}}{2}
        = \frac{\beta}{2} \big( 1 + \varepsilon_{\max} \big),
    \label{eq:appB-minirem-value}
\end{equation}
which is the frozen counting value \eref{eq:appB-lump-frozen} exactly.
For $\beta < \beta_f$ the stationary point falls at $x > 1$, the extremum over the physical interval is the endpoint $x = 1$, and the RS evaluation returns the annealed branch, again in agreement.
For the sign conventions of the $r \to 0$ extremization see \cite{mezard2009information}.
The one-step structure also fixes the weight statistics of the frozen phase: the lump weights $w_\mu = e^{\frac{\beta}{2}\norm{\xii{h}{v}_\mu}^2} / \tilde{Z}$ follow a Ruelle point process of parameter $x^\ast = T/T_f$ \cite{ruelle1987mathematical,mezard1987spin}, with participation ratio $\E \sum_\mu w_\mu^2 = 1 - T/T_f$.
Since the $w_\mu$ are the attention weights of the $f$ representation, this is a direct quantitative prediction for the attention statistics in the F phase.

\ntextbf{Clone method.}
Finally, the copy representation realizes the clone method of Monasson \cite{monasson1995structural} with a physical clone number.
The single-group sector of \appenref{append:model_b_finite_temperature_copy} (all $n$ copies on one pattern) has the rate
\begin{equation}
    \varphi
        = \max_{\varepsilon} \Big[ \underbrace{\alpha - I_1(1 + \varepsilon)}_{\text{complexity } \Sigma(\varepsilon)} + n \cdot \underbrace{\frac{\coupling}{2} (1 + \varepsilon)}_{\text{per-clone gain}} \Big],
    \label{eq:appB-clone}
\end{equation}
which is exactly the clone free energy: the choice of the occupied state (pattern) is counted once for the whole clone collective, while the energy is paid once per clone.
By the envelope theorem, $\partial \varphi / \partial n = \frac{\coupling}{2} (1 + \varepsilon^\ast)$ reads off the occupied level and $\varphi - n\, \partial \varphi / \partial n = \Sigma(\varepsilon^\ast)$ the complexity, so the clone-number derivative extracts the configurational entropy exactly as in the structural-glass application.
The distinctive feature of Model B is that the clone number $n = \beta / \coupling$ is not an auxiliary parameter but the physical temperature itself.
Scanning $n$ through real values is scanning the temperature, and the analytic control of this continuation is supplied by the generalized binomial expansion of \appenref{append:complete_phase_diagram}.

\subsection{Finite-temperature analysis in the copy representation}
\label{append:model_b_finite_temperature_copy}

\ntextbf{Counting lemma.}
We first derive the counting statement of \sref{sec:model_b_finite_temperature}.
Fix $M$ distinct indices $\mu_1, \dots, \mu_M$.
For each visible component $i$ the vector $x_i \coloneq (\xii{h}{v}_{\mu_1 i}, \dots, \xii{h}{v}_{\mu_M i})^\top$ is standard Gaussian in $\R^M$, i.i.d.~over $i$, and the normalized Gram matrix is the empirical covariance $\hat{Q} = \frac{1}{\Nv} \sum_i x_i x_i^\top$.
Its cumulant generating function per component is the matrix average \eref{eq:appB-matrix-cgf}, and Cram\'er's theorem gives $\P(\hat{Q} \approx Q) \asymp e^{- \Nv I_M(Q)}$ with the Legendre transform of \eref{eq:appB-legendre-pair}: the stationary point $\Theta^\ast = \frac12 (I - Q^{-1})$ yields $\Tr(\Theta^\ast Q) = \frac12 \Tr (Q - I)$ and $\Lambda(\Theta^\ast) = \frac12 \log \det Q$, which yields the rate \eref{eq:modelB-gram-rate}.
Multiplying by the $e^{M \alpha \Nv}$ choices of the indices, and repeating the second-moment and Markov arguments of \appenref{append:model_b_zero_temperature}, the number of $M$-tuples with Gram matrix $Q$ concentrates on $e^{\Nv [ M \alpha - I_M(Q) ]}$ when the exponent is positive and vanishes with high probability otherwise.
In the eigenbasis of $Q$ the rate decomposes as
\begin{equation}
    I_M(Q) = \sum_{a=1}^{M} I_1(\Lambda_a),
    \label{eq:appB-rate-eigen}
\end{equation}
with $\Lambda_a$ the eigenvalues of $Q$, and this decomposition organizes every computation below.

\ntextbf{Partition classes and the bulk branches.}
Following \sref{sec:model_b_finite_temperature}, classify the copy configurations by the partition of the $n$ copies over distinct patterns.
Consider first the symmetric classes: $M$ groups of equal size $n / M$, each occupying one of $M$ distinct patterns (the mixed classes relevant for retrieval are treated below, and the remaining channels in \appenref{append:complete_phase_diagram}).
The number of ways to distribute the copies is subexponential in $\Nv$, and the weight in \eref{eq:modelB-copy-expansion} depends on the configuration only through the Gram matrix $Q$ of the occupied patterns, with $\norm{\sum_j \xii{h}{v}_{\mu_j}}^2 = (n/M)^2\, \onevec^\top Q\, \onevec\, \Nv$, so the class exponent per visible neuron is
\begin{equation}
    \varphi[Q]
        = M \alpha - I_M(Q) + \frac{\beta}{2 M^2}\, \onevec^\top Q\, \onevec .
    \label{eq:appB-class-exponent}
\end{equation}
The energy couples only to the coherent quadratic form $\onevec^\top Q\, \onevec$.
Writing $w_a \coloneq (\onevec \cdot e_a)^2 / M$ for the weights of the eigenvectors of $Q$ in the coherent direction ($\sum_a w_a = 1$), so that $\onevec^\top Q \onevec = M \sum_a w_a \Lambda_a$, the convexity and positivity of $I_1$ give
\begin{equation}
    I_M(Q)
        = \sum_a I_1(\Lambda_a)
        \geq \sum_a w_a I_1(\Lambda_a)
        \geq I_1 \Big( \sum_a w_a \Lambda_a \Big),
    \label{eq:appB-jensen}
\end{equation}
with equality if and only if $\onevec / \sqrt{M}$ is an eigenvector and all remaining eigenvalues lie at the minimum of $I_1$, i.e.~at $1$.
The symmetric form $Q = (q_d - q) I + q\, \onevec \onevec^\top$ with transverse eigenvalue $\Lambda_2 = q_d - q = 1$ is therefore not an ansatz but the exact maximizer within each class, and the exponent reduces to one variable, the coherent eigenvalue $\Lambda_1$,
\begin{equation}
    \varphi(M, \Lambda_1)
        = M \alpha - I_1(\Lambda_1) + \frac{\beta}{2 M} \Lambda_1.
    \label{eq:appB-symmetric-family}
\end{equation}
For $M = 1$ this is precisely the lump sum \eref{eq:appB-lump-sum}, of which \eref{eq:appB-symmetric-family} is the generalization to collectively occupied patterns.
The maximization over $\Lambda_1$ repeats the annealed/frozen dichotomy of \appenref{append:model_b_zero_temperature}.
The interior stationary point $I_1^\prime(\Lambda_1^\ast) = \beta/2M$ gives $\Lambda_1^\ast = (1 - \beta/M)^{-1}$, which exists for $\beta < M$.
At the stationary point the identity $\frac{\beta}{2M} \Lambda_1^\ast - \frac12 (\Lambda_1^\ast - 1) = 0$ collapses the exponent to
\begin{equation}
    \varphi_{\mathrm{ann}}(M)
        = M \alpha - \frac12 \log \Big( 1 - \frac{\beta}{M} \Big),
    \qquad
    M \alpha \geq \kappa \Big( \frac{\beta}{M} \Big),
    \label{eq:appB-annealed-branch}
\end{equation}
in exact agreement with the annealed average of the class partition function (the rank-one tilt has the single nontrivial eigenvalue $\beta/2M$ along $\onevec$, so the determinant in \eref{eq:appB-matrix-cgf} is $(1 - \beta/M)^{-1/2}$): the annealed branch is the self-averaging regime.
When the stationary point violates the counting constraint, or when $\beta \geq M$, the maximum is pinned at the boundary of the populated set, where the counting exponent vanishes.
Since $I_1$ is convex with its minimum at $\Lambda = 1$, the counting condition $I_1(\Lambda) = M \alpha$ has one root below and one above unity, and the exponent \eref{eq:appB-symmetric-family} increases with $\Lambda_1$ throughout the populated range, so the boundary is the upper root $\Lambda_+ > 1$, the largest coherent eigenvalue carried by a populated class:
\begin{equation}
    \varphi_{\mathrm{froz}}(M)
        = \frac{\beta}{2 M} \Lambda_+,
    \qquad
    I_1(\Lambda_+) = M \alpha,
    \quad
    \Lambda_+ > 1 .
    \label{eq:appB-frozen-branch}
\end{equation}
At $M = 1$ one has $\Lambda_+ = 1 + \varepsilon_{\max}(\alpha)$, and \eref{eq:appB-frozen-branch} reduces to \eref{eq:appB-lump-frozen}.
The optimization over $M$ is now elementary.
On the annealed branch, $d\varphi_{\mathrm{ann}} / dM = \alpha - \beta / [2 M (M - \beta)]$ tends to $-\infty$ as $M \to \beta^+$ and to $\alpha > 0$ at large $M$: the branch has an interior minimum in $M$, so its maximum is attained at the endpoints, $M = n$ (all copies dispersed) or $M = 1$ (all copies aligned).
On the frozen branch, differentiating the constraint in \eref{eq:appB-frozen-branch} gives
\begin{equation}
    \frac{d \varphi_{\mathrm{froz}}}{d M}
        = - \frac{\beta}{2 M^2}\, \frac{\Lambda_+ \log \Lambda_+}{\Lambda_+ - 1} < 0,
    \label{eq:appB-frozen-monotone}
\end{equation}
so freezing always prefers full alignment, $M = 1$.
Three bulk branches survive.
For $M = n$ (using $\beta / n = \coupling$), $\varphi_{\mathrm{P}} = n \alpha - \frac12 \log (1 - \coupling)$.
For $M = 1$ annealed, $\varphi_{\mathrm{C}} = \alpha - \frac12 \log (1 - \beta)$ with validity $\beta < 1$ and $\alpha \geq \kappa(\beta)$.
For $M = 1$ frozen, $\varphi_{\mathrm{F}} = \frac{\beta}{2} (1 + \varepsilon_{\max})$.
Through $f = - T \varphi$ these are the free energies \eref{eq:modelB-branch-free-energies}.
The entropy of the C branch is $s_{\mathrm{C}} = - \partial f_{\mathrm{C}} / \partial T = \alpha - \kappa(\beta)$, which vanishes exactly on the freezing line \eref{eq:modelB-freezing-line}.
There $\beta_f = \varepsilon_{\max} / (1 + \varepsilon_{\max})$, and substituting into either branch gives the common value $\varphi = \varepsilon_{\max}/2$, so C and F connect continuously, the REM freezing scenario in the form used in \appenref{append:model_b_copy_representation}.
By the conditional Gaussian \eref{eq:appB-conditional-gaussian}, the F state is the complete retrieval state of the maximal-norm pattern, as asserted in the main text.
The P state, by contrast, is a weak thermal condensate: the tilted Gram matrix of the dispersed patterns follows from the stationary covariance $Q^\ast = (I - 2\Theta)^{-1}$ by the Sherman--Morrison formula, $q^\ast = \coupling^2 T / (1 - \coupling)$ off the diagonal, so the centroid in \eref{eq:appB-conditional-gaussian} carries $\norm{\frac1n \sum_j \xii{h}{v}_{\mu_j}}^2 / \Nv = \Lambda_1^\ast / n = \coupling T / (1 - \coupling)$, and adding the thermal variance $T$ of the Gaussian fluctuations,
\begin{equation}
    \frac{\ev{\norm{\xv}^2}}{\Nv}
        = \frac{\coupling T}{1 - \coupling} + T
        = \frac{T}{1 - \coupling},
    \label{eq:appB-P-condensate}
\end{equation}
the value quoted in \sref{sec:model_b_finite_temperature}: even the paramagnet weakly aligns the selected patterns ($q^\ast > 0$), a linear-response enhancement that vanishes as $T \to 0$.
Finally, the existence conditions unify: a symmetric class has attention weights $1/M$ on $M$ patterns, hence inverse participation ratio $\norm{\xf}^2 = 1/M$, and the annealed condition $\beta < M$ is $\beta \norm{\xf}^2 < 1$.
For P this reads $\coupling < 1$ and for C it reads $\beta < 1$, and the instability at $\beta \norm{\xf}^2 = 1$ is the divergence of the visible Gaussian integral in \eref{eq:modelB-f-representation}, i.e.~the softening of the visible fluctuations by concentrated attention.
For $\coupling \geq 1$ the paramagnetic branch does not exist.

\ntextbf{Retrieval branch.}
Condition on the retrieved pattern $\xii{h}{v}_1$, of typical norm, and consider the mixed classes of \sref{sec:model_b_finite_temperature}: $n \mem$ copies on $\xii{h}{v}_1$ and $N^\prime \coloneq n (1 - \mem)$ copies dispersed on distinct patterns $\mu \neq 1$, with overlaps $c_j = \xii{h}{v}_{\mu_j} \cdot \xii{h}{v}_1 / \Nv$ and mutual Gram matrix $Q^\prime$.
The energy depends only on $\sum_j c_j$ and $\onevec^\top Q^\prime \onevec$, so the Jensen argument \eref{eq:appB-jensen}, supplemented by the Cauchy--Schwarz inequality for the $c_j$, again makes the uniform and symmetric form ($c_j = c$, $Q^\prime = (q_d - q) I + q \onevec \onevec^\top$) exact within the family.
The counting rate conditioned on $\xii{h}{v}_1$ follows from the longitudinal--transverse decomposition $\xii{h}{v}_{\mu_j} = \ell_j\, e_1 + \xi^\perp_j$ with $e_1 = \xii{h}{v}_1 / \norm{\xii{h}{v}_1}$.
The longitudinal components are scalar unit Gaussians constrained to $\ell_j = c \sqrt{\Nv}$, at rate $c^2/2$ each, while the transverse vectors are i.i.d.~standard Gaussian in the orthogonal complement with mutual overlaps $\xi_j^\perp \cdot \xi_l^\perp / \Nv = Q^\prime_{jl} - c^2$, at rate $I_{N^\prime}(Q^\prime - c^2 \onevec \onevec^\top)$.
Adding the two, the longitudinal cost cancels against the trace shift,
\begin{equation}
    \begin{aligned}
        I_{\mathrm{cond}}
        &= \frac{N^\prime c^2}{2} + I_{N^\prime} \big( Q^\prime - c^2 \onevec \onevec^\top \big)  \\
        &= \frac12 \Big[ \Tr Q^\prime - \log \det \big( Q^\prime - c^2 \onevec \onevec^\top \big) - N^\prime \Big],
    \end{aligned}
    \label{eq:appB-conditional-rate}
\end{equation}
so only the Schur complement survives in the log-determinant.
Equivalently, apply \eref{eq:modelB-gram-rate} to the full $(N^\prime + 1)$-tuple including $\xii{h}{v}_1$ and use $\det Q_{\mathrm{full}} = \det (Q^\prime - c^2 \onevec \onevec^\top)$ together with the vanishing conditioning cost $I_1(1) = 0$.
In the symmetric form the eigenvalues of $Q^\prime - c^2 \onevec\onevec^\top$ are $\Lambda_1^\prime - N^\prime c^2$ (coherent, $\Lambda_1^\prime = q_d + (N^\prime - 1) q$) and $\Lambda_2^\prime = q_d - q$ (transverse), so
\begin{equation}
    I_{\mathrm{cond}}
        = (N^\prime - 1)\, I_1(\Lambda_2^\prime)
        + \frac12 \Big[ \Lambda_1^\prime - \log \big( \Lambda_1^\prime - N^\prime c^2 \big) - 1 \Big],
    \label{eq:appB-conditional-rate-eigen}
\end{equation}
where the trace carries $\Lambda_1^\prime$ but the logarithm its Schur shift: the mismatch is the imprint of the longitudinal cost.
The energy exponent follows from $\norm{n \mem\, \xii{h}{v}_1 + \sum_j \xii{h}{v}_{\mu_j}}^2 / \Nv = (n\mem)^2 + 2 n \mem N^\prime c + N^\prime \Lambda_1^\prime$, multiplied by $\coupling / 2n$:
\begin{equation}
    \frac{\beta}{2} \mem^2
    + \beta \mem (1 - \mem)\, c
    + \frac{\coupling (1 - \mem)}{2} \Lambda_1^\prime .
    \label{eq:appB-retrieval-energy}
\end{equation}
The coefficient of the last term is $\coupling$, not $\beta$: the coherent enhancement of a dispersed copy is that of a single copy, and this asymmetry between the condensed group and the leak drives the entire structure below.
Assembling the selection entropy $N^\prime \alpha$, the rate \eref{eq:appB-conditional-rate-eigen}, and the energy \eref{eq:appB-retrieval-energy}, and introducing $A \coloneq (\Lambda_1^\prime - N^\prime c^2)^{-1}$, the stationarity conditions are elementary.
The transverse eigenvalue carries no energy, so $I_1^\prime(\Lambda_2^\prime) = 0$ gives $\Lambda_2^\prime = 1$.
The $c$ derivative balances $- N^\prime c A$ from the logarithm against $\beta \mem (1 - \mem) = \coupling \mem N^\prime$ from the cross term, and the $\Lambda_1^\prime$ derivative balances $-\frac12 + \frac{A}{2}$ against $\coupling (1 - \mem)/2$.
Hence
\begin{equation}
    \Lambda_2^\prime = 1,
    \qquad
    A = 1 - \coupling (1 - \mem),
    \qquad
    c = \frac{\coupling \mem}{A}.
    \label{eq:appB-retrieval-stationarity}
\end{equation}
Substituting back, the $\Lambda_1^\prime$ terms combine into $- \frac{A}{2} \Lambda_1^\prime = - \frac12 - \frac{A}{2} N^\prime c^2$, the constant cancels, the two $c$-dependent terms add to $+ \frac{n \coupling^2}{2} \mem^2 (1 - \mem) / A$, and the identity $A + \coupling (1 - \mem) = 1$ collapses the sum with the condensate self-energy into a single term:
\begin{equation}
    \varphi_{\mathrm{R}}(\mem)
        = \frac{\beta}{\coupling} (1 - \mem)\, \alpha
        - \frac12 \log A
        + \frac{\beta}{2} \frac{\mem^2}{A},
    \label{eq:appB-landau}
\end{equation}
which is the Landau function \eref{eq:modelB-landau-function} through $f_{\mathrm{R}} = - T \varphi_{\mathrm{R}}$, with the endpoint values $\varphi_{\mathrm{R}}(0) = \varphi_{\mathrm{P}}$ and $\varphi_{\mathrm{R}}(1) = \beta/2$ quoted in the main text.
The visible overlap follows from the centroid \eref{eq:appB-conditional-gaussian},
\begin{equation}
    m = \frac{\xv \cdot \xii{h}{v}_1}{\Nv}
      = \mem + (1 - \mem)\, c
      = \frac{\mem}{A},
    \qquad
    c = \coupling m,
    \label{eq:appB-overlap}
\end{equation}
where the second form uses $A + \coupling(1 - \mem) = 1$ again.
The relation $c = \coupling m$ identifies the stationary overlap of the destinations as a linear response to the retrieval field, and the first form of \eref{eq:appB-overlap} is a self-consistent loop, $m = \mem + \coupling (1 - \mem)\, m$: the leak amplifies the visible overlap by the geometric series of the loop gain $\coupling (1 - \mem)$, and the positivity $A > 0$ of the loop stiffness is once more the criterion $\beta \norm{\xf}^2 < 1$, evaluated on the leak weights $\norm{\xf_{\mathrm{leak}}}^2 = (1 - \mem)/n$.
For later use we record the stationarity of \eref{eq:appB-landau} in $\mem$.
Using $1/A = (1 - \coupling m)/(1 - \coupling)$, the condition $\partial_\mem \varphi_{\mathrm{R}} = 0$ closes in the visible overlap alone,
\begin{equation}
    \alpha
        = F_T(m)
        \coloneq \coupling m \Big( 1 - \frac{\coupling m}{2} \Big)
        - \frac{\coupling^2 T}{2}\, \frac{1 - \coupling m}{1 - \coupling},
    \label{eq:appB-saddle-eq}
\end{equation}
and $F_T$ is strictly increasing on $m \in [0, 1]$ for $\coupling < 1$, so the interior stationary point $m^\dagger$ is unique.
It is the barrier of \sref{sec:model_b_finite_temperature}, with the zero-temperature root $\coupling m^\dagger = 1 - \sqrt{1 - 2\alpha}$.
The continuum endpoint criterion, $\alpha < F_T(1) = \coupling - \frac{\coupling^2}{2} (1 + T)$, is what a smooth treatment of $\mem$ would predict for the spinodal, and the next paragraph corrects it.

\ntextbf{One-copy defection and the spinodal.}
The attention weight moves on the lattice $\mem = 1 - k / n$, with spacing $\coupling T$ that is not small at finite temperature, so the local stability of retrieval is decided by the nearest sector, $\varphi_{\mathrm{R}}(1) \gtrless \varphi_{\mathrm{R}}(1 - 1/n)$, not by the endpoint slope.
We evaluate the one-copy defection directly, which also serves as an independent check of \eref{eq:appB-landau}.
Move one copy from $\xii{h}{v}_1$ to a pattern $\mu$ with overlap $x \coloneq \xii{h}{v}_\mu \cdot \xii{h}{v}_1 / \Nv$ and norm $\norm{\xii{h}{v}_\mu}^2 / \Nv = 1 + \varepsilon$.
Conditioned on $\xii{h}{v}_1$, the pair rate function is the $M = 2$ case of \eref{eq:modelB-gram-rate},
\begin{equation}
    I(x, \varepsilon) = \frac{x^2}{2} + I_1 \big( 1 + \varepsilon - x^2 \big),
    \label{eq:appB-defection-rate}
\end{equation}
the longitudinal Gaussian cost plus the transverse norm cost.
Note that conditioning on the overlap $x$ makes the typical norm excess $x^2$, at no additional cost.
The energy gain of the defection is
\begin{equation}
    \begin{aligned}
        \Delta(x, \varepsilon)
        &= \frac{\coupling}{2n} \Big[ (n-1)^2 + 2 (n-1) x + 1 + \varepsilon - n^2 \Big]  \\
        &= \tilde{\coupling}\, (x - 1) + \frac{\coupling^2 T}{2}\, \varepsilon,
        \qquad
        \tilde{\coupling} \coloneq \coupling (1 - \coupling T),
    \end{aligned}
    \label{eq:appB-defection-gain}
\end{equation}
an exchange term, the loss of alignment with the $n - 1$ copies left behind (the defector does not interact with itself, hence the thinning factor $1 - \coupling T$), and a norm term, an energy channel of order $T$ that favors defection toward patterns of large norm.
The defection sector carries the exponent $\sup_{I \leq \alpha} [ \alpha - I(x, \varepsilon) + \Delta(x, \varepsilon) ]$.
In the annealed regime the interior stationary point is, in the variable $y \coloneq 1 + \varepsilon - x^2$,
\begin{equation}
    y^\ast = \frac{1}{1 - \coupling^2 T},
    \qquad
    x^\ast = \frac{\tilde{\coupling}}{1 - \coupling^2 T} = \frac{\coupling (1 - \coupling T)}{1 - \coupling^2 T},
    \label{eq:appB-defection-saddle}
\end{equation}
and collecting the terms (the $1/n^2$ contributions cancel exactly) gives
\begin{equation}
    \Delta \varphi_{\mathrm{def}}
        = \alpha
        - \frac{\coupling ( 2 - \coupling - \coupling T )}{2 (1 - \coupling^2 T)}
        - \frac12 \log \big( 1 - \coupling^2 T \big),
    \label{eq:appB-defection-exponent}
\end{equation}
which coincides exactly with $\varphi_{\mathrm{R}}(1 - 1/n) - \varphi_{\mathrm{R}}(1)$ computed from \eref{eq:appB-landau}: the collective derivation and the single-defection computation, which optimizes the destination geometry independently, agree term by term.
Retrieval is locally stable while $\Delta \varphi_{\mathrm{def}} < 0$, which is the spinodal \eref{eq:modelB-spinodal}.
Its small-$T$ expansion, $\alpha_c = \coupling - \frac{\coupling^2}{2} - \frac{\coupling^2 T}{2} [ 1 + (1 - \coupling)^2 ] + O(T^2)$, lies strictly below the continuum criterion below \eref{eq:appB-saddle-eq} except at $\coupling = 1$: the continuum treatment resolves barriers thinner than one attention quantum and therefore overestimates the capacity at any $T > 0$, as stated in the main text.
The defection saddle \eref{eq:appB-defection-saddle} also encodes the physics: the optimal destination is slightly aligned with the signal ($x^\ast > 0$, the one-copy version of the linear response $c = \coupling m$) and slightly long ($y^\ast > 1$, the norm channel opened by the temperature).

\ntextbf{Frozen ceiling.}
The annealed defection assumed that destinations with the optimal $(x^\ast, \varepsilon^\ast)$ exist, i.e.~$I(x^\ast, \varepsilon^\ast) \leq \alpha$.
As $T \to 0$, $I(x^\ast, \varepsilon^\ast) \to \coupling^2/2$, so for $\alpha \lesssim \coupling^2 / 2$ (sharp attention) they do not, and the supremum is attained on the boundary $I = \alpha$: the defection freezes onto the most favorable pattern actually present, the one-copy analogue of the frozen branch of \eref{eq:modelB-leak-evaluation}.
On the boundary the counting term vanishes and one maximizes $\Delta(x, \varepsilon)$ alone subject to $I(x, \varepsilon) = \alpha$.
The Lagrange conditions, $\partial_x$: $\tilde{\coupling} = \eta\, x / y$ and $\partial_\varepsilon$: $\frac{\coupling^2 T}{2} = \frac{\eta}{2} ( 1 - 1/y )$, combine into
\begin{equation}
    y - 1 = \frac{\coupling T\, x}{1 - \coupling T}:
    \label{eq:appB-ceiling-lagrange}
\end{equation}
the optimal extreme destination shifts toward norm excess as $T$ grows, and returns to the conditionally typical norm ($y \to 1$) at $T = 0$.
Since $I_1(y) = O((y-1)^2)$, the transverse cost enters the constraint only at $O(T^2)$, so $x = \sqrt{2\alpha}\, (1 + O(T^2))$: the extreme overlap itself is temperature independent at this order, and the temperature enters only through the gain,
\begin{equation}
    \Delta = \coupling (x - 1)
        + \coupling^2 T \Big[ (1 - x) + \frac{x^2}{2} \Big] + O(T^2).
    \label{eq:appB-ceiling-gain}
\end{equation}
Near the threshold $x \approx 1$ the bracket equals $\frac12$, so stability $\Delta < 0$ requires $x < x_c = 1 - \coupling T/2 + O(T^2)$, i.e.~$\alpha < x_c^2 / 2$, which is the ceiling \eref{eq:modelB-ceiling}.
The two order-$T$ mechanisms quoted in the main text are the two terms of \eref{eq:appB-defection-gain}, the exchange thinning of the condensate and the norm channel.
The boundary optimization ranges over all patterns actually present, so it automatically includes the defection toward the maximal-norm pattern ($x \approx 0$, $\varepsilon = \varepsilon_{\max}$), the first step of nucleation toward the F phase.
The solution of \eref{eq:appB-ceiling-lagrange} dominates it at low temperature, so the one-copy instability channels are exhausted by this computation.
The finite-temperature capacity thus follows \eref{eq:modelB-spinodal} on the annealed side and \eref{eq:modelB-ceiling} on the frozen side of the branch switch at $\alpha \approx \coupling^2/2 + O(T)$, as in \sref{sec:model_b_finite_temperature}.
The switch is the finite-temperature continuation of the branch point $\coupling = \sqrt{2 \alpha}$ of \eref{eq:modelB-leak-evaluation}.

\subsection{Complete phase diagram for continuous temperature}
\label{append:complete_phase_diagram}

The results of \appenref{append:model_b_finite_temperature_copy} are exact at the integer temperature points $n = \beta / \coupling \in \Z_{>0}$.
In this appendix we reconstruct the phase diagram from the visible representation, which is defined at every real temperature from the outset, and show that all formulas of the main text hold as stated for arbitrary real $\beta$.%
\footnote{The results in this subsection are obtained with the help of Claude Fable 5.}
Along the way the fate of the fluctuation determinant and of the hidden-sector zero mode is clarified, and the analytic continuation off the integer lattice is proved unique.
Throughout, $s \coloneq \beta / \coupling$ denotes the real (in the uniqueness argument, complex) continuation of the copy number, with $s = n$ on the lattice.

\ntextbf{Constrained partition function.}
Fix the visible overlap $m = \xv \cdot \xii{h}{v}_1 / \Nv$ with a typical pattern $\xii{h}{v}_1$ as a reaction coordinate and define $Z(m) \coloneq \int \mathcal{D}\xv\, \delta( \xv \cdot \xii{h}{v}_1 / \Nv - m )\, e^{- \beta E(\xv)}$, with $E$ the effective energy \eref{eq:modelB-adiabatic} and $\mathcal{D}\xv$ the Gaussian reference measure.
The orthogonal decomposition $\xv = m\, \xii{h}{v}_1 + \xv_\perp$ fixes the longitudinal component with unit Jacobian, and the fields become
\begin{equation}
    a_1 = \coupling \Nv m,
    \qquad
    a_\mu = \coupling \big( \Nv m\, x_\mu + \xi^\perp_\mu \cdot \xv_\perp \big),
    \quad \mu \geq 2,
    \label{eq:appB-constrained-fields}
\end{equation}
with $x_\mu \coloneq \xii{h}{v}_\mu \cdot \xii{h}{v}_1 / \Nv$, $\xi^\perp_\mu \coloneq \xii{h}{v}_\mu - x_\mu \xii{h}{v}_1$, and $y_\mu \coloneq \norm{\xi^\perp_\mu}^2 / \Nv$.
Splitting off the signal and writing the leak sum \eref{eq:modelB-leak-sum} in the present variables, $L = \sum_{\mu \geq 2} e^{a_\mu}$, together with $u \coloneq e^{-a_1} L$,
\begin{equation}
    Z(m) = e^{\Nv \beta ( m - m^2/2 )}\, \big\langle (1 + u)^{s} \big\rangle_\perp,
    \label{eq:appB-Zm}
\end{equation}
where $\langle \cdot \rangle_\perp$ averages over $\xv_\perp \sim \mathcal{N}(0, \beta^{-1} I_{\Nv - 1})$ and is normalized so that the term $u^0$ contributes $1$.
Conditioned on $\xii{h}{v}_1$, the variables $(x_\mu, y_\mu)$ are independent across $\mu$, with joint rate $I(x, y) = x^2/2 + I_1(y)$, which is \eref{eq:appB-defection-rate} in the variables $y = 1 + \varepsilon - x^2$.
The counting dichotomy of \appenref{append:model_b_zero_temperature} applies to them unchanged.

\ntextbf{Sector decomposition of the retrieval basin.}
By Newton's generalized binomial theorem, for every real (indeed complex) exponent $s$,
\begin{equation}
    (1 + u)^{s} = \sum_{k = 0}^{\infty} \binom{s}{k}\, u^{k},
    \qquad
    \binom{s}{k} = \frac{s (s-1) \cdots (s - k + 1)}{k!},
    \label{eq:appB-binomial}
\end{equation}
convergent for $u < 1$.
In the retrieval basin $u$ is exponentially small, term-by-term evaluation is justified, and the expansion decomposes $Z(m)$ into sectors labeled by the integer defection number $k$, of condensed weight $\mem_k = 1 - k \coupling T$: the attention lattice and its quantum $\coupling T$ are properties of the model at every real temperature, not artifacts of integer $n$.
Only the combinatorial coefficients $\binom{s}{k}$ are continued, and they are subexponential in $\Nv$.
The convergence condition $u < 1$ is the retrieval condition itself, so the expansion is precisely the excitation theory of the retrieval state.
The $k = 0$ sector gives $\varphi_0(m) = \beta ( m - m^2/2 )$, maximal at $m = 1$ with value $\beta/2$.
For $k = 1$, using the Gaussian generating function $\langle e^{\coupling\, \xi^\perp_\mu \cdot \xv_\perp} \rangle_\perp = e^{\Nv (\coupling^2 T / 2)\, y_\mu}$ and the counting dichotomy,
\begin{equation}
    \Delta_1(m)
        = \sup_{I(x,y) \leq \alpha} \Big[ \alpha - I(x, y) + \coupling m (x - 1) + \frac{\coupling^2 T}{2} y \Big].
    \label{eq:appB-sector1}
\end{equation}
The annealed stationary point in $y$ is governed by the identity
\begin{equation}
    - I_1(y_\ast) + \frac{t}{2}\, y_\ast = - \frac12 \log (1 - t),
    \qquad
    y_\ast = \frac{1}{1 - t},
    \label{eq:appB-det-identity}
\end{equation}
here at $t = \coupling^2 T$, while the stationary overlap is $x_\ast = \coupling m$, so that
\begin{equation}
    \Delta_1(m)
        = \alpha - \coupling m \Big( 1 - \frac{\coupling m}{2} \Big)
        - \frac12 \log \big( 1 - \coupling^2 T \big),
    \label{eq:appB-sector1-value}
\end{equation}
valid while $I(x_\ast, y_\ast) = \coupling^2 m^2 / 2 + \kappa(\coupling^2 T) \leq \alpha$.
Maximizing $\varphi_0 + \Delta_1$ over $m$ gives $m^\ast_1 = (1 - \coupling T)/(1 - \coupling^2 T)$, which is the overlap amplification $\mem_1 / A_1$ of \eref{eq:appB-overlap} evaluated in this sector, and the value $\varphi_{\mathrm{R}}(1 - \coupling T)$, i.e.~exactly the copy result \eref{eq:appB-landau} at $\mem = 1 - 1/n$, now for arbitrary real $\beta$.
In particular the one-quantum stability criterion, hence the spinodal \eref{eq:modelB-spinodal}, holds at every real temperature.
For general $k$ with distinct destinations, the transverse average factorizes over the $\Nv - 1$ transverse coordinates.
Per coordinate the destinations contribute $w = \sum_{j=1}^k z_j \sim \mathcal{N}(0, k)$ and
\begin{equation}
    \E\, e^{\frac{\coupling^2 T}{2} w^2}
        = \big( 1 - k \coupling^2 T \big)^{-1/2},
    \label{eq:appB-sector-k-det}
\end{equation}
which resums the destination norms and their mutual response to all orders.
The longitudinal tilts contribute $e^{\Nv \coupling^2 m^2 / 2}$ each, independently, since $m$ is held fixed.
With the signal loss $e^{- k \coupling m \Nv}$ and the counting $e^{k \alpha \Nv}$,
\begin{equation}
    \Delta_k(m)
        = k \Big[ \alpha - \coupling m \Big( 1 - \frac{\coupling m}{2} \Big) \Big]
        - \frac12 \log \big( 1 - k \coupling^2 T \big),
    \label{eq:appB-sector-k}
\end{equation}
and in terms of $\mem_k = 1 - k \coupling T$ and $A_k = 1 - \coupling (1 - \mem_k) = 1 - k \coupling^2 T$,
\begin{equation}
    \begin{aligned}
        \varphi_k(m)
            &= \varphi_0(m) + \Delta_k(m)  \\
            &= \frac{\beta}{\coupling} (1 - \mem_k)\, \alpha
            - \frac12 \log A_k
            + \beta \Big( \mem_k m - \frac{A_k}{2} m^2 \Big).
    \end{aligned}
    \label{eq:appB-landau-surface}
\end{equation}
This is the two-variable Landau surface $\varphi(m, \mem)$ of the model, derived exactly on the lattice $\mem_k = 1 - k \coupling T$ at arbitrary real temperature.
Maximizing over $m$ returns \eref{eq:appB-landau}, so the entire retrieval branch of \appenref{append:model_b_finite_temperature_copy} is recovered without any integer assumption.
The $m$ direction is exactly continuous and smooth, while the $\mem$ direction is quantized: the discreteness of the finite-temperature physics resides solely in the attention coordinate.
The validity conditions are the annealed counting condition per quantum, $\alpha \geq \coupling^2 m^2/2 + \kappa(\coupling^2 T)$, and the positivity $A_k > 0$, i.e.~$k < 1/(\coupling^2 T)$.
The physical lattice ends at $\mem_k \geq 0$, i.e.~$k \leq s$, beyond which the binomial coefficients alternate in sign and the bulk is described by the dual expansion below.

\ntextbf{Co-defection channels.}
The $u^k$ terms with coinciding destination indices describe $j$ copies defecting to the same pattern.
The coherent coupling is enhanced by $j^2$, giving
\begin{equation}
    \Delta_j^{\mathrm{co}}(m)
        = \alpha - j \coupling m + \frac{j^2 \coupling^2 m^2}{2}
        - \frac12 \log \big( 1 - j^2 \coupling^2 T \big),
    \label{eq:appB-codefection}
\end{equation}
with tilted overlap $x_\ast = j \coupling m$, which requires the existence condition $I = j^2 \coupling^2 m^2 / 2 \leq \alpha$.
Formally \eref{eq:appB-codefection} would overtake $j$ separate defections for $\alpha < \coupling^2 m^2$, but that region violates the existence condition already at $j = 2$ (which demands $\alpha \geq 2 \coupling^2 m^2$), so the annealed co-defection channel is never dominant where it is valid.
The frozen co-defection, $j$ quanta onto the most favorable pattern actually present, gains only $j$ times the single-quantum exchange term plus the $O(T)$ norm enhancement, and therefore does not destabilize retrieval before the one-copy channel does.
The spinodal thus remains the one-copy criterion, $\min$ of \eref{eq:modelB-spinodal} and \eref{eq:modelB-ceiling}, at every real temperature.
The role of the co-defection ladder is instead to provide the nucleation path toward the condensed phases, as shown below.
Similarly, the frozen-leak evaluation of \appenref{append:model_b_finite_temperature_copy} carries over at fixed $m$: the boundary optimization gives $\Delta_1^{\mathrm{froz}}(m) = \coupling m ( \sqrt{2\alpha} - 1 ) + \coupling^2 T / 2 + O(T^2)$, and relaxing $m$ to its optimum $m_\ast = 1 - \coupling T (1 - \sqrt{2 \alpha})$ yields
\begin{equation}
    \Delta^{\mathrm{froz}}
        = \coupling ( \sqrt{2 \alpha} - 1 )
        + \coupling^2 T \big[ 1 - \sqrt{2 \alpha} + \alpha \big] + O(T^2),
    \label{eq:appB-frozen-leak-m}
\end{equation}
which agrees term by term with \eref{eq:appB-ceiling-gain} at $x = \sqrt{2\alpha}$ (the exchange factor $1 - \coupling T$ of the copy computation reappears here as the $m$-relaxation term: the grouping of terms differs, the total does not), and reproduces the ceiling \eref{eq:modelB-ceiling} at real temperature.

\ntextbf{Bulk branches and endpoint closure.}
Outside the basin ($u > 1$) one expands dually, $(e^{a_1} + L)^{s} = L^{s} (1 + u^{-1})^{s}$: leak-dominated states into which signal is injected.
For the paramagnetic branch the annealed leak exponent $\Nv^{-1} \log L = \alpha + \coupling^2 \rho^2 / 2$ is linear in $\rho^2 = \norm{\xv}^2 / \Nv$, so the visible integral is exactly Gaussian and
\begin{equation}
    \varphi_{\mathrm{P}}(m)
        = s \alpha - \frac12 \log (1 - \coupling) - \frac{\beta (1 - \coupling)}{2} m^2,
    \label{eq:appB-P-branch}
\end{equation}
maximal at $m = 0$ where it reproduces $\varphi_{\mathrm{P}}$ of \appenref{append:model_b_finite_temperature_copy} with $n \to s$ real.
The longitudinal stiffness $\beta (1 - \coupling)$ is the inverse of the condensate size \eref{eq:appB-P-condensate}.
For the condensed sector, a state $\xv \approx \xii{h}{v}_\mu$ with overlap $x = m$ and norm $1 + \varepsilon = m^2 + y$ carries
\begin{equation}
    \varphi_{\mathrm{cond}}(m)
        = \sup_{y :\, \frac{m^2}{2} + I_1(y) \leq \alpha}
        \Big[ \alpha - \frac{m^2}{2} - I_1(y) + \frac{\beta}{2} \big( m^2 + y \big) \Big].
    \label{eq:appB-condensed-m}
\end{equation}
The interior stationary point, $y_\ast = (1 - \beta)^{-1}$ for $\beta < 1$ by \eref{eq:appB-det-identity} at $t = \beta$, gives
\begin{equation}
    \varphi_{\mathrm{C}}(m)
        = \alpha - \frac12 \log (1 - \beta) - \frac{1 - \beta}{2} m^2,
    \label{eq:appB-C-branch}
\end{equation}
whose existence condition $\beta < 1$ appears automatically.
The boundary evaluation, $I_1(y) = \alpha - m^2/2$, gives
\begin{equation}
    \varphi_{\mathrm{F}}(m)
        = \frac{\beta}{2} \Big[ m^2 + 1 + \varepsilon_{\max} \Big( \alpha - \frac{m^2}{2} \Big) \Big],
    \qquad
    m^2 \leq 2 \alpha,
    \label{eq:appB-F-branch}
\end{equation}
where $\varepsilon_{\max}(a)$ solves $I_1(1 + \varepsilon) = a$.
Since $\partial_m \varphi_{\mathrm{F}} = - \beta m / \varepsilon_{\max} < 0$, the frozen branch is maximal at $m = 0$: the extreme-norm pattern is typically orthogonal to $\xii{h}{v}_1$.
The freezing line at fixed $m$ is $\kappa(\beta) = \alpha - m^2/2$, reducing at $m = 0$ to \eref{eq:modelB-freezing-line}.
The decisive structural fact is that the defection lattice closes exactly onto these bulk branches.
Substituting the terminus $k = n$ of the distinct-defection ladder (at integer $n$, with $\mem = 0$, $k \coupling = \beta$, and $k \coupling^2 T = \coupling$) into \eref{eq:appB-landau-surface},
\begin{equation}
    \varphi_n(m)
        = n \alpha - \frac12 \log (1 - \coupling) - \frac{\beta (1 - \coupling)}{2} m^2
        = \varphi_{\mathrm{P}}(m),
    \label{eq:appB-closure-P}
\end{equation}
and the terminus $j = n$ of the co-defection ladder, whose gain $- \frac{\beta}{2} m^2 + \beta m x + \frac{\beta}{2} y$ completes the square as $\frac{\beta}{2} [ 1 + \varepsilon - (m - x)^2 ]$, is optimized on the boundary at $x = m$ and closes onto \eref{eq:appB-F-branch} (interior stationarity giving \eref{eq:appB-C-branch} instead).
The retrieval state is thus connected to the bulk phases by two quantized nucleation ladders living on a single surface, distinct defections leading to P and co-defections to C or F.
At non-integer $s$ the terminus $k = s$ is not a lattice point, but the dual expansion supplies the bulk values independently, and they agree with the $\mem \to 0$ values of \eref{eq:appB-landau-surface}, so the two expansions connect without a gap.

\ntextbf{Fluctuation determinant, zero mode, and uniqueness.}
Three structural questions raised in \appenref{append:model_b_copy_representation} are settled by this construction.

(a) \emph{One determinant, two interpretations.}
The only Gaussian fluctuation determinant surviving in the entire construction is the transverse visible one, and it admits two equivalent evaluations.
Averaging over the destination patterns first, $\E_{\xi^\perp} \exp [ \coupling \sum_j \xi^\perp_j \cdot \xv_\perp ] = \exp [ \frac{k \coupling^2}{2} \norm{\xv_\perp}^2 ]$, which renormalizes the transverse stiffness $\beta \to \beta - k \coupling^2 = \beta A_k$.
The ratio to the reference measure is
\begin{equation}
    \Big( \frac{\beta}{\beta - k \coupling^2} \Big)^{\!(\Nv - 1)/2}
        = A_k^{-(\Nv-1)/2}
        \;\Longrightarrow\;
        - \frac12 \log A_k,
    \label{eq:appB-two-readings}
\end{equation}
so the logarithm in \eref{eq:appB-sector-k} is the determinant of the visible transverse fluctuations softened by the attention leak, $A_k > 0$ is their stability condition, and $A_k \to 0$ at $k = s$ is the point $\coupling \to 1$: the paramagnetic instability.
Averaging over $\xv_\perp$ first instead gives the Gram-fluctuation evaluation \eref{eq:appB-sector-k-det}, whose large-deviation form is the variational problem $\sup_{Q^\perp} [ - I_k(Q^\perp) + \frac{\coupling^2 T}{2} \onevec^\top Q^\perp \onevec ]$.
The symmetric decomposition puts the transverse eigenvalues at $1$ and the coherent one at $\Lambda_\ast = (1 - k \coupling^2 T)^{-1}$, with the same value by \eref{eq:appB-det-identity}.
The two interpretations are the two sides of the Legendre pair \eref{eq:appB-legendre-pair} realized sector by sector: the energy-side $\log\det$ of a replica computation and the entropy-side rate function of the counting derivation are one object viewed from its two convex-dual sides.

(b) \emph{The fate of the zero mode.}
No pseudo-determinant appears above, and the reason is instructive.
The visible representation is written in terms of the log-sum-exp, i.e.~after the softmax gauge orbit of \appenref{append:model_b_copy_representation} has been fixed.
The divergent zero-mode volume $V_0$ of \eref{eq:appB-adiabatic-corrected} is a configuration-independent constant absorbed into the reference measure.
Conversely, a Gaussian fluctuation expansion of the $f$ representation around a retrieval vertex is structurally degenerate: by \eref{eq:appB-detp}, $\detp \hess = \Nh \prod_\mu \xf_\mu \to 0$ as $\xf \to e_1$, since every non-condensed weight is exponentially small, reflecting the divergence of the entropy curvature $\partial^2_{\xf} ( \xf \log \xf ) = 1/\xf$ at the boundary of the simplex.
A Gaussian theory of fluctuations around the vertex therefore does not exist.
The correct fluctuation theory is the discrete sector decomposition itself, whose elementary excitations are the quantized defections of attention weight $\coupling T$ and whose exact generating function is the binomial series \eref{eq:appB-binomial}.
This also sharpens the distinction, noted in \sref{sec:model_b_copy_representation}, from a variant model in which the attention entropy is rescaled by hand to be extensive: such a model equilibrates at an interior saddle point with macroscopically spread attention, where $1/\xf$ remains finite and the Gaussian expansion is legitimate.
The two models differ already at the level of their fluctuation theories, discrete versus Gaussian.

(c) \emph{Uniqueness of the continuation.}
At finite $\Nv$ and fixed disorder, $Z(m; \beta)$ is defined by the visible integral for every real $\beta > 0$ and is analytic in $\beta$.
The multinomial expansion at integer $n$ and the binomial expansion \eref{eq:appB-binomial} are exact rewritings of this single function on their respective domains, so within the retrieval basin there is no continuation to choose.
The only question is the exchange of the thermodynamic limit with the continuation, and in the basin the series \eref{eq:appB-binomial} converges locally uniformly in $\beta$, the dominant sector index is finite, and each $\Delta_k$ is analytic by \eref{eq:appB-sector-k}, so $\varphi(m; \beta)$ is analytic up to branch switches and continuous across them.
Agreement with the copy values at the integer points was verified at \eref{eq:appB-sector1-value}.
A stronger statement holds: the integer-point data alone already determine the answer.
By Carlson's theorem \cite[Sec.~9]{boas1954entire}, a function analytic in $\Re s \geq 0$, of exponential type with $|F(s)| \leq C e^{\tau |s|}$ for some $\tau < \pi$, and vanishing on the non-negative integers, vanishes identically.
In the basin, $|(1 + u)^s| = (1 + u)^{\Re s} \leq 2^{\Re s}$, so the candidate functions have type at most $\log 2 < \pi$, and the difference of any two continuations that agree on the lattice is identically zero.
The growth condition is what excludes the spurious continuations $\sin(\pi s)$, of type exactly $\pi$.
With it, the caveat usually attached to integer-power representations is removed for the entire retrieval basin.

\ntextbf{Assembly of the phase diagram.}
Table~\ref{tab:appB-branches} collects the branches constructed above with their validity conditions.
All live on the single $m$ axis and are connected through the defection lattice by the endpoint closure.

\begin{table}[t]
    \caption{
        Branches of the constrained rate $\varphi(m)$ at real temperature.
        The retrieval sectors live on the attention lattice $\mem_k = 1 - k \coupling T$.
        P, C, and F are maximal at $m = 0$, and R at $\mem = 1$.
        }
    \label{tab:appB-branches}
    \begin{ruledtabular}
        \begin{tabular}{lll}
            Branch & $\varphi(m)$ & Validity \\
            \colrule
            R sectors & \eref{eq:appB-landau-surface} & $u < 1$, $A_k > 0$, $\alpha \geq \frac{\coupling^2 m^2}{2} + \kappa(\coupling^2 T)$ \\
            R frozen leak & \eref{eq:appB-frozen-leak-m} & $u < 1$, $\alpha < \frac{\coupling^2 m^2}{2} + O(T^2)$ \\
            P & \eref{eq:appB-P-branch} & $\coupling < 1$ \\
            C & \eref{eq:appB-C-branch} & $\beta < 1$, $\alpha \geq \kappa(\beta) + \frac{m^2}{2}$ \\
            F & \eref{eq:appB-F-branch} & $m^2 \leq 2 \alpha$ \\
        \end{tabular}
    \end{ruledtabular}
\end{table}

\begin{figure*}[t]
    \centering
    \includegraphics[width=\textwidth]{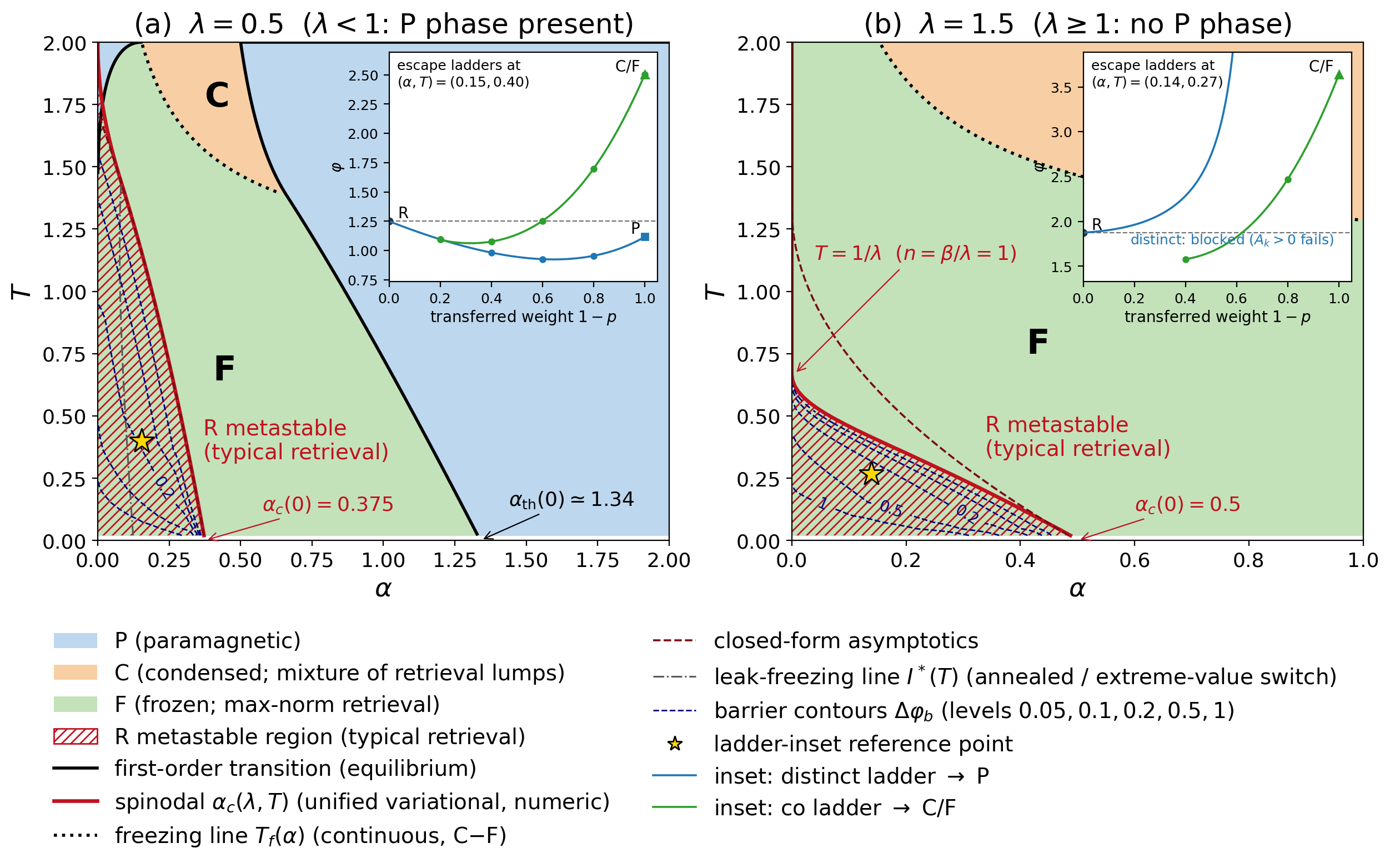}
    \caption{
        Same as \fref{fig:modelB-phase-diagram}, with the dynamical content of the real-temperature construction overlaid.
        The blue dashed curves are contours of the escape barrier \eref{eq:appB-barrier}, evaluated along the lower of the two escape ladders, at the levels $\Delta \varphi_{\mathrm{b}} = 0.05$, $0.1$, $0.2$, $0.5$, and $1$.
        The barrier vanishes on the spinodal and grows toward small $\alpha$ and low temperature.
        The insets show the constrained rate $\varphi$ along the two escape ladders at the starred reference points, $(\alpha, T) = (0.15, 0.40)$ in (a) and $(0.14, 0.27)$ in (b), against the transferred attention weight $1 - \mem$: the distinct-defection ladder (blue, annealed curve with rungs at $\mem_k = 1 - k \coupling T$) ends at the paramagnetic value, and the co-defection ladder (green) at the condensed value, whose frozen terminus reproduces $\varphi_{\mathrm{F}}$ exactly.
        In (b) the distinct ladder is cut off by the transverse stability condition $A_k > 0$ before reaching its terminus, and only the co-defection ladder connects retrieval to the bulk.
        }
    \label{fig:appB-phase-diagram-full}
\end{figure*}

The spinodal read off the table is the one-quantum criterion established above, with the annealed and frozen branches \eref{eq:modelB-spinodal} and \eref{eq:modelB-ceiling}, now as a statement at every real temperature.
The two branches follow from a single expression: eliminating $m$ at its saddle, $1 - m = \coupling T (1 - x)$, reduces the one-quantum criterion to the variational problem
\begin{equation}
    \Delta(\alpha; \coupling, T)
        = \max_{I(x, y) \leq \alpha}
        \Big[ \alpha - I(x, y) - \coupling (1 - x)
        + \frac{\coupling^2 T}{2} \big( (1 - x)^2 + y \big) \Big],
    \label{eq:appB-unified-spinodal}
\end{equation}
with retrieval locally stable while $\Delta < 0$.
The interior stationary point of \eref{eq:appB-unified-spinodal} reproduces \eref{eq:modelB-spinodal}, the boundary evaluation reproduces \eref{eq:modelB-ceiling}, the switch between the two is the existence line of the annealed destination, $\alpha = I^\ast(T) \coloneq (x^\ast)^2 / 2 + \kappa(\coupling^2 T)$ with $x^\ast$ the annealed saddle \eref{eq:appB-defection-saddle} (the dash-dotted curve of the figures), and the root $\Delta = 0$ is the spinodal drawn in \fref{fig:modelB-phase-diagram} and \fref{fig:appB-phase-diagram-full}.
The same surface carries the barrier.
The increments of \eref{eq:appB-sector-k} in $k$ start at $\alpha - \coupling m ( 1 - \coupling m / 2 )$ and grow convexly through the logarithm, so for $\alpha < \alpha_c$ the sector weights first decrease and then increase along the ladder.
The continuum stationary point is \eref{eq:appB-saddle-eq}, and the barrier height of the retrieval state is
\begin{equation}
    \Delta \varphi_{\mathrm{b}}
        = \frac{\beta}{2} - \varphi_{\mathrm{R}}(\mem^\dagger),
    \label{eq:appB-barrier}
\end{equation}
with $\mem^\dagger$ the sector corresponding to the root $m^\dagger$ of \eref{eq:appB-saddle-eq} through $m^\dagger = \mem^\dagger / A^\dagger$.
At $T = 0$, $\coupling m^\dagger = 1 - \sqrt{1 - 2 \alpha}$ gives a closed form, and at finite $T$ a single numerical root suffices.
In the Arrhenius sense $\tau_{\mathrm{esc}} \sim e^{\Nv \Delta \varphi_{\mathrm{b}}}$, the barrier turns the equilibrium construction into the lifetime map of \fref{fig:appB-phase-diagram-full}: the contours of $\Delta \varphi_{\mathrm{b}}$ fan out from the spinodal, on which the barrier vanishes, and grow toward small $\alpha$ and low temperature, so that retrieval is protected by an order-one rate barrier already at moderate depth inside the metastable region.
The insets resolve the competition between the two ladders of the endpoint closure at a fixed reference point.
For $\coupling = 0.5$ the co-defection ladder has both the lower barrier ($0.19$ against $0.34$ at the starred point) and a terminus, the frozen value $\varphi_{\mathrm{F}}$, far above the retrieval value $\beta/2$, while the distinct ladder terminates at a paramagnetic value below $\beta/2$: escape toward P is closed there and opens only at larger $\alpha$, where $\varphi_{\mathrm{P}}$ exceeds $\varphi_{\mathrm{R}}$, so distinct defections dominate the escape at large load and co-defections at small load and low temperature.
For $\coupling = 1.5$ the inset shows the distinct ladder running into the divergence of the determinant $- \frac12 \log A_k$ as $A_k \to 0$, so the escape proceeds through the co-defection ladder alone.
The general statement is given below.
The equilibrium boundaries follow by comparing the $m$-optimized branches.
Equating \eref{eq:appB-P-branch} and \eref{eq:appB-F-branch} at $m = 0$ gives the P--F line
\begin{equation}
    \frac{\alpha}{\coupling} + \frac{T}{2} \big| \log (1 - \coupling) \big|
        = \frac{1 + \varepsilon_{\max}(\alpha)}{2},
    \label{eq:appB-PF-boundary}
\end{equation}
whose $T \to 0$ limit is the zero-temperature equilibrium boundary of \appenref{append:model_b_zero_temperature}.
Equating \eref{eq:appB-P-branch} and \eref{eq:appB-C-branch} gives the P--C line
\begin{equation}
    (s - 1)\, \alpha = \frac12 \log \frac{1 - \coupling}{1 - \beta},
    \label{eq:appB-PC-boundary}
\end{equation}
both first order, while C--F is the continuous freezing line \eref{eq:modelB-freezing-line}.
The status of retrieval is unchanged from the main text: $\varphi_{\mathrm{F}} - \varphi_{\mathrm{R}}(1) = \frac{\beta}{2} \varepsilon_{\max} > 0$ at every $\alpha > 0$ and every temperature, so typical retrieval never becomes the equilibrium phase for Gaussian patterns.
For $\coupling \geq 1$ the transverse stability $A_k > 0$ fails at $k = 1/(\coupling^2 T) \leq s$: the distinct-defection ladder terminates before reaching its P terminus, which is the ladder-level manifestation of the absence of the paramagnetic phase.
The equilibrium is then C or F alone, with retrieval metastable below the ceiling \eref{eq:modelB-ceiling}.
This is the structure displayed in \fref{fig:modelB-phase-diagram} and, with the barrier contours and escape ladders overlaid, in \fref{fig:appB-phase-diagram-full}, now established at all real temperatures: relative to the integer-lattice derivation, every curve is promoted from an interpolation to an exact solution, the restriction that the lattice caps the accessible temperatures at $T = 1/\coupling$ disappears (so the C region stands for every $\coupling$), and the barrier \eref{eq:appB-barrier} adds dynamical information not visible in the equilibrium diagram.

\ntextbf{Remarks and open problems.}
We close with the points left open by the present analysis.
(i) \emph{Hidden-sector corrections.}
The shifts of \eref{eq:appB-shifted-copy}, the exponent $\beta/\coupling \to \beta/\coupling + \Nh/2$ with its constraint $\bar{n} > \Nh/2$ and the pattern shift $- \frac12 \sum_\nu \xii{h}{v}_\nu$, are not small at exponential load.
Whether they can be absorbed as a change of the reference measure within the sector decomposition, leaving the rate-level phase diagram intact, is the main structural question left open.
(ii) \emph{The basin edge.}
At $u \to 1$ the binomial coefficients alternate in sign and the series requires resummation.
A uniform treatment of the basin edge, e.g.~through the Mellin--Barnes representation of $(1 + u)^s$ \cite{paris2001asymptotics}, would give the precise structure on the spinodal line itself, though the location of the spinodal, fixed by the $k = 0$ versus $k = 1$ comparison, is not affected.
(iii) \emph{Subexponential corrections.}
The prefactors of $\binom{s}{k}$, the fluctuation determinants collected here only at rate level, and the resulting finite-size shifts of $\alpha_c$, of order $\log \Nv / \Nv$, are relevant to numerical verification of the $T$ slopes in \eref{eq:modelB-spinodal} and \eref{eq:modelB-ceiling}.
(iv) \emph{Stability beyond one step.}
Within each sector the freezing prescription is exact (\appenref{append:model_b_copy_representation}), but fluctuations between sectors (an analogue of the AT condition) and the predicted Ruelle statistics of the F-phase attention weights \cite{ruelle1987mathematical} remain unexamined.
(v) \emph{Ensemble dependence.}
For spherical patterns $\varepsilon_{\max} \equiv 0$ and the F phase loses its advantage, so typical retrieval can equilibrate, while for binary patterns the Gaussian overlap rate $x^2/2$ is replaced by a binary entropy.
Redoing the construction for these ensembles and matching \cite{lucibello2024exponential,demircigil2017model} quantitatively would separate what is universal in the phase diagram from what is a norm-fluctuation effect.
(vi) \emph{Dynamics.}
The barrier \eref{eq:appB-barrier} and the two escape ladders are equilibrium statements about a reaction coordinate.
Connecting them to genuine relaxation dynamics, nucleation prefactors, and the multi-step retrieval dynamics of attention networks is left for future work.